%% file: main.tex
\documentclass{article}

\usepackage[utf8]{inputenc}
\usepackage{float}
\usepackage[T1]{fontenc}
\usepackage{hyperref}
\usepackage{url}
\usepackage{booktabs}
\usepackage{amsfonts}
\usepackage{amsmath}
\usepackage{graphicx}
\usepackage{microtype}
\usepackage{xcolor}
\usepackage{subcaption}
\usepackage{tikz}
\usetikzlibrary{calc,positioning}
\usepackage[final,main]{neurips_2026}

\hypersetup{
    colorlinks=true,
    linkcolor=blue,
    citecolor=blue,
    urlcolor=blue
}

\newcommand{\fblock}{g}
\newcommand{\mypar}[1]{\textbf{#1.}}
\definecolor{myorange}{HTML}{D95F02}  
\definecolor{mygreen}{HTML}{1B7F3A}

\usetikzlibrary{decorations.pathreplacing}

\title{Two Stages of Folding: Convergent Mechanisms in AI Protein Folding Trunks}
\author{%
  \textbf{Kevin Lu}\textsuperscript{1\,*\,\textdaggerdbl} \quad
  \textbf{Jannik Brinkmann}\textsuperscript{1\,2\,\textdagger} \quad
  \textbf{Stefan Huber}\textsuperscript{3\,\textdagger} \quad
  \textbf{Aaron Mueller}\textsuperscript{4} \\[2pt]
  \textbf{Yonatan Belinkov}\textsuperscript{3\,5} \quad
  \textbf{David Bau}\textsuperscript{1\,\textdagger} \quad
  \textbf{Chris Wendler}\textsuperscript{1\,*\,\textdaggerdbl}%
}

\begin{document}
\maketitle

\renewcommand{\thefootnote}{}
\footnotetext{%
  \textsuperscript{1}Northeastern University \quad
  \textsuperscript{2}TU Clausthal \quad
  \textsuperscript{3}Harvard University \quad
  \textsuperscript{4}Boston University \quad
  \textsuperscript{5}Technion -- Israel Institute of Technology
  \textsuperscript{*}Lead author. \quad
  \textsuperscript{\textdagger}Core contributor. \quad
  \textsuperscript{\textdaggerdbl}Corresp. authors: lu.kev@northeastern.edu, ch.wendler@northeastern.edu%
}
\renewcommand{\thefootnote}{\arabic{footnote}}
\vspace{-20pt}

\begin{figure}[h!]
    \centering
    \includegraphics[trim=0bp 0.000bp 0bp 2.000bp,clip,width=0.90\linewidth]{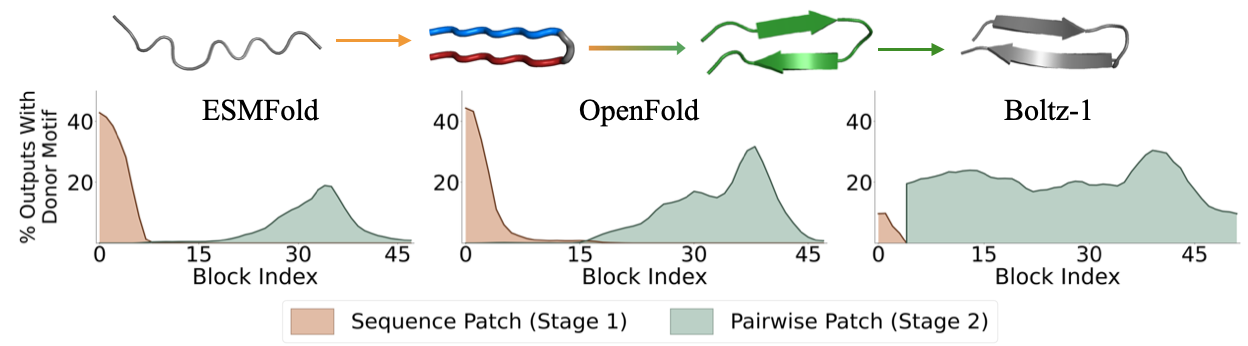}
    \caption{\textbf{Two computational stages in AI protein folding trunks.}
We identify which latent representations in ESMFold, OpenFold, and Boltz-1 influence secondary structure formation by patching activations from a hairpin protein into a helical protein at each block of the trunk, then measuring whether the output folds as a hairpin (or vice versa). Sequence patches (\textcolor{myorange}{orange}) induce hairpin formation in early blocks; pairwise patches (\textcolor{mygreen}{green}) are effective in later blocks.}
    \label{fig:stages_of_computation}
\end{figure}

\begin{abstract}
How do protein structure prediction models fold proteins? We investigate this question through causal interventions on the folding trunks of ESMFold, OpenFold, and Boltz-1. Across all three models, we find a shared two-stage computational structure for certain structural decisions. In the first stage, early blocks initialize \textit{pairwise biochemical signals}: features like charge propagate from sequence into pairwise representations through architecture-specific pathways. In the second stage, late blocks develop \textit{pairwise spatial features}: distance and contact information accumulate in the pairwise representation. We support this mechanistic account with causal evidence, showing that steering charge and distance features induces predictable structural changes. Furthermore, these representations support partial functional transfer: pairwise states can be linearly aligned and substituted across models. Together, these results suggest that folding trunks with different architectures, inputs, and training procedures converge on a shared representational organization for mapping sequence chemistry into spatial geometry.
\end{abstract}

\section{Introduction}
Have neural protein folding models like AlphaFold2~\citep{jumper2021highly} truly learned the physical principles of how proteins fold? Our community has studied these models extensively as predictors, with benchmarks, scaling curves, and downstream applications, but comparatively little on what they have figured out about folding itself. We ask whether physically interpretable decisions can be identified within the internal computations of these models, conducting the first cross-architectural mechanistic account of three modern folding models: ESMFold~\citep{lin2023evolutionary}, OpenFold~\citep{Ahdritz2024-yd}, and Boltz-1~\citep{Wohlwend2024.11.19.624167}.

We begin with alpha helices and beta hairpins, using motif transfer as a controlled way to change a model's structural decision. Hairpins are among the simplest motifs requiring coordination between separated strands~\citep{munoz1998statistical,blanco1998formation}, making them a natural starting point for tracing how sequence information becomes geometry. We then extend the analysis to four-strand sheets, contacts between distant sequence regions, and pathway measurements across 1,000 diverse proteins. We ask three questions:

\textbf{\emph{When} does a folding model commit to a structural decision?} Using activation  patching~\citep{vig2020causal,meng2022locating} we localize the computation that determines whether a region folds as a
hairpin or as a helix, identifying a two-stage structure shared across all three architectures: an early sequence-driven commitment phase and a late pairwise-refinement phase. In Boltz-1, the early phase is compressed into the upstream MSA module, followed by a broad window of pairwise influence.

\textbf{\emph{What} features drive each stage?} Using linear probing and representation steering, we find that stage 1 writes biochemical signals (like charge and hydrophobicity related information) into pairwise space. Steering these features induces strand attraction, repulsion, and even solvent exposure. Stage 2 develops pairwise representations as approximate distance maps, and steering them induces contacts and hairpin-like structure.

\textbf{Do \emph{different} architectures share an internal organization?} Using linear projections between pairwise representations of pairs of models, we find that representations from one model are partially functionally transplantable: states from one model can drive structural changes in another at the corresponding stage of computation. Folding trunks trained with different recipes, encoders, and decoders converge on a shared internal organization for mapping sequence chemistry into spatial geometry.

\section{AI Protein Folding Background}
For background on protein structure (amino acids, secondary structure motifs like alpha helices and beta hairpins), see Appendix~\ref{app:protein_structure} and \citet{branden1999introduction}.

\begin{figure}[t]
    \centering
    \begin{subfigure}[t]{0.48\linewidth}
        \centering
        \includegraphics[width=\linewidth]{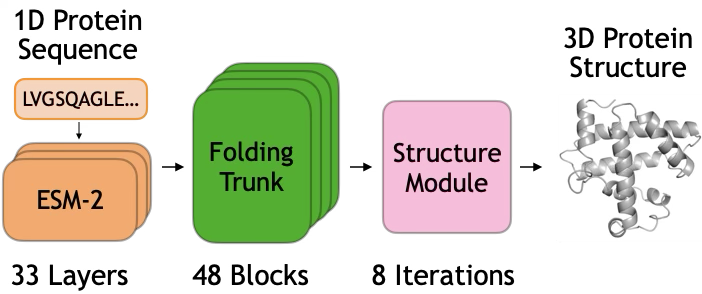}
        \caption{\textbf{ESMFold.} A protein language model (ESM-2) encodes an amino acid sequence into an initial sequence representation; the pairwise representation is initialized with learned positional embeddings. The folding trunk iteratively refines both representations over $48$ blocks (consisting of multiple layers each). The structure module converts these into 3D coordinates for each residue. Adapted from \citep{lin2023evolutionary}.}
        \label{fig:model_background}
    \end{subfigure}
    \hfill
    \begin{subfigure}[t]{0.48\linewidth}
        \centering
        \includegraphics[width=\linewidth]{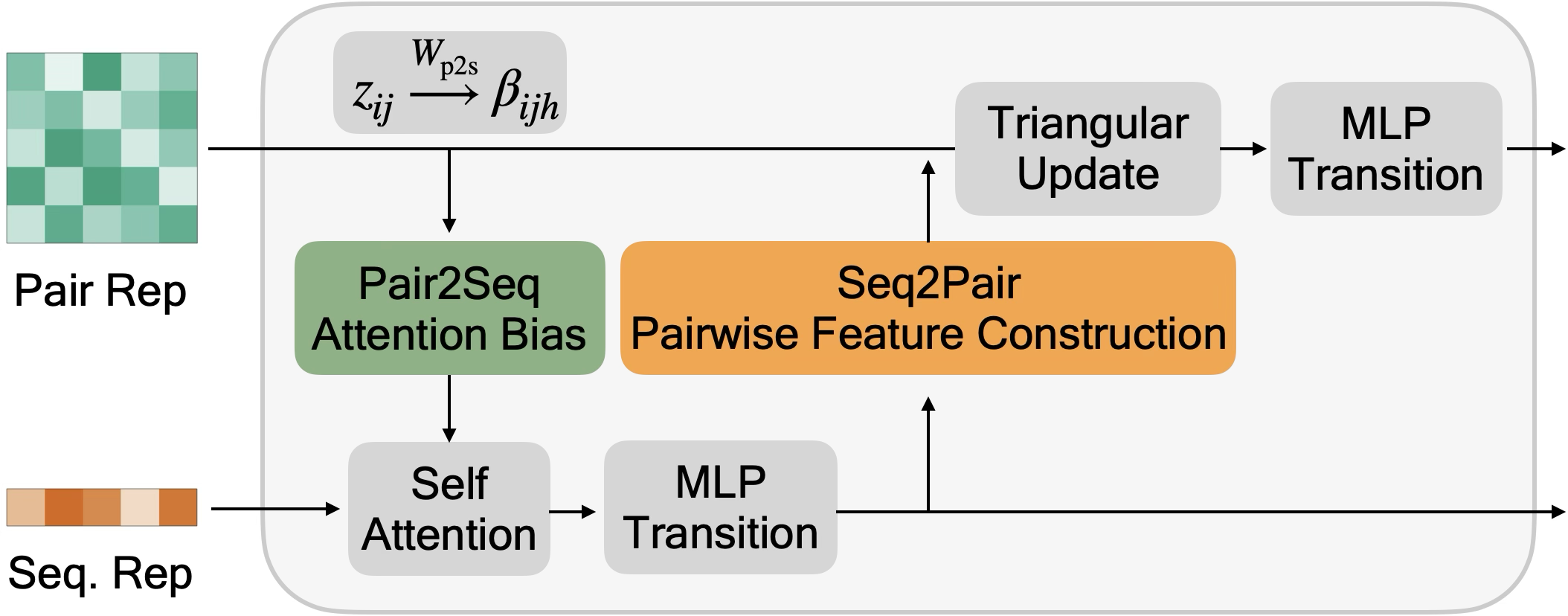}
        \caption{\textbf{Folding block.} \emph{A sequence update} consisting of \texttt{Pair2Seq} and a sequence transformer layer (attention plus MLP) followed by a \emph{pairwise update} consisting of \texttt{Seq2Pair} and multiplicative and attention based updates of the pairwise representation (triangular update, see \citep{jumper2021highly}) plus MLP. Adapted from \citep{lin2023evolutionary}.}
        \label{fig:folding_block}
    \end{subfigure}
    \caption{ESMFold architecture and folding block.}
    \label{fig:architecture_overview}
\end{figure}

We provide background on ESMFold as an example (Fig.~\ref{fig:model_background}), which consists of three modules. (1)~A protein language model (ESM-2~\citep{lin2023evolutionary}) maps an amino acid sequence of length $L$ to an initial sequence representation $s^{(0)} \in \mathbb{R}^{L \times d_s}$. (2)~A \emph{folding trunk} iteratively refines $s$ alongside a pairwise representation $z \in \mathbb{R}^{L \times L \times d_z}$ (initialized from learned positional embeddings) over $K{=}48$ blocks. (3)~A structure module based on invariant point attention~\citep{jumper2021highly} decodes the final $(s^{(K)}, z^{(K)})$ into 3D coordinates. We focus on the folding trunk.

\mypar{Folding trunk} Each of the $K{=}48$ blocks (Fig.~\ref{fig:folding_block}) applies a \emph{sequence update} followed by a \emph{pairwise update}, with residual connections throughout:
$(s^{(k)}, z^{(k)}) = \fblock^{(k)}(s^{(k-1)}, z^{(k-1)})$.

\mypar{Sequence update} Self-attention is modulated by a \texttt{pair2seq} bias derived from $z$:
\begin{equation}\label{eq:pair2seq}
  A^{(h)}_{ij} = \frac{\langle q^{(h)}_i, k^{(h)}_j \rangle}{\sqrt{d_h}} + \beta_{ijh}(z_{ij}),
\end{equation}
where $\beta_{ijh}(z_{ij}) = (W_{\beta} z_{ij})_h$ is a learned linear projection producing one scalar bias per attention head.

\mypar{Pairwise update} A \texttt{seq2pair} operation constructs pairwise features from sequence representations:
\begin{equation}\label{eq:seq2pair}
  \phi_{ij} = \bigl[u_i \odot v_j \;;\; u_i - v_j\bigr],
\end{equation}
where $u_i = W_u s_i$ and $v_j = W_v s_j$ are linear projections of $s$. A learned projection maps these features into an update to $z$, which is then refined via triangular multiplicative updates and attention~\citep{jumper2021highly}.

\mypar{OpenFold and Boltz-1} OpenFold~\citep{Ahdritz2024-yd} is a faithful reimplementation of AlphaFold2 with a 48-block Evoformer trunk. Boltz-1~\citep{Wohlwend2024.11.19.624167} is an AlphaFold-3-style architecture with an MSA module followed by a 48-block Pairformer. Both use multiple sequence alignments (MSAs) and maintain sequence and pairwise representations analogous to ESMFold's. ESMFold and OpenFold exchange information in both directions throughout the trunk (\texttt{seq2pair} and \texttt{pair2seq} in every block). Boltz-1 confines \texttt{seq2pair} to its upstream MSA module (which we treat as blocks 0–3 of the trunk). Within the remaining Pairformer blocks, the only cross-representation pathway is \texttt{pair2seq}: sequence representations are updated by pair-biased self-attention and an MLP, but never write back into the pairwise track.

\section{When Does the Model Fold a Hairpin or Helix?}
\label{sec:patching_setup}

\begin{figure*}[t]
    \centering
    \includegraphics[trim=0bp 6.000bp 0bp 0.000bp,clip,width=0.85\linewidth]{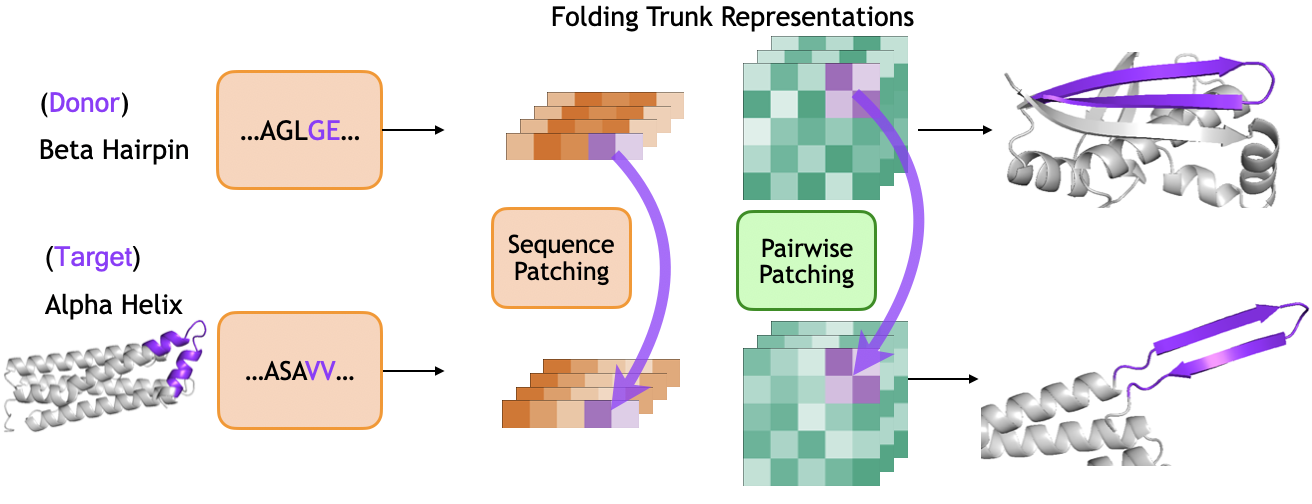}
    \caption{\textbf{Activation patching setup.} We extract donor hairpin representations (top) and replace the aligned target motif's sequence states (orange), pairwise states (green), or both during a selected trunk block. Purple marks the motif. We measure whether the output adopts the donor motif; the helix-donor experiment reverses the motif roles.}
    \label{fig:activation_patching}
\end{figure*}

\mypar{Activation patching setup} We use activation patching to find the blocks at which replacing internal states changes the predicted motif. We select a \emph{donor protein} containing a secondary structure motif (a beta hairpin or alpha-helical region) and a \emph{target protein} containing the opposite motif. We run both proteins through the folding trunk, extracting the sequence representation $s^{(k)}$ and pairwise representation $z^{(k)}$ at each block $k$. During the target's forward pass, we replace representations in the target's motif region with the donor's. We then observe whether the output structure contains the donor motif in the patched region (Fig.~\ref{fig:activation_patching}). We disable recycling to isolate a single pass through the trunk (App.~\ref{app:recycling}).

\mypar{Dataset construction} We selected 200 target proteins and approximately 130,000 donor motif regions from the PDB~\citep{berman2000protein}. Using DSSP~\citep{kabsch1983dictionary}, we identified motif regions in each target, sampled 10 donors per motif, and aligned motif positions to define 15--20 residue intervention regions, yielding $\approx$10,000 patching experiments (App.~\ref{app:experiment_datasets}).

\mypar{Full patching establishes feasibility} We first verify that patching can induce donor motif formation by patching both sequence and pairwise representations across all 48 blocks of the folding trunk. Approximately 40\% of patches ($\approx$4,000 cases) successfully produce the donor motif in the target region, as measured by DSSP secondary structure assignment (App.~\ref{app:hairpin_detection_algorithm}). In ESMFold, patching at the encoder or structure module is substantially less effective (App.~\ref{app:other_modules}), identifying the trunk as the most effective site for motif transfer in this comparison.

\mypar{Single-block patching reveals two regimes across architectures} To localize the computation, we repeat the patching experiment but intervene at a single block, patching either sequence or pairwise representations alone (restricted to the $\approx$4,000 donor--target pairs where full patching was successful).
Sequence patches are most effective early, while pairwise patches become effective later in ESMFold and OpenFold (Fig.~\ref{fig:stages_of_computation}). Repeating the analysis over all donor--target pairs lowers success rates but preserves the early/late pattern (App.~\ref{app:revision_scope}). We patch sequence positions in the aligned motif and, by default, pairwise entries with both indices in that region. A broader ``touch'' mask extends the effective pairwise window (App.~\ref{app:touch_patching}). We report the effective windows for each mask, since the patch must also integrate with the surrounding structure.

\mypar{Structural quality and scope} We evaluate intervention success alongside chain integrity, clashes, confidence, design/refold self-consistency TM-score (scTM), and motif recovery (App.~\ref{app:revision_quality}, Table~\ref{tab:quality_independent_revision}). Filtering invalid or non-designable outputs lowers success but preserves the localization. After filtering, Boltz-1 pairwise patches remain effective from blocks 3--47, covering most of the analyzed network rather than a narrow late interval.

The ordering also appears in four-strand-sheet patches spanning 40--60 residues, long-range contacts with median sequence separation 132 residues, and length-stratified tests up to 1,000 residues (App.~\ref{app:revision_scope}). Long-range pairwise patches induce contacts in 46--73\% of cases, versus 6--9\% for norm-matched random perturbations. We check whether the new contacts match the donor and whether the rest of the protein contracts, to distinguish targeted transfer from global collapse. Across 1,000 diverse proteins, motif-independent sequence-to-pair updates peak at blocks 1--2, before pair-to-sequence updates at blocks 11--34. Together, these experiments extend the two-stage organization from local motif transfer to larger structures, long-range interactions, and whole-protein information flow.

\section{Early Blocks: Building Pairwise Chemistry}
\subsection{Sequence Information Flows Into Pairwise Space}
\label{sec:information_flow}

Why is sequence patching only effective in early blocks? We hypothesize that early blocks serve a specific role: transferring biochemical information from sequence representations into pairwise representations, where it remains accessible for downstream geometric computation. We present the analysis in detail for ESMFold using 400 successful cases of block-0 sequence patching, and report parallel results for OpenFold and Boltz-1 in App.~\ref{app:triple_model_info_flow}.

Within the ESMFold folding trunk, sequence representations $s$ and pairwise representations $z$ interact through two pathways (Figure \ref{fig:folding_block}). The \texttt{seq2pair} pathway converts the sequence representation into a pairwise update via Eq.~\eqref{eq:seq2pair}. The \texttt{pair2seq} pathway projects $z$ into a scalar bias for each attention head and residue pair, which modulates sequence self-attention, via Eq.~\eqref{eq:pair2seq}.

\begin{figure}[t]
    \centering
    \includegraphics[trim=0bp 4.000bp 0bp 13.001bp,clip,width=1.0\linewidth]{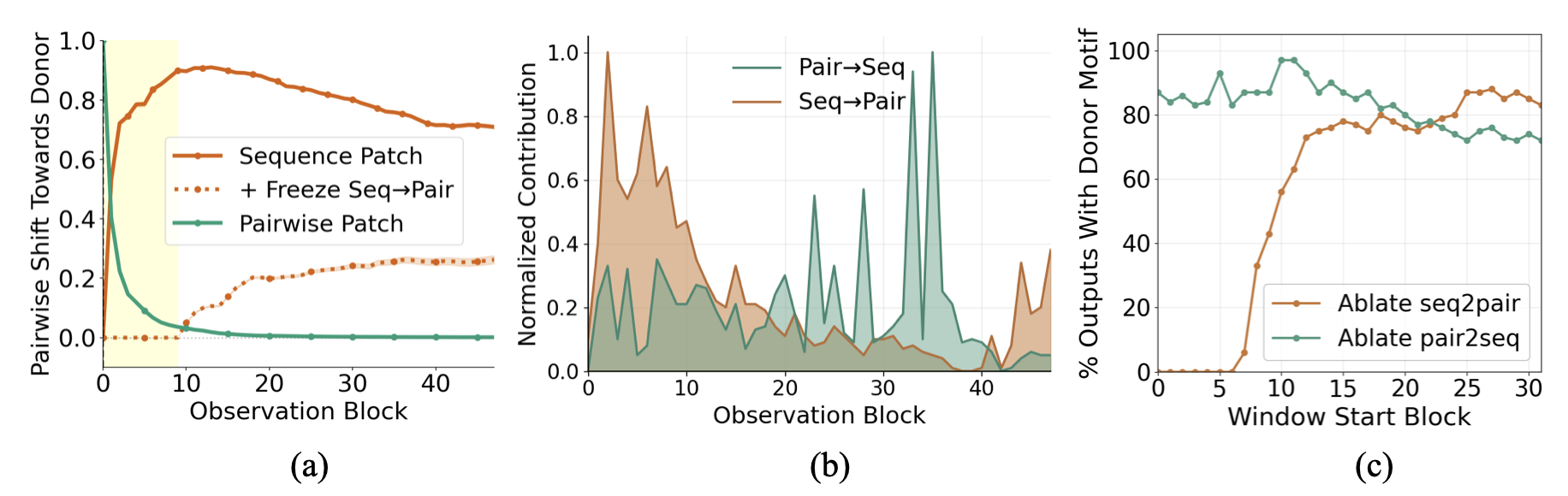}
    \caption{\textbf{Information flows from sequence into pairwise representations during a short early window.} (ESMFold; replicated for OpenFold and Boltz-1 in App.~\ref{app:triple_model_info_flow})
    \textbf{(a)} Shift in $z$ during the target forward pass under three interventions: sequence patching at block 0, sequence patching with \texttt{seq2pair} frozen during blocks 0--10, and pairwise patching at block 0. Sequence patching shifts $z$ toward the donor (Eq.~\eqref{eq:representation_similarity}), but freezing \texttt{seq2pair} during the early window prevents this shift.
    \textbf{(b)} Normalized contribution magnitudes of \texttt{seq2pair} (orange) and \texttt{pair2seq} (green) at each block. \texttt{seq2pair} peaks early; \texttt{pair2seq} peaks later, mirroring the staged patching pattern.
    \textbf{(c)} Sliding-window ablation (window size 15) during block-0 sequence patching. Ablating \texttt{seq2pair} early sharply reduces motif formation.}
    \label{fig:information_flow}
    \vspace{-5pt}
\end{figure}

\mypar{Representation similarity across layers} To quantify how much the patched representation resembles the donor versus the target, we compute an interpolation coefficient:
\begin{equation}\label{eq:representation_similarity}
    \alpha = \frac{\langle z_{\text{patched}} - z_{\text{target}}, z_{\text{donor}} - z_{\text{target}} \rangle}{\|z_{\text{donor}} - z_{\text{target}}\|^2}
\end{equation}
Here $\alpha$ measures how far the patched state moves from the target toward the donor along the donor--target direction. Values of 0 and 1 match the target and donor on this axis; the states can still differ in other directions.

After sequence patching at block 0, $z$ rapidly becomes donor-like within the first $\approx$10 blocks, then changes only gradually (Fig.~\ref{fig:information_flow}a). Blocking \texttt{seq2pair} during blocks 0--10 prevents this shift entirely, while directly patching $z$ at block 0 does not persist: the pairwise state returns toward the target trajectory. This confirms a short early ``write-in'' window during which sequence information is consolidated into $z$ via \texttt{seq2pair}.

\mypar{Pathway Contributions} The asymmetric importance of \texttt{seq2pair} and \texttt{pair2seq} across blocks is also visible without intervention. We measure the relative contribution of each pathway at every block by computing the norm of the corresponding update at that block, normalized within each pathway across the trunk. The \texttt{seq2pair} contribution to $z$ peaks in the first ten blocks and declines through the middle of the trunk, while \texttt{pair2seq} is small in early blocks and rises sharply in late blocks (Fig. ~\ref{fig:information_flow}b). The two pathways are not just \emph{used} at different times: their update magnitudes themselves follow the staged pattern, with sequence-to-pair flow concentrated early and pair-to-sequence flow concentrated late.

\mypar{Pathway ablations} Sliding-window ablations confirm the directionality of the early-window effect (Fig.~\ref{fig:information_flow}c). When the ablation window starts in early blocks (0–7), ablating \texttt{seq2pair} during block-0 sequence patching reduces hairpin formation to near zero; the effect recovers fully once the window starts past block 12, indicating that \texttt{seq2pair} is specifically required during the early write-in phase. As a control, ablating \texttt{pair2seq} has little effect at early window positions, confirming that the loss of structure is specific to the sequence-to-pairwise pathway rather than a generic disruption from ablating either inter-representation pathway.

\mypar{Generalization across architectures} We repeat all three analyses on OpenFold and Boltz-1 (App.~\ref{app:triple_model_info_flow}, Fig.~\ref{fig:triple_model_info_flow}). OpenFold and Boltz-1 show the same sequence of events: sequence patches shift $z$ toward the donor within an early write-in window, freezing \texttt{seq2pair} prevents the shift, \texttt{seq2pair} contributions concentrate in early blocks, and early-block \texttt{seq2pair} ablations reduce motif formation.

\begin{figure*}[t]
    \centering
    \includegraphics[trim=0bp 1.000bp 0bp 1.500bp,clip,width=0.9\linewidth]{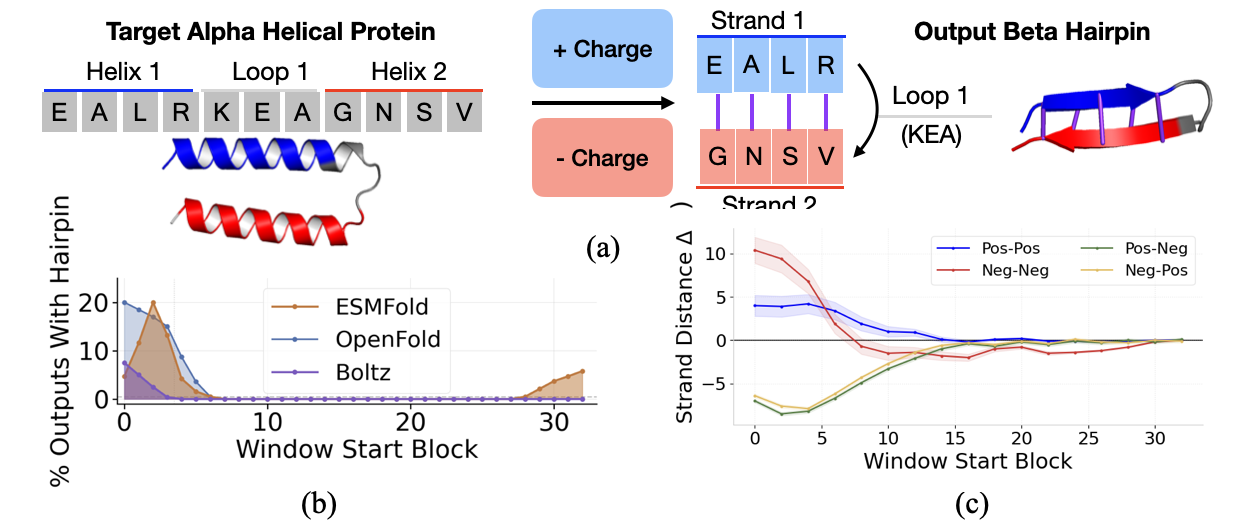}
\caption{\textbf{Charge steering induces hairpin formation.} 
(a) Intervention setup: we steer the sequence representation toward opposite charges on the two helical regions flanking the loop, motivated by electrostatic complementarity. 
(b) Early-block steering (window size 15) induces hairpins in all three models. 
(c) ESMFold control: same-charge steering on $\beta$-hairpin strands and opposite-charge steering on $\alpha$-helical targets produce sign-sensitive changes (strands separate; helices approach), concentrated in early blocks.}
\vspace{-10pt}
    \label{fig:charge_boom}
\end{figure*}

\subsection{Which Sequence Features are Propagated?}\label{sec:biochem-features}

What biochemical information passes through the early sequence-to-pair pathway? We first focus on charge, which is relevant to hairpin stability: beta-hairpins can be stabilized by cross-strand salt bridges between oppositely charged residues (e.g., lysine (K) and glutamate (E)) on facing positions~\citep{doi:10.1021/ja030074l}. We illustrate the construction and controls on ESMFold and report cross-architectural results in Fig.~\ref{fig:charge_boom}b.

\mypar{Charge information is linearly decodable} To test whether charge-related information can be read out linearly, we use a difference-in-means approach to identify a ``charge direction'' in the sequence representation space. We collect sequence representations from a new set of 200 alpha-helical proteins sampled from the PDB (using helical proteins avoids confounding charge with the specific distance patterns present in beta hairpins). Let $\mathcal{P} = \{\text{K}, \text{R}, \text{H}\}$ denote positively charged residues and $\mathcal{N} = \{\text{D}, \text{E}\}$ denote negatively charged residues. We compute the charge direction as:
\begin{equation}\label{eq:dom}
\begin{aligned}
v_{\text{charge}}
&= \frac{\bar{s}_{\mathcal{P}} - \bar{s}_{\mathcal{N}}}{\|\bar{s}_{\mathcal{P}} - \bar{s}_{\mathcal{N}}\|}, \\
\text{where}\quad
\bar{s}_{\mathcal{P}}
&= \frac{1}{|\mathcal{P}|} \sum_{a \in \mathcal{P}} \mathbb{E}[s_i \mid x_i = a],
\end{aligned}
\end{equation}
and analogously for $\bar{s}_{\mathcal{N}}$. Projections onto $v_{\text{charge}}$ separate charge classes with high ROC-AUC across all three models throughout the trunk (App.~\ref{app:charge_doms}), showing that a linear readout can distinguish the charge labels. Linear probing on $z$ shows that charge information enters the pairwise representation through each model's sequence-to-pairwise pathway in  early trunk blocks (App.~\ref{app:charge_probing}). Is this information used to drive folding decisions?

\mypar{Electrostatic complementarity steering} We add or subtract $v_{\text{charge}}$ to steer the internal representation toward positive or negative charge while keeping the amino acid sequence fixed. For a target helix-turn-helix region with residue indices $\mathcal{S}_1$ (helix 1) and $\mathcal{S}_2$ (helix 2), we steer $s_i' = s_i + \alpha v_{\text{charge}}$ for $i \in \mathcal{S}_1$ and $s_i' = s_i - \alpha v_{\text{charge}}$ for $i \in \mathcal{S}_2$, leaving other positions unchanged. The strength $\alpha$ controls the magnitude, and the construction mimics the cross-strand electrostatic complementarity found in natural hairpins (Fig.~\ref{fig:charge_boom}a).

\mypar{Steering induces hairpin formation across architectures} We apply this intervention to motif regions in our 100 target helical proteins, yielding approximately 500 cases per model, with steering strength $\alpha=3\sigma$. Complementarity steering induces hairpin formation in all three models, with the effect concentrated in early blocks coinciding with stage 1 (Fig.~\ref{fig:charge_boom}b).

\mypar{Sign and magnitude controls} Two controls test whether the effect is sign-sensitive and graded. \emph{Same-charge steering} on 400 sampled beta-hairpins (steering both strands to the same charge) increases mean cross-strand distance by $\approx$10\AA{} in early blocks (Fig.~\ref{fig:charge_boom}c, Pos-Pos/Neg-Neg), and opposite-charge steering on $\alpha$-helical targets decreases it (Pos-Neg/Neg-Pos), as expected from the attraction/repulsion analogy that motivated the experiment. \emph{Scaling charge strength} on small single-motif targets ($\leq$40 residues; App.~\ref{app:charge_size_scaling}) shows that cross-strand distance changes smoothly and monotonically as steering strength increases in all three models.

\mypar{Matched controls and a second biochemical feature} Norm- and variance-matched random directions have weaker effects than charge steering (App.~\ref{app:revision_biochem}). To test whether the early pathway also carries information about hydrophobicity, we derive a hydrophobicity direction from uncharged sets $\{I,V,L,F,M,A\}$ and $\{N,Q,S,T,G\}$ using Eq.~\eqref{eq:dom}. Steering opposing faces of amphipathic helices increases exposure of the originally hydrophobic face by 43, 39, and 24 percentage points in ESMFold, OpenFold, and Boltz-1, respectively ($n=500$ targets). Effects localize to early blocks, including Boltz-1's MSA module. The distinct geometric effects of charge and hydrophobicity steering show that the early pathway carries multiple biochemically meaningful features that causally influence structure.

\begin{figure}
    \centering
    \includegraphics[trim=0bp 1.500bp 0bp 9.000bp,clip,width=0.95\linewidth]{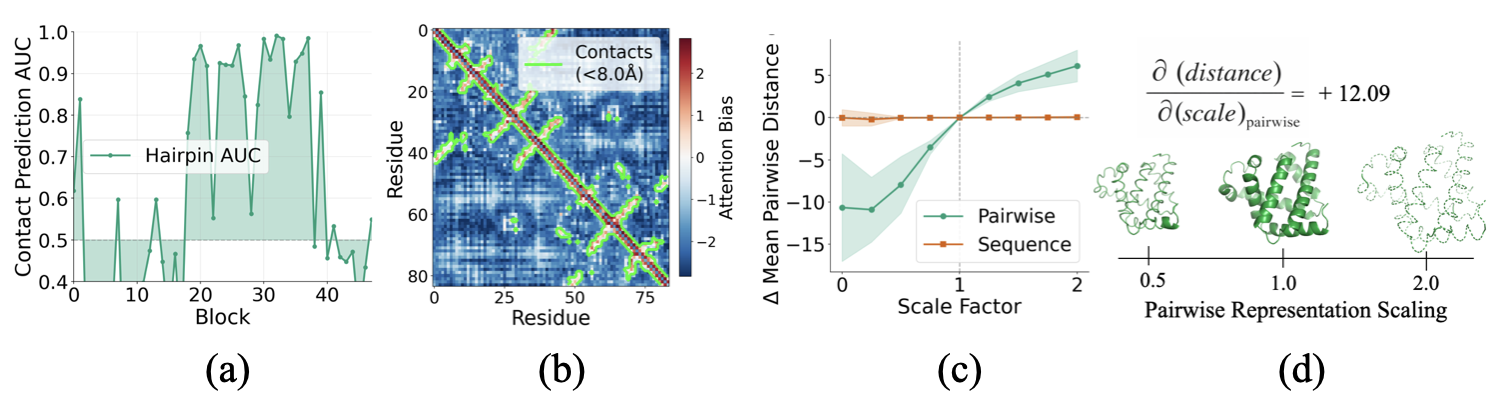}
    \caption{\textbf{Late blocks read $z$ as geometric information (ESMFold).} \textbf{(a)} ROC-AUC for classifying contacting (C$\alpha$ $<$ 8\AA) versus non-contacting residue pairs from \texttt{pair2seq} bias values alone, computed per block. The bias cleanly separates contacts in middle and late blocks. \textbf{(b)} Head-averaged \texttt{pair2seq} bias at block 32 (red = positive, blue = negative); green contours mark residue pairs within 8\AA. Positive bias concentrates at spatial contacts. \textbf{(c)} Effect of scaling the pairwise ($z$) versus sequence ($s$) representation by a factor in $[0, 2]$ before the structure module, shown as deviation in mean pairwise C$\alpha$ distance from baseline (scale = 1.0). \textbf{(d)} Example structures at different pairwise scaling factors, with the gradient $\partial(\text{distance})/\partial(\text{scale})$ at the normal operating point.}
    \label{fig:pair2seq_combined}
\end{figure}

\section{Late Blocks: Pairwise Geometry Emerges}
By the late blocks, the pairwise representation $z$ has taken over as the primary carrier of folding-relevant information: pairwise patching induces structural change while sequence patching does not. We characterize how downstream computation reads $z$ and verify across architectures (App.~\ref{app:triple_model_late_blocks}).

\subsection{Downstream Pathways Read $z$ as Geometric Information}
\label{sec:What_does_z_do}

Two pathways consume $z$ in the ESMFold trunk: the \texttt{pair2seq} bias modulates sequence self-attention (Eq.~\eqref{eq:pair2seq}), and $z$ feeds directly into the structure module that produces 3D coordinates.

\mypar{\texttt{pair2seq} broadcasts contacts to sequence attention} For each block, we extract the \texttt{pair2seq} bias values across all residue pairs and measure how well bias magnitudes separate contacts (C$\alpha$ distance $<$ 8\AA) from non-contacts, quantified by ROC-AUC. In middle and late blocks, the bias cleanly separates contacts from non-contacts (Fig.~\ref{fig:pair2seq_combined}a), and visualizing bias values alongside contact maps shows positive bias concentrated at spatial contacts (Fig.~\ref{fig:pair2seq_combined}b). Thus, late pairwise states make spatial contacts available to sequence attention.

\mypar{The structure module reads $z$ as a geometric signal} To isolate the structure module's use of $z$, we scale $z$ by factors $0$--$2$ before it enters the structure module while holding $s$ fixed, and vice versa, across 600 proteins. Scaling $z$ monotonically scales mean pairwise distance in the output: scaling up expands the predicted structure, scaling down contracts it (Fig.~\ref{fig:pair2seq_combined}c,d).

\mypar{Generalization across architectures} Both pathway behaviors replicate in OpenFold and Boltz-1 (App.~\ref{app:triple_model_late_blocks}, Fig.~\ref{fig:triple_model_late_blocks}). The \texttt{pair2seq} bias (and its Pairformer analog) separates contacts from non-contacts in middle and late blocks, and scaling pairwise representations before the structure module monotonically scales output distances.

\subsection{Pairwise Representations Encode Distance}
\label{sec:distance_steering}

An intuitive hypothesis is that $z$ linearly encodes pairwise distance. We test this by training linear probes to predict pairwise C$\alpha$ distance from $z$ at each block, $\hat{d}_{ij} = \mathbf{w}^\top z_{ij} + b$, where $z_{ij} \in \mathbb{R}^{d_z}$ is the pairwise representation for residues $i$ and $j$. We use 400 proteins for training and 200 for evaluation (App.~\ref{app:experiment_datasets}), training a separate probe per block per model.

\begin{figure*}[t]
    \centering
    \includegraphics[trim=0bp 2.500bp 0bp 4.000bp,clip,width=0.9\linewidth]{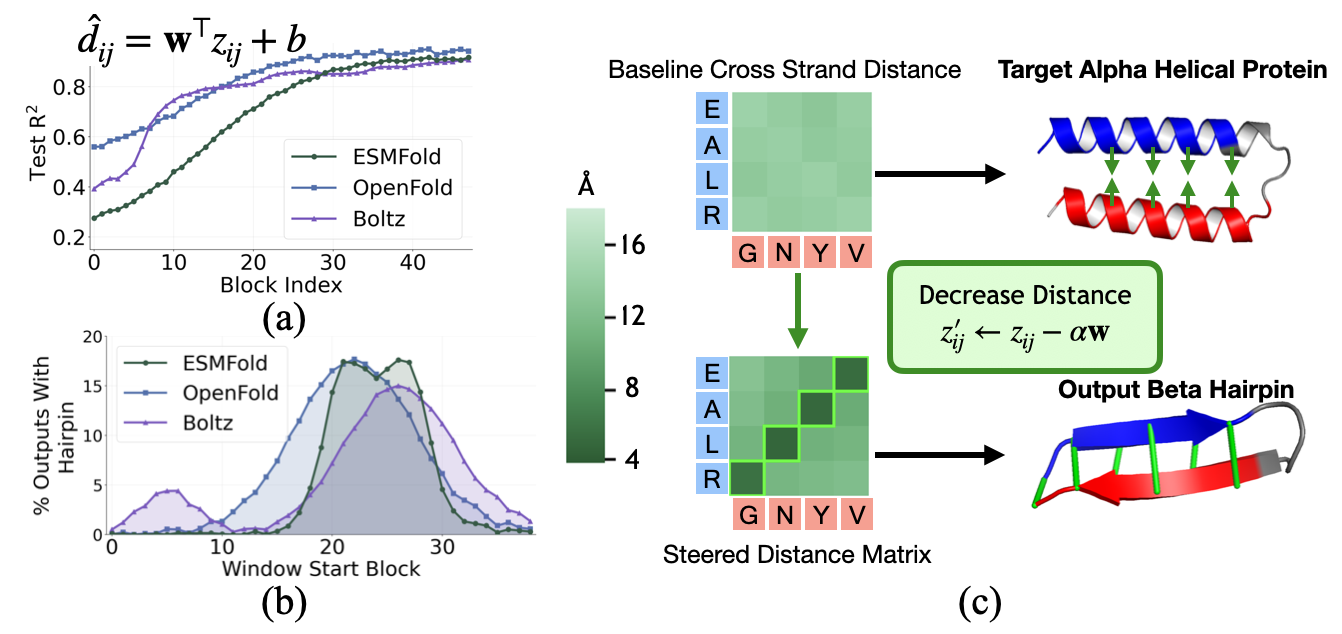}
\caption{\textbf{Distance information is linearly accessible in late blocks and can be manipulated via steering.} (a) $R^2$ score of linear distance probes per block, reaching $\approx 0.9$ at late blocks for all three models. (b) Hairpin induction rate (DSSP) after distance steering with window size 10, for ESMFold (green), OpenFold (blue), and Boltz-1 (purple). Induction peaks when steering middle-to-late blocks. (c) Intervention setup: we move cross-strand pair states along the direction that reduces predicted distance, using 5.5\AA{} as a reference contact distance: $z'_{ij} \leftarrow z_{ij} - \alpha \mathbf{w}$. Successful steering produces a distance matrix with cross-strand residues in contact (dark green), decreasing cross-strand distances and inducing hairpin formation.}
    \label{fig:contact_steering}
\end{figure*}

\mypar{Distance is linearly encoded in late blocks across architectures} Probe accuracy is low at block 0 in all three models, before substantial trunk refinement, but rises through the trunk and plateaus at $R^2 \approx 0.9$ in late blocks (Fig.~\ref{fig:contact_steering}a). Distance is therefore linearly accessible from $z$ in late blocks across all three architectures.

\mypar{Steering $z$ induces hairpin formation across architectures} Since the probe is linear, its gradient with respect to $z_{ij}$ is the weight vector: $\nabla_{z_{ij}} \hat{d}_{ij} = \mathbf{w}$. We can therefore steer the representation toward a target distance by moving along this direction. We choose $5.5$\AA{} as the reference cross-strand contact distance in the schematic. We steer by a fixed displacement along the probe direction. For cross-strand pairs $(i, j) \in \mathcal{C}$ that would form contacts in a hairpin, we steer $z'_{ij} = z_{ij} - \alpha \hat{\mathbf{w}}$, where $\hat{\mathbf{w}}$ is the normalized probe weight direction and $\alpha = 20\sigma$, with $\sigma$ the standard deviation of residue-pair representations projected onto the distance direction. We apply the intervention symmetrically to both $z_{ij}$ and $z_{ji}$ across trunk blocks with window size $10$. Steering produces measurable structural changes in all three models: cross-strand distances decrease and hairpins form in the target region (Fig.~\ref{fig:contact_steering}c). Hairpin induction peaks when steering middle-to-late blocks, where high probe accuracy coincides with the stage 2 effective window.
\section{Folding Trunks Share a Representational Geometry}
\label{sec:cross-model}

The shared stages and causal features suggest that different folding architectures may also develop a common representational geometry. We test this by aligning pairwise states across models, then asking whether those aligned states can drive the receiving model's structural predictions. We illustrate with ESMFold as the receiver; all-pairs results appear in App.~\ref{app:all_pairs_alignment}.

\begin{figure}[t]
    \centering
    \includegraphics[trim=0bp 2.500bp 0bp 7.000bp,clip,width=0.95\linewidth]{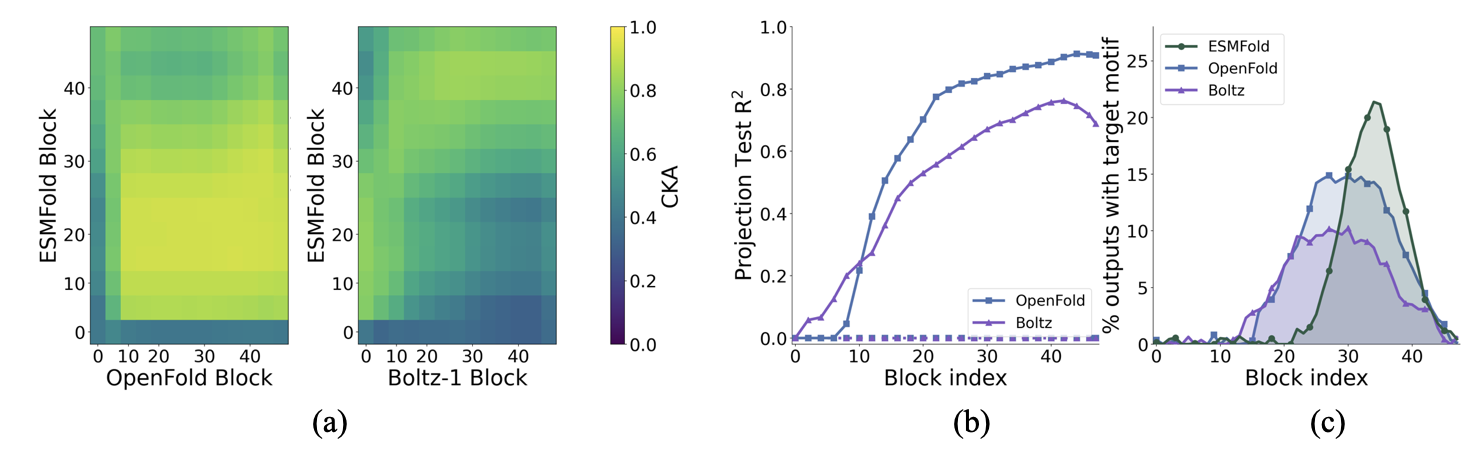}
    \caption{\textbf{Pairwise representations share geometry and support cross-model motif transfer.}
\textbf{(a)} CKA between pairwise representations at every block pair across ESMFold and OpenFold, and ESMFold and Boltz-1.
\textbf{(b)} Test $R^2$ of whitened Procrustes projections from OpenFold and Boltz-1 pairwise representations into ESMFold's space, evaluated per block. Solid lines use true correspondences; dotted lines are shuffled-correspondence baselines.
\textbf{(c)} Single-block pairwise patching success when donor representations come from OpenFold or Boltz-1 (after Procrustes projection into ESMFold's space) versus from ESMFold itself. Cross-model patching reproduces the late-block pairwise window, with strength tracking projection quality from (b).}
    \label{fig:cross-model-patching}
\end{figure}

We first measure representational similarity using Centered Kernel Alignment~\citep{kornblith2019similarity}, a training-free measure of how much two representations share linear structure. Fig.~\ref{fig:cross-model-patching}a shows CKA between every pair of blocks across ESMFold and OpenFold, and across ESMFold and Boltz-1. Similarity is high throughout middle and late blocks (CKA $> 0.8$ for ESMFold vs.\ OpenFold; CKA $> 0.7$ for ESMFold vs.\ Boltz-1) and concentrates along the diagonal: late blocks of one model are most similar to late blocks of the other.

\mypar{Linear alignment captures this similarity} CKA suggests shared structure but does not show whether it is recoverable by a simple map. We test this with whitened Procrustes: for each model pair $(M_A, M_B)$ and block $k$, we whiten paired representations $Z_A, Z_B \in \mathbb{R}^{N \times d}$ as $\tilde{Z}_X = Z_X \Sigma_X^{-1/2}$, then solve $\min_{R^\top R = I} \| \tilde{Z}_A R - \tilde{Z}_B \|_F^2$ for the optimal rotation $R$, giving the full projection $P = \Sigma_A^{-1/2} R \Sigma_B^{1/2}$. Whitening removes scale and covariance differences, leaving rotation to capture geometric correspondence. No folding signal or task supervision is used.

Fig.~\ref{fig:cross-model-patching}b reports per-block test $R^2$. Alignment is poor at the earliest measured block, rises through blocks 5--15, and plateaus thereafter. OpenFold aligns to ESMFold more tightly ($R^2 \approx 0.9$) than Boltz-1 does ($R^2 \approx 0.75$), consistent with their closer architectural relationship. Shuffling the residue-pair correspondences removes this strong alignment, indicating that it depends on matching the same pairs across models. We compare against other projection methods in App.~\ref{app:projection-comparison}.

\mypar{Aligned representations support partial functional transfer} Geometric alignment does not by itself show the representations carry the same \emph{functional} information. We test this by patching projected donor representations into ESMFold's forward pass (single-block protocol of §\ref{sec:patching_setup}).

Cross-model patching reproduces the late-block pairwise window from within-model patching (Fig.~\ref{fig:cross-model-patching}c), with strength tracking alignment quality: ESMFold $\rightarrow$ ESMFold peaks near 22\%, OpenFold $\rightarrow$ ESMFold near 15\%, Boltz-1 $\rightarrow$ ESMFold near 10\%. Cross-model patches induce hairpins only in ESMFold's late blocks, matching the pairwise stage from within-model analysis.

\mypar{Generalization across all model pairs} App.~\ref{app:all_pairs_alignment} repeats these analyses across all three model pairings and both directions of each. The pattern is consistent: high CKA and Procrustes $R^2$ in middle and late blocks, poor held-out alignment under shuffled controls, and cross-model patching reproducing the late-block window in all six directions.

\mypar{Matched calibration and failure analysis} We compare cross-model and within-model patches on the same cases, using the same receiving model, mask, and late-block windows. Cross-model peak success is 20--75\% of the within-model peak. We bootstrap whole cases so that evaluations at different blocks of the same case stay together (App.~\ref{app:revision_transfer}). Final pairwise-state swaps into another model's decoder retain hairpins in 65--78\% of $n=500$ cases per direction, showing that final states can transfer more successfully when the receiving trunk does not process them further. On proteins with divergent predictions, decoded structures resemble the donor more than the host; an ESMFold2 decoder check gives the same qualitative result. These interventions establish functional compatibility across architectures while revealing where transfer depends on the receiving model. The remaining directional asymmetry is discussed in App.~\ref{app:revision_transfer}.

\section{Related Work}

\mypar{Probing and causal intervention} Probes identify information that can be recovered from a representation~\citep{belinkov-2022-probing}. Activation interventions test which states affect model outputs, through causal mediation analysis~\citep{vig2020causal} and causal tracing of factual recall~\citep{meng2022locating}. Steering connects latent directions to changes in generation~\citep{subramani-etal-2022-extracting,marks2024the}. We combine these methods to connect biochemical and geometric features in the trunk to predicted structure.

\mypar{Stages and shared representations} Layer-deletion and layer-swapping experiments motivate stages of inference in language models~\citep{lad2024remarkable}. The Platonic Representation Hypothesis~\citep{huh2024platonicrepresentationhypothesis} connects growing representational alignment across models and modalities to a shared statistical account of their data. We test staged computation and functional transfer of aligned states in folding architectures.

\mypar{Protein model interpretability} InterPLM~\citep{simon2025interplm} and \citet{adams2025from} use sparse autoencoders to identify features in protein language models associated with structural motifs, functional annotations, and other biological properties. \citet{gujral2025sparse} use sparse autoencoders and transcoders to extract features associated with Gene Ontology terms and protein families. \citet{garcia2025interpreting} use such features to steer sequence generation, while \citet{murakami2025mechanistic} study features of an ESM-2 model fine-tuned for nanobody thermostability. Moving beyond sequence encoders, \citet{10.1093/bioadv/vbae187} examine AlphaFold2's structural capabilities through template perturbations and limited sequence information. Our focus is the folding trunk itself: we intervene on its sequence and pairwise states to trace how biochemical information becomes geometry across three architectures.

\mypar{Beta-hairpin folding} Hairpins are a well-studied folding model. Statistical-mechanical models~\citep{munoz1998statistical} and atomistic simulations~\citep{dinner1999betahairpin} investigate their kinetics and stabilizing interactions. We use motif transfer to study structural decisions inside prediction models.

\section{Limitations}
\label{sec:limitations}
\vspace{-10pt}
We test local helices and hairpins, four-strand sheets, sequence-distant contacts, and proteins up to 1,000 residues (App.~\ref{app:revision_scope}). Other topologies, larger proteins, multimers, and pairwise-free architectures such as SimpleFold~\citep{wang2025simplefoldfoldingproteinssimpler} remain open. Interventions can produce clashes, chain breaks, and reduced confidence. Design/refold and cross-model checks support many coherent backbones (App.~\ref{app:revision_quality}), but do not establish experimental stability or that the target sequence adopts the patched fold.

\section{Discussion}
We have conducted a cross-architectural mechanistic analysis of protein folding trunks, revealing a shared two-stage computational structure across the tested structural decisions in ESMFold, OpenFold, and Boltz-1. In the first stage, early computation propagates biochemical information from sequence into pairwise representations; steering a charge-related direction induces hairpin formation. In the second stage, later computation develops geometric features in the pairwise representation; distances are linearly decodable with high accuracy, and steering along the distance direction induces hairpin formation. These features are not merely correlational: interventions show that they influence structural predictions across architectures. Cross-model patching demonstrates that pairwise representations are alignable via linear projection, and representations from one model can partially substitute for another’s at the corresponding stage. The two-stage structure holds for both hairpin and helix induction, suggesting it reflects a recurring computational strategy across the folding trunks studied here.

Protein folding models contain learned structural knowledge that may be valuable beyond their ability to output coordinates. Interpretability has supported scientific discovery in other domains~\citep{davies2021advancing,davies2024signature,wong2024discovery,schut2025bridging,wang2026interpretability}, raising the possibility of extracting biological insights from folding models as well. Our results provide a starting point by connecting biochemical features to the internal computations that shape a predicted structure. Extending this approach could help reveal which residue interactions a model relies on, explain its structural preferences and errors, and turn these explanations into experimentally testable hypotheses about protein behavior.

Conditional protein behavior offers one concrete opportunity. Proteins can adopt different conformations as their biochemical environment changes; for example, specific histidine residues influence the pH sensitivity of influenza hemagglutinin's conformational changes~\citep{mair2014histidine}. Our charge-steering results suggest a way to investigate whether folding models have learned information relevant to such transitions: intervene on biochemical features and examine how the predicted structure responds. Comparing these responses with experimentally measured conformations and mutations could establish whether latent interventions capture meaningful aspects of conditional behavior. This would be a step toward using the knowledge inside folding models to study how proteins respond to changing conditions.

\newpage

\bibliographystyle{plainnat}
\bibliography{references_camera_ready}

\newpage

\appendix

\section{Protein Structure}
\label{app:protein_structure}

\mypar{Chains of amino acids} Proteins are linear chains of amino acids; once incorporated into a protein, each amino acid is called a \textit{residue}. Each residue consists of a backbone (common to all residues) and a side chain (which differs between residue types and drives folding). The backbone contains the amino nitrogen (N), central carbon (C$\alpha$), and carbonyl group (C=O); C$\alpha$--C$\alpha$ distances are commonly used to measure inter-residue distances, while backbone N--H and C=O groups form the hydrogen bonds that stabilize secondary structures. The linear order of residues---the \textit{protein sequence}---determines how they interact when the chain folds into a three-dimensional structure.

\input{Figures/seq_example}

\begin{figure}[h!]
    \centering
    \includegraphics[width=0.95\linewidth]{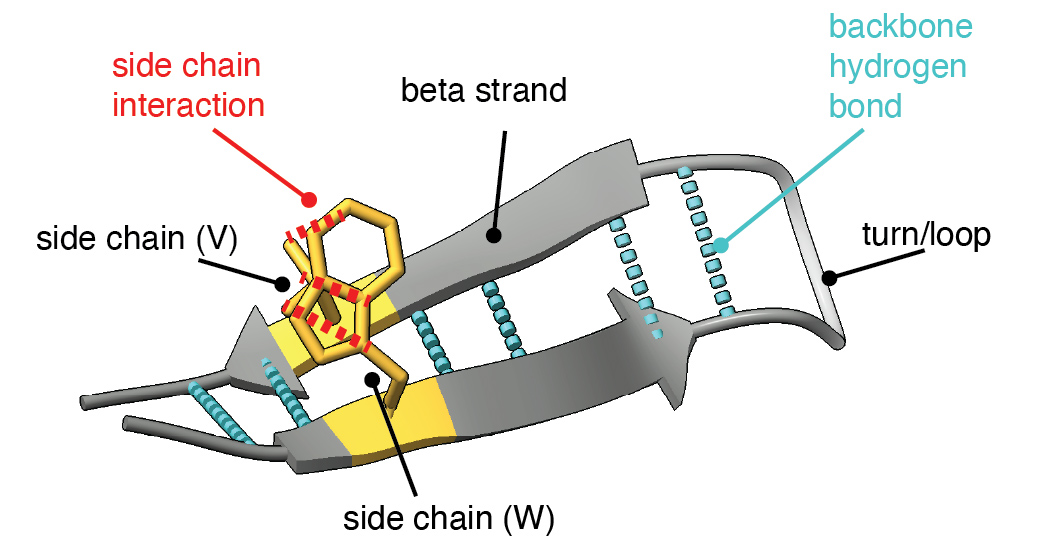}
    \caption{\textbf{3D cartoon diagram of a beta-hairpin}, a common secondary-structure motif consisting of two antiparallel $\beta$-strands (side chain V and side chain W) connected by a short turn/loop. Interactions between side chains on opposing strands and backbone hydrogen bonding together determine the folded hairpin geometry.}
    \label{fig:hairpin-secondary}
\end{figure}

\mypar{From sequence to structure} The amino acid sequence (Fig.~\ref{fig:1d-sequence}) defines the linear order of residues in the chain. During folding, residues far apart in sequence can become neighbors in 3D space.  Protein structure is described at multiple levels: \textit{secondary structure} refers to local motifs such as alpha helices and beta strands (Fig.~\ref{fig:hairpin-secondary}).  We focus on these two motifs as canonical structural decisions a folding model must make. 

\mypar{Beta-hairpins} A beta-hairpin (Fig.~\ref{fig:hairpin-secondary}) is a common secondary-structure motif formed by two beta strands that run in opposite directions along the protein backbone (antiparallel orientation). 
The two strands are connected by a short loop that allows the chain to reverse direction. This loop brings the strands into close proximity, enabling them to align side by side. Once aligned, the strands are stabilized primarily by backbone hydrogen bonds; side-chain interactions provide additional, sequence-dependent stabilization.

Beta-hairpins constitute a non-trivial minimal example of a protein folding motif, and have a long history as a model system in experimental, theoretical, and molecular dynamics studies of folding \citep{munoz1998statistical,blanco1998formation,dinner1999betahairpin}. Simpler features, such as isolated secondary structure elements, do not involve longer-range residue–residue interactions, while larger motifs and full protein folds introduce many interacting components that obscure causal analysis.

\section{Impact Statement}
\label{app:impact_statement}
This work studies interpretability in protein structure prediction models. Improved mechanistic understanding can support safer and more reliable use of these models in basic research and downstream applications, including protein design, by enabling targeted diagnosis of failure modes and more controllable interventions. The methods are analytical and do not introduce new predictive capabilities or training data. Potential negative impacts include dual-use concerns if interpretability insights accelerate protein engineering in harmful contexts; such risks are best mitigated through standard biosecurity governance and careful application review. Overall, we view the primary impact as advancing scientific understanding of how folding models operate.

\section{Code and Computational Cost}
\label{app:code-and-compute}

\paragraph{Code.}
Code, ESMFold demonstration notebooks, and reproduction instructions are available at \url{https://github.com/kevinlu4588/ProteinFolding}. Data and saved results are available at \url{https://huggingface.co/datasets/kevinlu4588/ProteinFolding}.
The repository contains a single \texttt{reproduce.sh} entry point that
runs the key experiments in sequence for ESMFold, along with three Jupyter notebooks
that demonstrate the patching, charge-steering, and distance-steering
interventions on a single example without requiring the full dataset.

\paragraph{Hardware.}
The original experiments were run on a single workstation with
2$\times$ NVIDIA RTX A6000 GPUs (48\,GB VRAM each). Each model
inference fits on a single A6000; the second GPU was used to
parallelize across cases rather than to shard a single forward pass.

\paragraph{Wall-clock budget.}
The dominant cost is repeated forward passes under intervention. We study three models, ESMFold, OpenFold, and Boltz-1, and run the same intervention sweeps on each. Approximate per-model wall-clock times on the setup above are:

\begin{table}[h]
\centering
\small
\begin{tabular}{lr}
\toprule
\textbf{Experiment} & \textbf{Wall-clock (per model)} \\
\midrule
Single-block patching (all trunk blocks)        & $\sim$1.5 days \\
Sliding-window ablations                        & $\sim$1 day    \\
Charge / distance steering sweeps               & $\sim$1 day    \\
Cross-model patching                            & $\sim$1.5 days \\
Module patching, representation tracking, misc. & $\sim$0.5 day  \\
\bottomrule
\end{tabular}
\caption{Approximate wall-clock cost on $2\times$ A6000 per model. Costs scale roughly linearly with the number of cases, which is configurable in \texttt{reproduce.sh}.}
\label{tab:compute}
\end{table}

Aggregated across the three models and including pilot runs and re-runs after iteration, total compute spent on the experiments reported in this paper is approximately \textbf{2--3 weeks of wall-clock time on $2\times$ A6000} (roughly 1{,}000--1{,}500 GPU-hours). These estimates describe the original workstation experiments. The longer-protein extension used two H100 GPUs; a complete compute budget for the added validation experiments is not reported here.

\section{Secondary Structure Detection Algorithm}
\label{app:hairpin_detection_algorithm}
\subsection{Hairpin Detection Algorithm}
A beta hairpin is defined as two antiparallel $\beta$-strands connected by a short loop. We detect hairpins from predicted structures using the following criteria:
\begin{enumerate}
    \item \textbf{Secondary structure assignment}: We run DSSP \citep{Hekkelman2025.04.11.648460} on the predicted structure to identify $\beta$-strand regions (DSSP codes E or B).
    \item \textbf{Strand length}: Each strand must contain at least 2 residues.
    \item \textbf{Loop length}: The connecting loop must be 0--5 residues.
    \item \textbf{Adjacency}: The two strands must be sequential in the primary sequence (no intervening strands).
\end{enumerate}

\subsection{Helix-Loop-Helix Detection Algorithm}
A helix-loop-helix motif is defined as two $\alpha$-helices connected by a short loop. We detect helix-loop-helix motifs from predicted structures using the following criteria:
\begin{enumerate}
    \item \textbf{Secondary structure assignment}: We run DSSP \citep{Hekkelman2025.04.11.648460} on the predicted structure to identify $\alpha$-helical regions (DSSP code H).
    \item \textbf{Helix length}: Each helix must contain at least 4 residues.
    \item \textbf{Loop length}: The connecting loop must be 2--8 residues.
    \item \textbf{Adjacency}: The two helices must be sequential in the primary sequence (no intervening helices).
\end{enumerate}

\section{Licensing of existing assets}
\label{app:licenses}
We use several existing models and public protein-structure resources. ESMFold/ESM is released under the MIT license, OpenFold is released under the Apache 2.0 license, and Boltz-1 is released under the MIT license. For data resources, the Protein Data Bank (PDB) archive is distributed under CC0 1.0, CATH and UniProt are distributed under CC BY 4.0, and DSSP is used for secondary-structure annotation under its published usage terms. We use these resources for research purposes and will include full license and attribution details in the accompanying code and data release.

\section{Experiment Datasets}
\label{app:experiment_datasets}

\textbf{Alpha-helical protein dataset (CATH Class 1).}
To obtain a diverse set of all-$\alpha$ proteins for representation analysis, we constructed a dataset from CATH Class 1 (``mainly $\alpha$'') domains. For each unique Class 1 PDB chain in the CATH domain list, we retrieved the full structure and extracted the complete chain sequence.

To ensure that these proteins were purely helical, we verified secondary structure assignments using DSSP and excluded any chain containing $\beta$-strand residues (DSSP codes E or B). We further filtered sequences to length 90--400 residues and removed duplicate sequences, yielding 5,000 protein chains. From this remaining set, we selected 200 proteins spanning diverse CATH architectures and topologies, including helix bundles, helix--turn--helix motifs, coiled coils, EF-hand proteins, and helical repeat proteins.

This dataset is also used for training the difference-of-means (DoM) vectors that identify biochemical directions (e.g., charge) in sequence representations.

\textbf{Beta Hairpin Dataset}
Protein structures were selected from the PISCES database using a non-redundant culling at 25\% sequence identity, restricting to X-ray structures with resolution between 0.0 and 2.5~\AA\ and R-factor $\leq$ 0.3.
Only protein chains with lengths between 40 and 10,000 residues were retained.
We used the PISCES culling set
\texttt{cullpdb\_pc25.0\_res0.0-2.5\_ len40-10000\_R0.3\_Xray\_ d2025\_02\_19\_chains11652},
which yields 11,652 high-quality protein chains.
Subsequent identification of beta-hairpins was performed on this filtered set as described in App. \ref{app:hairpin_detection_algorithm}.

\textbf{Activation patching target dataset.}
For activation patching experiments, we selected 100 alpha helical proteins and 100 hairpin proteins from the datasets described above. Sequence lengths were restricted to 100--400 residues. These proteins serve as targets into which donor motif representations are patched.

\vspace{0.5em}
\textbf{Probing dataset.}
To evaluate whether biochemical and geometric features are linearly decodable from trunk representations, we built a train/test dataset containing both $\alpha$-helical and $\beta$-hairpin proteins.

The \emph{training set} consists of:
\begin{itemize}
    \item the 200 $\alpha$-helical proteins described above (CATH Class 1), and
    \item 200 full-chain $\beta$-hairpin proteins sampled from a curated motif dataset.
\end{itemize}

The \emph{test set} consists of:
\begin{itemize}
    \item all $\alpha$-helical target proteins used in patching experiments, and
    \item 100 additional full-chain $\beta$-hairpin proteins disjoint from the training hairpins.
\end{itemize}

All sequences were required to contain only the 20 standard amino acids. Hairpin sequences were further filtered to ensure $\leq 25\%$ pairwise sequence identity to any $\alpha$-helical sequence or previously selected hairpin, and to exclude exact matches to donor sequences used in activation patching. The hairpin filters constrain sequence overlap within and between these cohorts.

\section{Effect of Recycling on ESMFold Predictions}
\label{app:recycling}
We evaluated the effect of ESMFold recycling by comparing predictions generated with different numbers of recycles for a randomly sampled subset of UniProt protein sequences. 
For each protein, structures obtained with fewer recycles were compared to the highest-recycle prediction using C$\alpha$ RMSD. 
Short proteins ($<100$ aa) show a rapid decrease in RMSD between 0 and 1 recycle, followed by progressively smaller changes with additional recycling, indicating early structural convergence.
In contrast, medium and long proteins continue to exhibit substantial refinement across multiple recycles, suggesting that recycling is most critical for larger systems and has limited impact for small proteins.

\begin{figure}[h]
    \centering
    \includegraphics[width=0.5\linewidth]{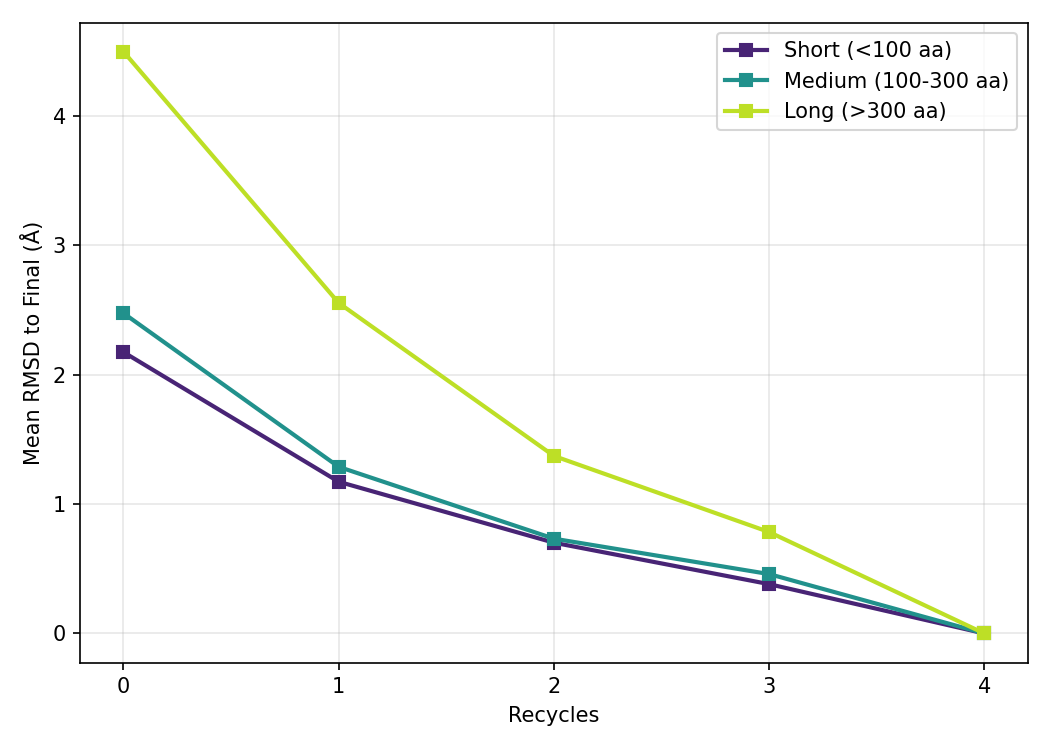}
    \caption{Mean C$\alpha$ RMSD to the highest-recycle ESMFold prediction as a function of the number of recycles, shown for short ($<100$ aa), medium (100–300 aa), and long ($>300$ aa) proteins sampled from UniProt. For short sequences, the RMSD drops sharply between 0 and 1 recycle and decreases only modestly thereafter, indicating that most structural convergence occurs within the first recycle. In contrast, longer proteins continue to show substantial refinement with additional recycling.}
    \label{fig:recycle-analysis}
\end{figure}

\section{Additional Patching Experiments}

\begin{figure}[h]
    \centering
    \includegraphics[width=0.7\linewidth]{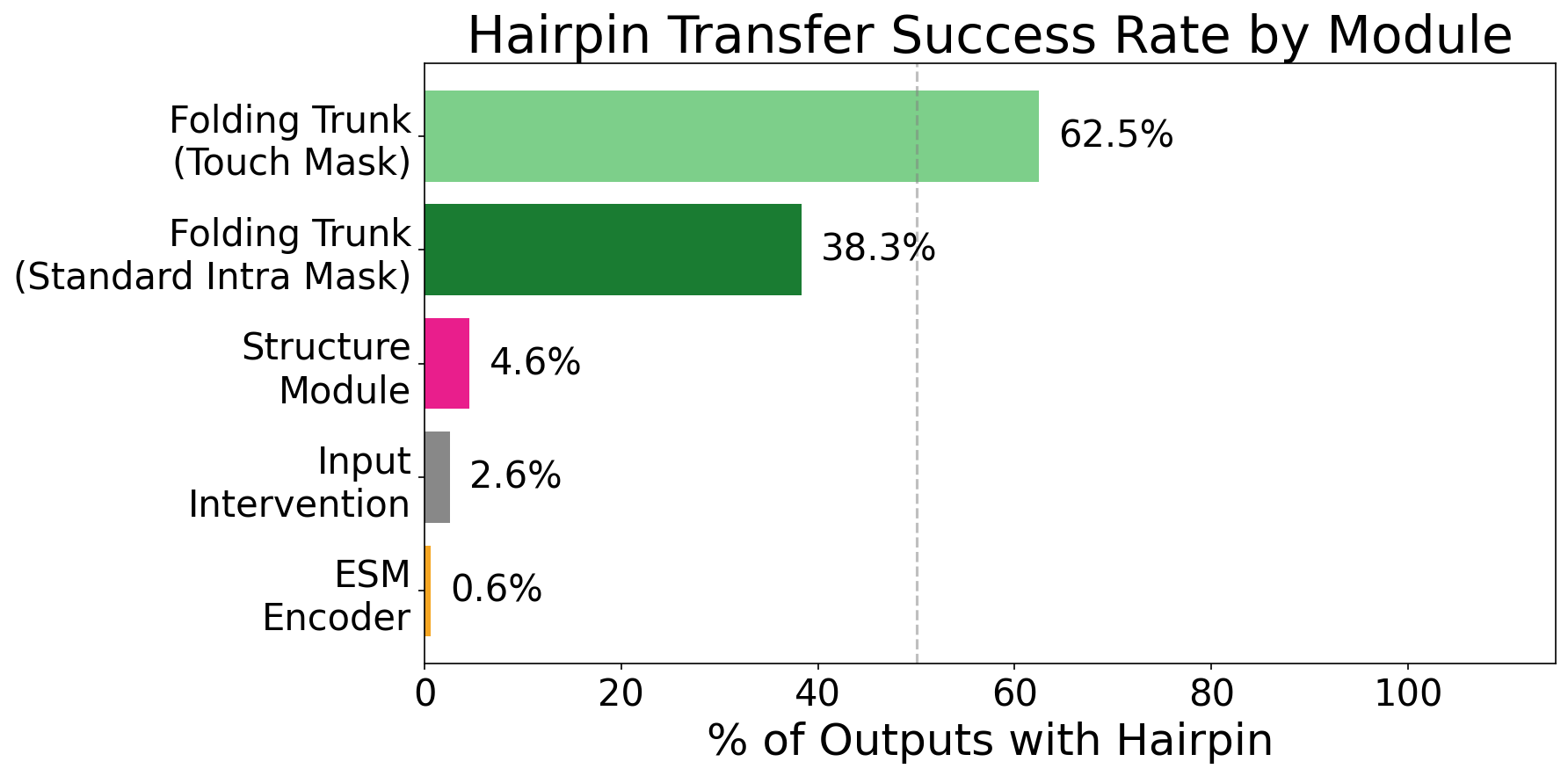}
    \caption{We apply activation patching to the ESM protein language model encoder, folding trunk, and structure module. Patching in the folding trunk is the only place where hairpins consistently transfer. Input intervention: alter the sequence residue text to match hairpin residues before inputting into ESMFold model}
    \label{fig:module_patching}
\end{figure}

\subsection{Performance in Other Modules}
\label{app:other_modules}

To localize where hairpin-relevant computation occurs within ESMFold, we compared activation patching across three architectural components: the ESM-2 encoder, the folding trunk, and the structure module. For each module, we patched donor hairpin representations into the corresponding target representations and measured whether the output structure contained a hairpin.

\mypar{ESM encoder patching} We extracted hidden states from all 33 layers of the ESM-2 encoder for the donor hairpin region and patched them into the corresponding positions during the target's forward pass. Despite replacing the encoder's learned sequence representations, encoder patching rarely induced hairpin formation (Fig.~\ref{fig:module_patching}), suggesting that the encoder representations alone do not determine secondary structure.

\mypar{Structure module patching} We patched the outputs of the invariant point attention (IPA) module at each of its eight iterations. Structure module patching was similarly ineffective: the IPA operates on representations that have already been shaped by the trunk, and patching at this late stage cannot override the geometric information established earlier.

\mypar{Trunk patching} In contrast, patching both sequence and pairwise representations across all 48 blocks of the folding trunk successfully induced hairpin formation in approximately 40\% of cases using standard patching. This makes the trunk a useful site for studying how these motif predictions can be changed, motivating the main-text analysis.

\begin{figure}[h!]
    \centering
    \includegraphics[width=0.5\linewidth]{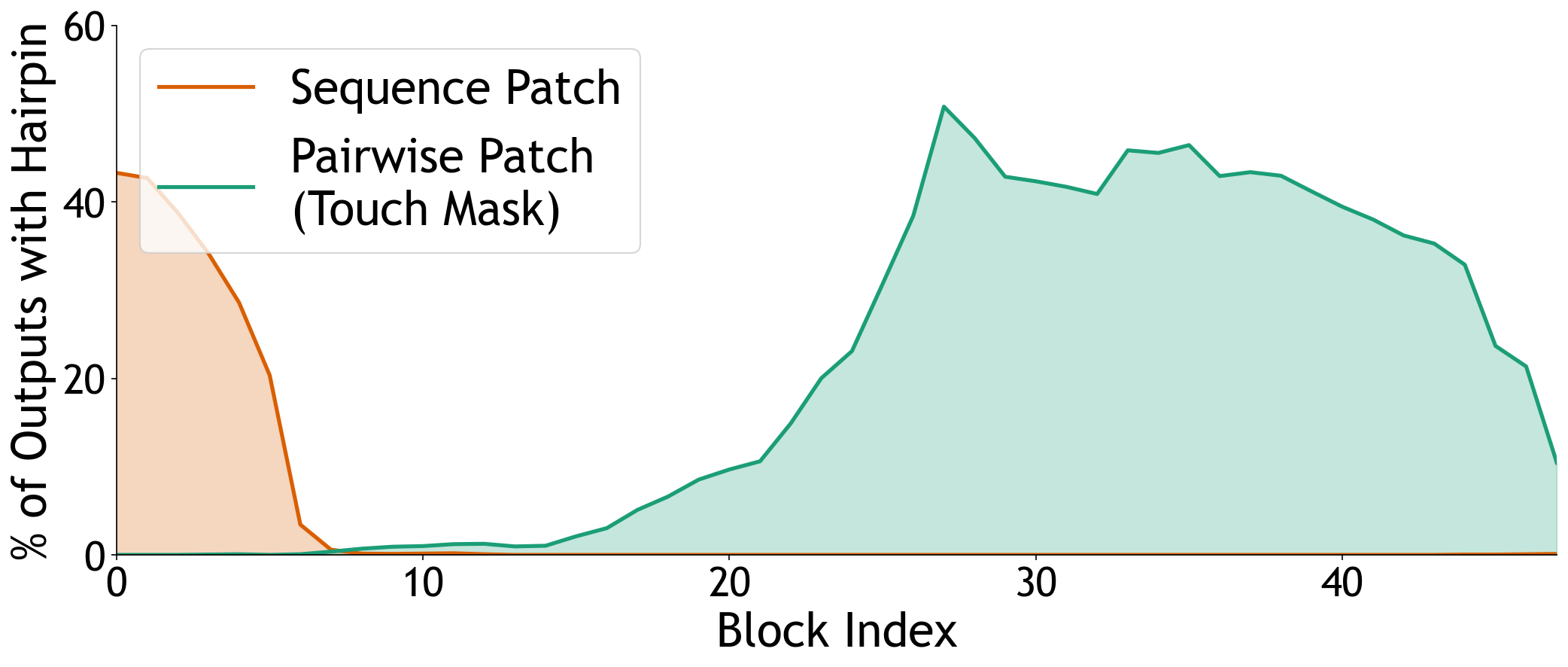}
    \caption{\textbf{Single-block patching with touch mask.} Pairwise patching using the touch mask (green) shows a broader effective window extending into late blocks (25--47), compared to sequence patching (orange) which is effective only in early blocks (0--8). The touch mask patches interactions between hairpin residues and flanking regions, bypassing the need for information to propagate from the patched region to neighboring residues.}
    \label{fig:touch_rate}
\end{figure}

\mypar{Pairwise patching masks} When patching the pairwise representation $z$, we must specify which entries $z_{ij}$ to replace. We experimented with two masking strategies (illustrated in Fig.~\ref{fig:touch_rate}):

\begin{itemize}
    \item \textbf{Standard intra patching}: We patch only entries where both residues $i$ and $j$ fall within the hairpin region, i.e., $z_{ij}$ for $i, j \in [\text{start}, \text{end})$. This captures interactions strictly within the hairpin motif.
    
    \item \textbf{Touch patching}: We patch a cross-shaped region that includes (1) all intra-region entries, plus (2) entries where exactly one residue is in the hairpin region and the other is in the flanking sequence. Formally, we patch $z_{ij}$ whenever $i \in [\text{start}, \text{end})$ or $j \in [\text{start}, \text{end})$. This captures how the hairpin region interacts with its sequential neighbors, which may be important for properly situating the motif within the larger structure.
\end{itemize}

\mypar{Input intervention baseline} As an additional baseline, we performed a literal sequence substitution: replacing the target's amino acid sequence in the patch region with the donor's hairpin sequence before running ESMFold. This provides a direct comparison between replacing the input sequence and replacing internal states. As shown in Fig.~\ref{fig:module_patching}, input intervention induces hairpins at a lower rate than trunk patching, indicating that the learned representations in the trunk carry information beyond what is present in the raw sequence.

\subsection{Touch Patching Extends the Effective Window for Pairwise Interventions}
\label{app:touch_patching}

In the main text (Fig.~\ref{fig:stages_of_computation}), we report pairwise patching results using standard (intra-region) masking, which patches only entries $z_{ij}$ where both $i$ and $j$ fall within the hairpin region. Here we compare this to touch patching, which additionally patches entries where the hairpin region interacts with flanking residues.

Touch patching yields a broader effective window that extends further into late blocks (Fig.~\ref{fig:touch_rate}). We attribute this to information propagation constraints. With standard patching, donor information is injected only into the intra-hairpin entries of $z$; this information must then propagate outward through subsequent blocks to update how the hairpin interacts with flanking residues. If we patch too late, insufficient blocks remain for this propagation to occur, and the resulting structure is incoherent---the hairpin region conflicts with its surroundings.

Touch patching bypasses this bottleneck by directly providing the hairpin-to-flanking interactions, eliminating the need for propagation. This allows effective patching even in very late blocks where standard patching fails. The result highlights that successful structural transfer requires not only the correct intra-motif geometry but also sufficient time (i.e., remaining blocks) for the model to integrate the patched region with the rest of the chain.

\section{Generalization of Information Flow Across Architectures}
\label{app:triple_model_info_flow}

\begin{figure}[h]
    \centering
    \includegraphics[width=\linewidth]{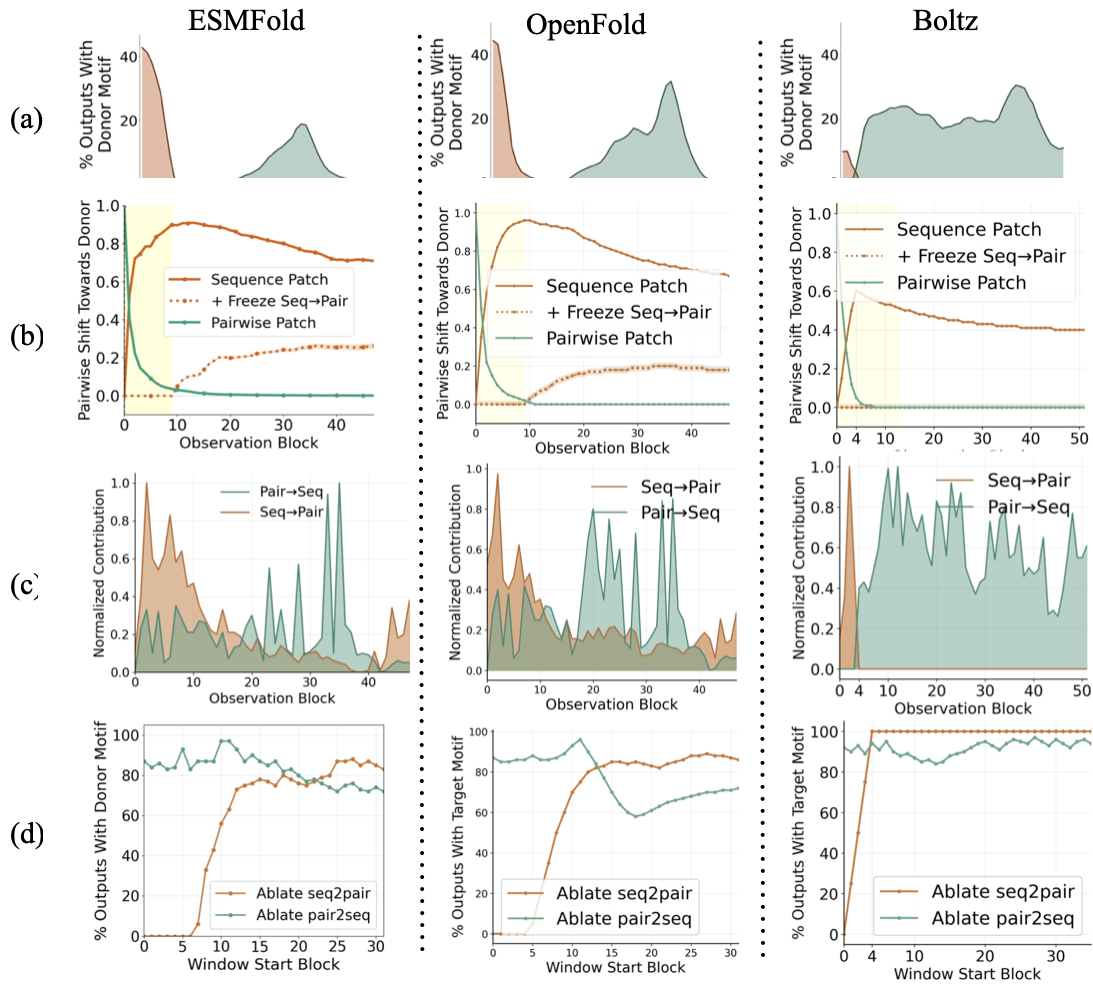}
    \caption{\textbf{Information flow analysis across ESMFold, OpenFold, and Boltz-1.}
    \textbf{(a)} Single-block patching success rate as a function of block index. Sequence patches (orange) are effective in early blocks across all three models; pairwise patches (green) are effective in late blocks. Boltz-1 additionally shows a broad pairwise window across the whole trunk.
    \textbf{(b)} Shift in $z$ during the target forward pass under sequence patching at block 0, sequence patching with \texttt{seq2pair} frozen during the early window, and pairwise patching at block 0. All three models show a clear early write-in window (yellow) during which sequence patches shift $z$ toward the donor and freezing \texttt{seq2pair} prevents this shift.
    \textbf{(c)} Normalized contribution magnitudes of \texttt{seq2pair} (orange) and \texttt{pair2seq} (green) at each block. ESMFold and OpenFold show earlier \texttt{seq2pair} and later \texttt{pair2seq} peaks. In Boltz-1, the \texttt{seq2pair} contribution is concentrated in the upstream MSA module (blocks 0--3) and essentially zero across the rest of the Pairformer trunk.
    \textbf{(d)} Sliding-window ablation (window size 15) during block-0 sequence patching. In all three models, ablating \texttt{seq2pair} during early blocks reduces motif formation, with full recovery once the window starts past the early write-in phase. Ablating \texttt{pair2seq} has little effect at any window position.}
    \label{fig:triple_model_info_flow}
\end{figure}

In §\ref{sec:information_flow}, we showed that early ESMFold blocks transfer biochemical information from sequence into pairwise representations through the \texttt{seq2pair} pathway, and that this transfer underlies the early-block sequence patching window. Here we replicate the four analyses underlying that conclusion, single-block patching, representation similarity under interventions, pathway contributions, and sliding-window ablations, across all three folding architectures. Results are summarized in Fig.~\ref{fig:triple_model_info_flow}.

\paragraph{Two-stage patching pattern (Fig.~\ref{fig:triple_model_info_flow}a).} The single-block patching analysis from §\ref{sec:patching_setup} produces qualitatively the same pattern in all three models: sequence interventions act earlier than sustained pairwise interventions. ESMFold and OpenFold show stronger sequence-patching success; Boltz-1 sequence success is 15\% before filtering and 7\% after designability filtering. Boltz-1 has a broader filtered pairwise window, blocks 3--47 (App.~\ref{app:revision_quality}). One possible contributor is the absence of \texttt{seq2pair} write-back in Boltz-1's downstream Pairformer: once $z$ enters the Pairformer, it is refined within the pairwise track, without further sequence-to-pair updates. This may make donor pairwise states harder to overwrite. In ESMFold and OpenFold, by contrast, early-block pairwise patches are rewritten by subsequent \texttt{seq2pair} updates, narrowing the effective window to late blocks.

\paragraph{Representation similarity under interventions (Fig.~\ref{fig:triple_model_info_flow}b).} We replicate the interpolation-coefficient analysis (Eq.~\ref{eq:representation_similarity}) on OpenFold and Boltz-1. In all three models, sequence patching at block 0 shifts $z$ toward the donor within the first $\approx$10 blocks, while freezing \texttt{seq2pair} during this window prevents the shift entirely; direct pairwise patches at block 0 do not persist. The early write-in window (highlighted in yellow) is therefore present in all three architectures. In Boltz-1, the relevant \texttt{seq2pair} pathway lives in the upstream MSA module (blocks 0--3) rather than throughout the trunk; the freeze condition correspondingly targets these blocks.

\paragraph{Pathway contributions (Fig.~\ref{fig:triple_model_info_flow}c).} Without intervention, the relative magnitudes of \texttt{seq2pair} and \texttt{pair2seq} updates follow the staged pattern in all three models: \texttt{seq2pair} peaks during early blocks and declines, while \texttt{pair2seq} is small early and rises in middle and late blocks. The location of the \texttt{seq2pair} peak shifts to track each model's sequence-to-pairwise pathway: early throughout the trunk in ESMFold and OpenFold, confined to blocks 0--3 in Boltz-1 (corresponding to its upstream MSA module).

\paragraph{Sliding-window ablations (Fig.~\ref{fig:triple_model_info_flow}d).} In all three models, ablating \texttt{seq2pair} during a window starting in early blocks reduces motif formation, with recovery once the window starts past the early write-in phase (block 12 in ESMFold and OpenFold, block 4 in Boltz-1, again tracking the location of each model's \texttt{seq2pair} pathway). Ablating \texttt{pair2seq} has little effect at any window position in any model, ruling out generic disruption from removing either inter-representation pathway. Together with the patching, similarity, and contribution analyses, these results support the interpretation that stage 1, the biochemical write-in from sequence into pairwise space, is a shared property of all three folding trunks, with its location varying based on where each architecture implements its \texttt{seq2pair} pathway.

\section{Additional Charge Experiments}

\subsection{Charge Direction Separability}
\label{app:charge_doms}

To assess the linear decodability of charge information from each model's sequence representation, we computed a charge direction using the difference-of-means approach (Eq.~\ref{eq:dom}) and measured how well projections onto this direction separate positively charged residues (K, R, H) from negatively charged residues (D, E). For each block in each model, we report ROC-AUC of the projection as a binary classifier.

\begin{figure}[h]
    \centering
    \includegraphics[width=0.7\linewidth]{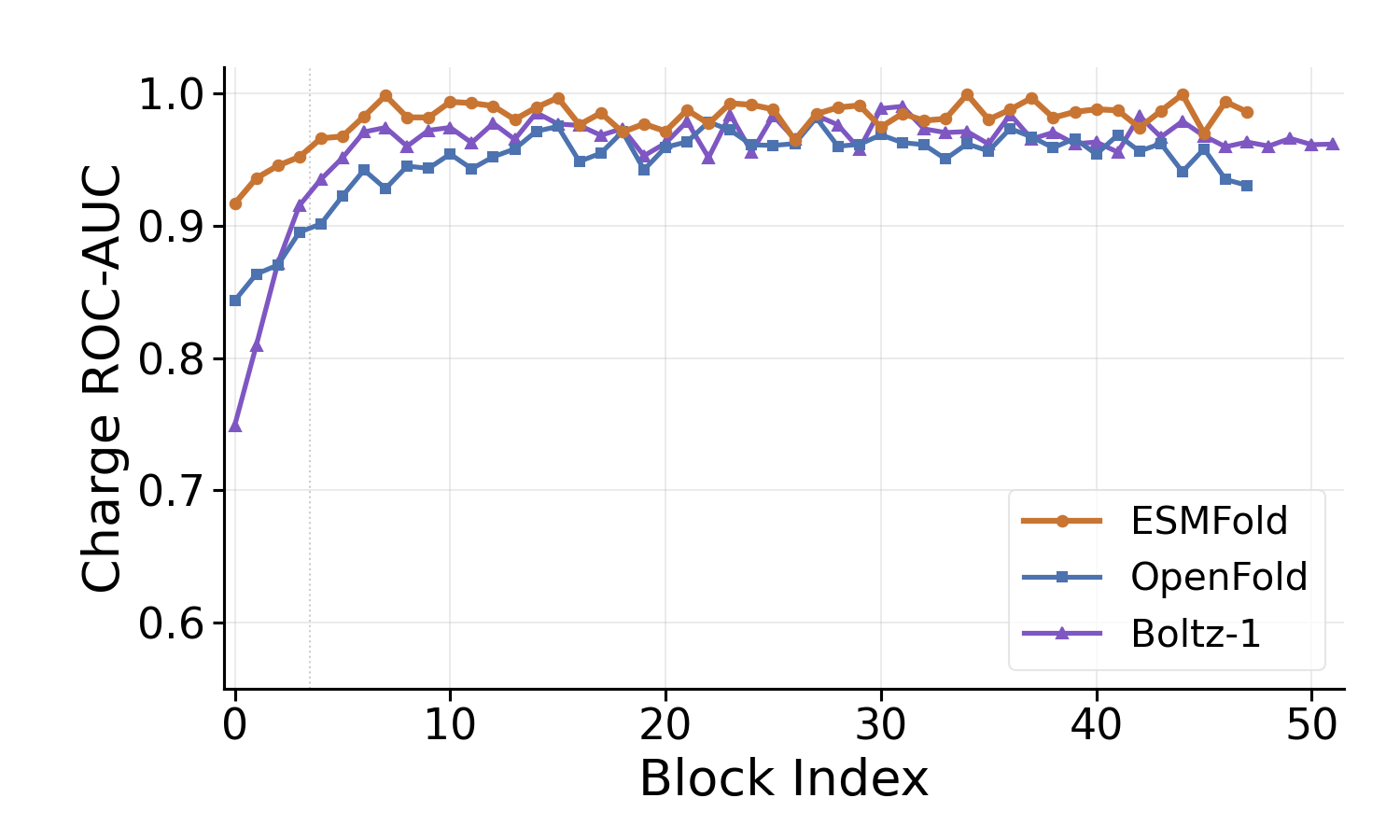}
    \caption{\textbf{Charge information is linearly decodable from sequence representations across all three models.} ROC-AUC of the difference-of-means charge direction $v_{\text{charge}}$ as a binary classifier separating positively charged residues (K, R, H) from negatively charged residues (D, E), computed per block. All three models reach AUC near 1.0 within the first few blocks and maintain it throughout the trunk.}
    \label{fig:charge_doms}
\end{figure}

Fig.~\ref{fig:charge_doms} shows that projections onto $v_{\text{charge}}$ distinguish the charge groups throughout most of the trunk in all three models. ESMFold and OpenFold start at AUC $\approx 0.85$--$0.9$ at block 0 and rise to near 1.0 within the first 5--10 blocks. Boltz-1's curve begins lower at the earliest MSA block (AUC $\approx 0.75$) and rises sharply through the MSA module to AUC $\approx 0.95$ by block 0 of the Pairformer, after which it remains high.

These results show that charge-related information remains linearly decodable throughout most of the trunk in each model. This property enables our charge steering intervention (§\ref{sec:biochem-features}): the difference-of-means construction yields a meaningful charge direction in every architecture, supporting the cross-architectural steering results in Fig.~\ref{fig:charge_boom}b.

\subsection{Charge Probing}
\label{app:charge_probing}

To verify that charge information propagates from sequence representations into the pairwise representation via \texttt{seq2pair}, we trained linear probes to predict residue charge directly from pairwise features. We extend this analysis across all three models to test whether the architectural variation identified in App.~\ref{app:triple_model_info_flow} (specifically, that Boltz-1's seq2pair pathway lives in the upstream MSA module rather than the trunk) shows up at the level of a specific biochemical feature.

\mypar{Probe setup} For each block $k$ in each model, we train two binary linear classifiers on the pairwise representation $z^{(k)}_{ij}$: one to detect positively charged residues (K, R, H) and one to detect negatively charged residues (D, E). Given a pair of residues $(i, j)$ with sufficient sequence separation ($|i - j| \geq 4$), the probe predicts the charge class of residue $i$ from the pairwise feature vector $z_{ij}$. We train analogous probes on each model's \texttt{seq2pair} output (or its architectural analog) to track when charge information first enters the pairwise pathway. We train probes on the probing dataset described in App.~\ref{app:experiment_datasets}, reporting balanced accuracy to account for class imbalance (charged residues comprise approximately 15\% of natural proteins).

\begin{figure}[h]
    \centering
    \includegraphics[width=1.0\linewidth]{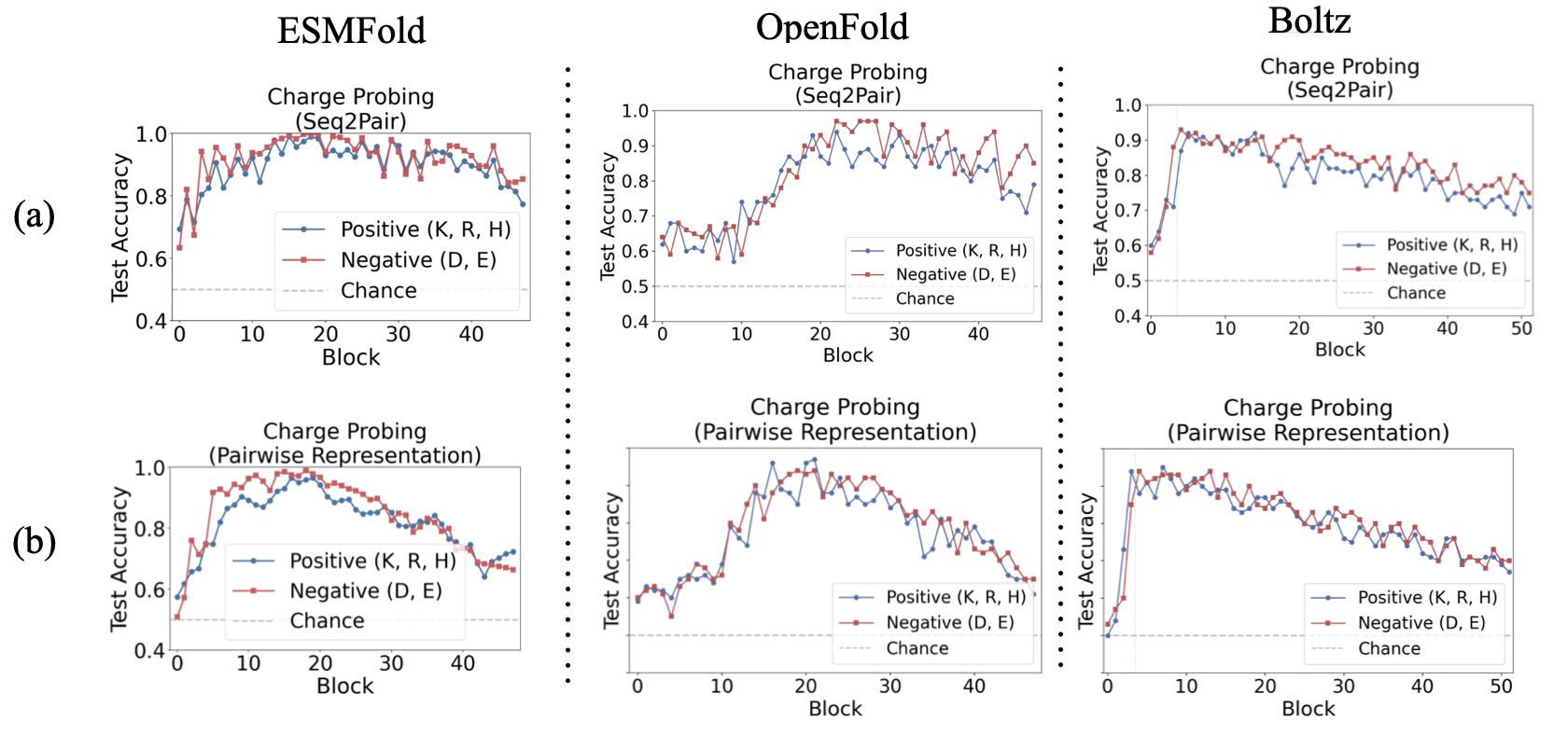}
    \caption{\textbf{Charge information is linearly accessible in pairwise representations across all three architectures.} Balanced accuracy of linear probes predicting positive charge (K, R, H; blue) and negative charge (D, E; red) from \textbf{(a)} the \texttt{seq2pair} output and \textbf{(b)} the pairwise representation $z$. Probes predict the charge of residue $i$ from pairwise features at position $(i, j)$ for sequence-distant pairs ($|i - j| \geq 4$). \texttt{seq2pair} output maintains high accuracy from early blocks throughout the trunk, while $z$ probes start near chance at block 0 and rise through blocks 0--15, consistent with charge information transferring into $z$ via \texttt{seq2pair} during early trunk blocks. The location of the rise tracks each model's sequence-to-pairwise pathway.}
    \label{fig:charge_probing}
\end{figure}

\mypar{Results} Figure~\ref{fig:charge_probing} shows probe accuracy across blocks for all three models.

In all three models, probes on $z$ start near chance at block 0 (before substantial pairwise refinement) and rise through the early blocks, plateauing at accuracies comparable to the corresponding \texttt{seq2pair} probes. The location of the rise tracks each model's \texttt{seq2pair} pathway: blocks 0--15 in ESMFold and OpenFold, and the narrower window of blocks 0--3 in Boltz-1, where the \texttt{seq2pair} pathway lives in the MSA module before the Pairformer trunk begins. This timing aligns with our patching results in §\ref{sec:information_flow}: early trunk blocks constitute a ``write-in'' window during which biochemical features including charge are transferred from sequence representations into $z$ via \texttt{seq2pair}.

In all three models, $z$ probe accuracy declines somewhat in late blocks (35--47), suggesting charge information is progressively transformed into more geometric features as the trunk shifts to stage 2 computation. This is consistent with our finding that pairwise representations encode distance with high accuracy in late blocks (§\ref{sec:distance_steering}).

\mypar{Summary} Charge probing supports the pathway analysis in App.~\ref{app:triple_model_info_flow}: charge information enters $z$ during the early-block write-in window in all three models, with the location of the window tracking where each architecture implements its \texttt{seq2pair} pathway.

\subsection{Charge Steering Dose-Response}
\label{app:charge_size_scaling}

We measure how cross-strand distance changes as a function of charge steering magnitude across all three models. To isolate the dose-response relationship from confounding structural variation, we use small single-motif proteins for this experiment: short ($\leq$ 40 residue) PDB proteins consisting primarily of a beta-hairpin (for same-charge steering) or a helix-turn-helix (for opposite-charge steering), rather than the larger, more diverse proteins used elsewhere in the paper. This lets us measure the geometric effect of charge steering without variability from surrounding structural context.

For each model, we apply four steering conditions: same-charge steering on beta-hairpin targets (Both +, Both $-$), expecting strands to repel; and opposite-charge steering on $\alpha$-helical targets (+/$-$, $-$/+), expecting strands to attract. Steering magnitude is reported in units of the standard deviation of residue projections onto the model's charge direction.

\begin{figure}[h]
    \centering
    \includegraphics[width=1.0\linewidth]{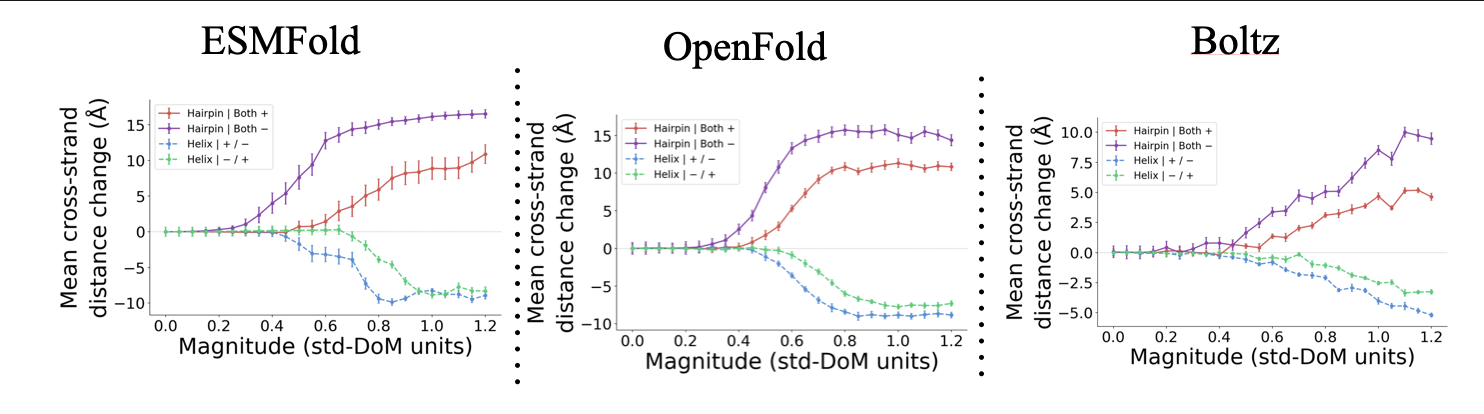}
    \caption{\textbf{Charge steering produces a smooth, monotonic dose-response across all three models.} Mean cross-strand distance change as a function of steering magnitude (in units of std-DoM) on small single-motif target proteins ($\leq$ 40 residues). Same-charge steering on hairpin targets (red, purple) increases cross-strand distance; opposite-charge steering on helical targets (blue, green) decreases it. All three models show effects emerging above a threshold around 0.4 std-DoM units and saturating thereafter, showing graded sensitivity to this charge-related latent direction in these structural contexts. Effect magnitudes are largest in OpenFold, intermediate in ESMFold, and smallest in Boltz-1, consistent with the location of the sequence-to-pair pathway in App.~\ref{app:triple_model_info_flow}: Boltz-1's seq2pair pathway lives upstream of the trunk, so trunk-level steering has a smaller direct effect.}
    \label{fig:charge_size_scaling}
\end{figure}

\begin{figure}[h]
    \centering
    \begin{tikzpicture}
        \def\boxwidth{0.22\linewidth}
        \def\boxheight{3.2cm}
        \def\boxsep{0.01\linewidth}
        
        \node[anchor=south west, inner sep=0, minimum width=\boxwidth, minimum height=\boxheight] (img1) at (0, 0)
            {\includegraphics[width=\boxwidth, height=\boxheight, keepaspectratio]{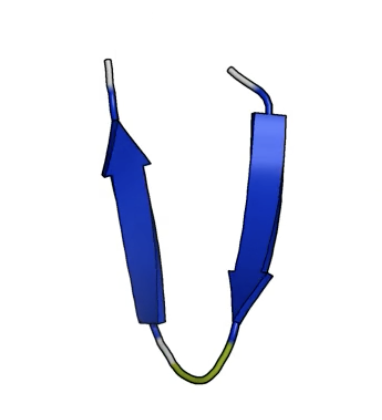}};
        \node[anchor=south west, inner sep=0, minimum width=\boxwidth, minimum height=\boxheight] (img2) at ($(img1.south east) + (\boxsep, 0)$)
            {\includegraphics[width=\boxwidth, height=\boxheight, keepaspectratio]{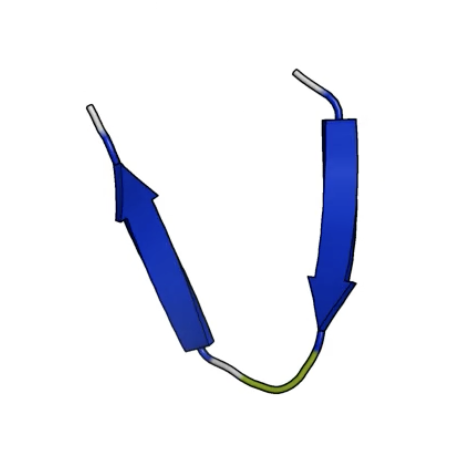}};
        \node[anchor=south west, inner sep=0, minimum width=\boxwidth, minimum height=\boxheight] (img3) at ($(img2.south east) + (\boxsep, 0)$)
            {\includegraphics[width=\boxwidth, height=\boxheight, keepaspectratio]{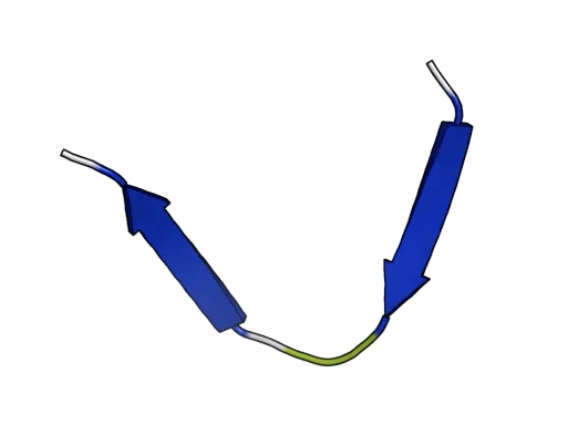}};
        \node[anchor=south west, inner sep=0, minimum width=\boxwidth, minimum height=\boxheight] (img4) at ($(img3.south east) + (\boxsep, 0)$)
            {\includegraphics[width=\boxwidth, height=\boxheight, keepaspectratio]{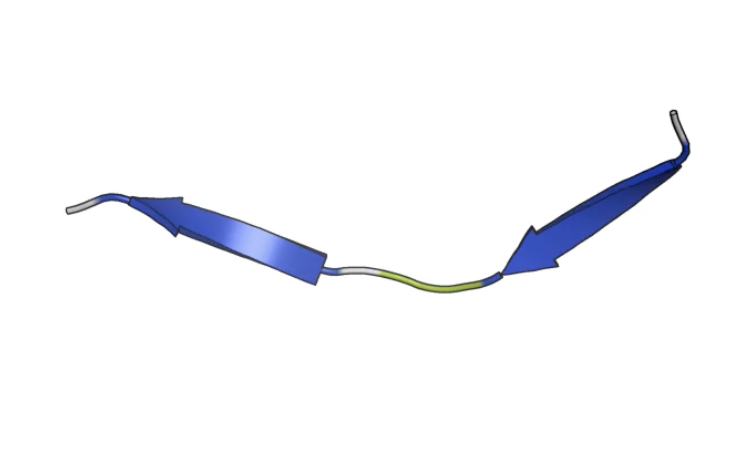}};
        
        \coordinate (arrowL) at (img1.north west |- img1.north);
        \coordinate (arrowR) at (img4.north east |- img1.north);
        
        \draw[->, line width=2.5pt, draw=red!80!black]
            ($(arrowL) + (0, 0.5)$) -- ($(arrowR) + (0, 0.5)$)
            node[midway, above=2pt, black] {\textbf{Increasing same-charge steering magnitude}};
        
        \node[font=\small\bfseries] at (img1.north) {(a)};
        \node[font=\small\bfseries] at (img2.north) {(b)};
        \node[font=\small\bfseries] at (img3.north) {(c)};
        \node[font=\small\bfseries] at (img4.north) {(d)};
        
        \node[font=\small, below=2pt of img1.south] {0.2 std-DoM};
        \node[font=\small, below=2pt of img2.south] {0.4 std-DoM};
        \node[font=\small, below=2pt of img3.south] {0.6 std-DoM};
        \node[font=\small, below=2pt of img4.south] {0.8 std-DoM};
        
    \end{tikzpicture}
    \caption{\textbf{Progressive separation of beta-hairpin strands under increasing same-charge steering (Pos-Pos).} Predicted structures of a small beta-hairpin target protein under same-charge steering at four magnitudes (0.2, 0.4, 0.6, 0.8 std-DoM). At low magnitude (a), the hairpin retains its native folded geometry with strands in close cross-strand contact. As steering magnitude increases (b--d), the strands progressively separate under the latent intervention, producing the dose-response curve quantified in Fig.~\ref{fig:charge_size_scaling}.}
    \label{fig:hairpin_separation_frames}
\end{figure}

The same dose-response shape appears in all three models: a near-zero region below threshold, a monotonic rise (or fall) through intermediate magnitudes, and saturation at high magnitudes. The threshold and saturation magnitudes are similar across models. The smaller effect size in Boltz-1 is consistent with the location of the sequence-to-pair pathway in App.~\ref{app:triple_model_info_flow}: charge written into the sequence track at trunk blocks 0--3 has limited downstream influence on the Pairformer's pairwise representation, since Boltz-1 lacks a seq2pair write-back within the trunk. Fig. \ref{fig:hairpin_separation_frames} visualizes this gradual separation of beta hairpin strands via the charge steering. 

\section{Generalization of Late-Block Geometric Computation Across Architectures}
\label{app:triple_model_late_blocks}

\begin{figure}[h]
    \centering
    \includegraphics[width=\linewidth]{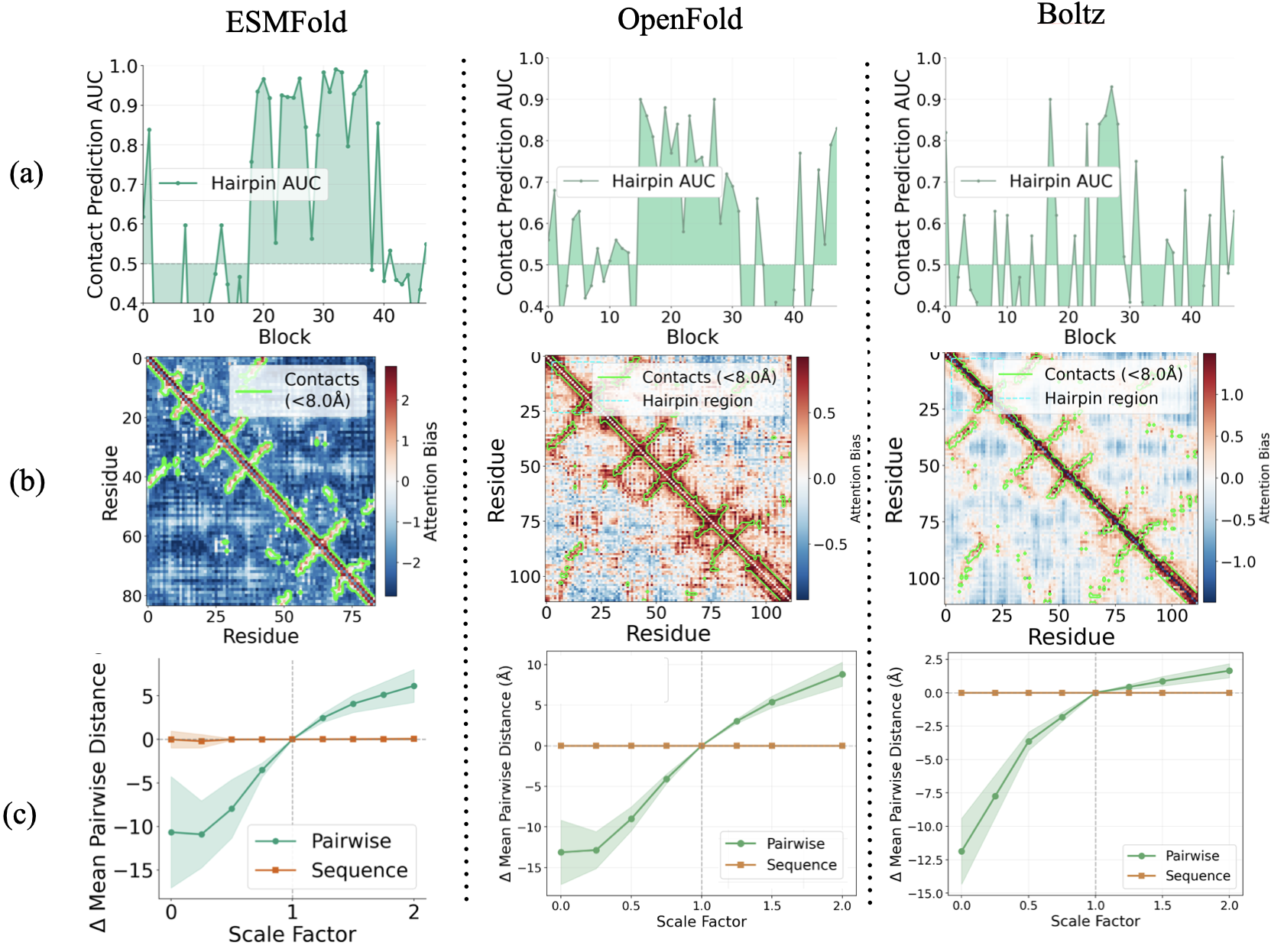}
    \caption{\textbf{Late-block pathway analysis across ESMFold, OpenFold, and Boltz-1.}
    \textbf{(a)} ROC-AUC for classifying contacting (C$\alpha$ $<$ 8\AA) versus non-contacting residue pairs from \texttt{pair2seq} bias values alone, computed per block. All three models exhibit a middle-to-late block window in which the bias cleanly separates contacts from non-contacts.
    \textbf{(b)} Head-averaged \texttt{pair2seq} bias for an example protein at a representative late block. Heatmap entries are bias values (red = positive, blue = negative); green contours mark residue pairs within 8\AA. Positive bias concentrates at spatial contacts in all three models.
    \textbf{(c)} Effect of scaling the pairwise ($z$, green) or sequence ($s$, orange) representation by a factor in $[0, 2]$ before it enters the structure module. Mean pairwise C$\alpha$ distance is shown as deviation from the unscaled baseline (scale = 1.0); shaded regions indicate one standard deviation across the test set. Scaling $z$ monotonically scales output distances in all three models; scaling $s$ has virtually no effect.}
    \label{fig:triple_model_late_blocks}
\end{figure}

In §\ref{sec:What_does_z_do}, we showed that two pathways read $z$ as geometric information in ESMFold's late trunk: contact information can be read from the \texttt{pair2seq} attention bias, and scaling $z$ changes distances in the decoded structure. Here we replicate both analyses on OpenFold and Boltz-1. Results are summarized in Fig.~\ref{fig:triple_model_late_blocks}.

\paragraph{\texttt{pair2seq} makes contacts accessible to sequence attention (Fig.~\ref{fig:triple_model_late_blocks}a,b).} For each block in each model, we extract the bias term that modulates sequence attention from the pairwise representation and measure how well its magnitude separates contacting from non-contacting residue pairs (C$\alpha$ distance $<$ 8\AA), reporting ROC-AUC. All three models show a clear middle-to-late block window in which the bias cleanly separates contacts. ESMFold's separation is the sharpest, with AUC approaching 1.0 in late blocks; OpenFold and Boltz-1 reach AUC $\approx$ 0.9 over a similar block range. Visualizing head-averaged bias values alongside contact maps for a representative protein (panel b) confirms that positive bias concentrates at spatial contacts in all three models, with the off-diagonal contact structure clearly visible. Attention has access to this contact information in all three models, although these readouts do not measure its contribution to the final structure.

\paragraph{The structure module reads $z$ as a geometric signal (Fig.~\ref{fig:triple_model_late_blocks}c).} We replicate the scaling analysis from §\ref{sec:What_does_z_do} on OpenFold and Boltz-1: we multiply either the pairwise representation $z$ or the sequence representation $s$ by a factor in $[0, 2]$ immediately before the structure module, holding the other fixed, and measure the change in mean pairwise C$\alpha$ distance. All three models exhibit the same qualitative behavior: scaling $z$ produces a monotonic, approximately linear change in output distance, while scaling $s$ produces no detectable change. The magnitude of the effect varies between models. ESMFold shows an intermediate effect, with mean pairwise distance changing by roughly 15\AA{} across the scaling range. OpenFold shows the largest swing (about 20\AA{} swing) and Boltz-1 the smallest (about 13\AA{} swing, but with most of the change concentrated below scale = 1). The direction and qualitative shape are preserved across architectures: this protocol detects greater sensitivity to scaling $z$ than $s$. It does not exclude geometric information or causal contributions in $s$.

\paragraph{Summary.} Together, these analyses establish that the stage-2 picture from ESMFold generalizes across all three folding trunks. In all three models, contact information is present in the attention bias, and scaling the pairwise input changes output distances.

\subsection{Bias Maps}
\label{app:bias_maps}

We visualize the \texttt{pair2seq} attention bias (Section~\ref{sec:What_does_z_do}) across ESMFold blocks and across individual attention heads. For a single protein (PDB: 6rwc), we extract the bias term $\beta_{ij}(z_{ij})$ from Eq.~\eqref{eq:pair2seq} and average over heads. Green contours indicate structural contacts (C$\alpha$ distance $<$ 8\AA).

\begin{figure*}[H]
    \centering

    \begin{subfigure}[b]{0.24\textwidth}
        \includegraphics[width=\textwidth]{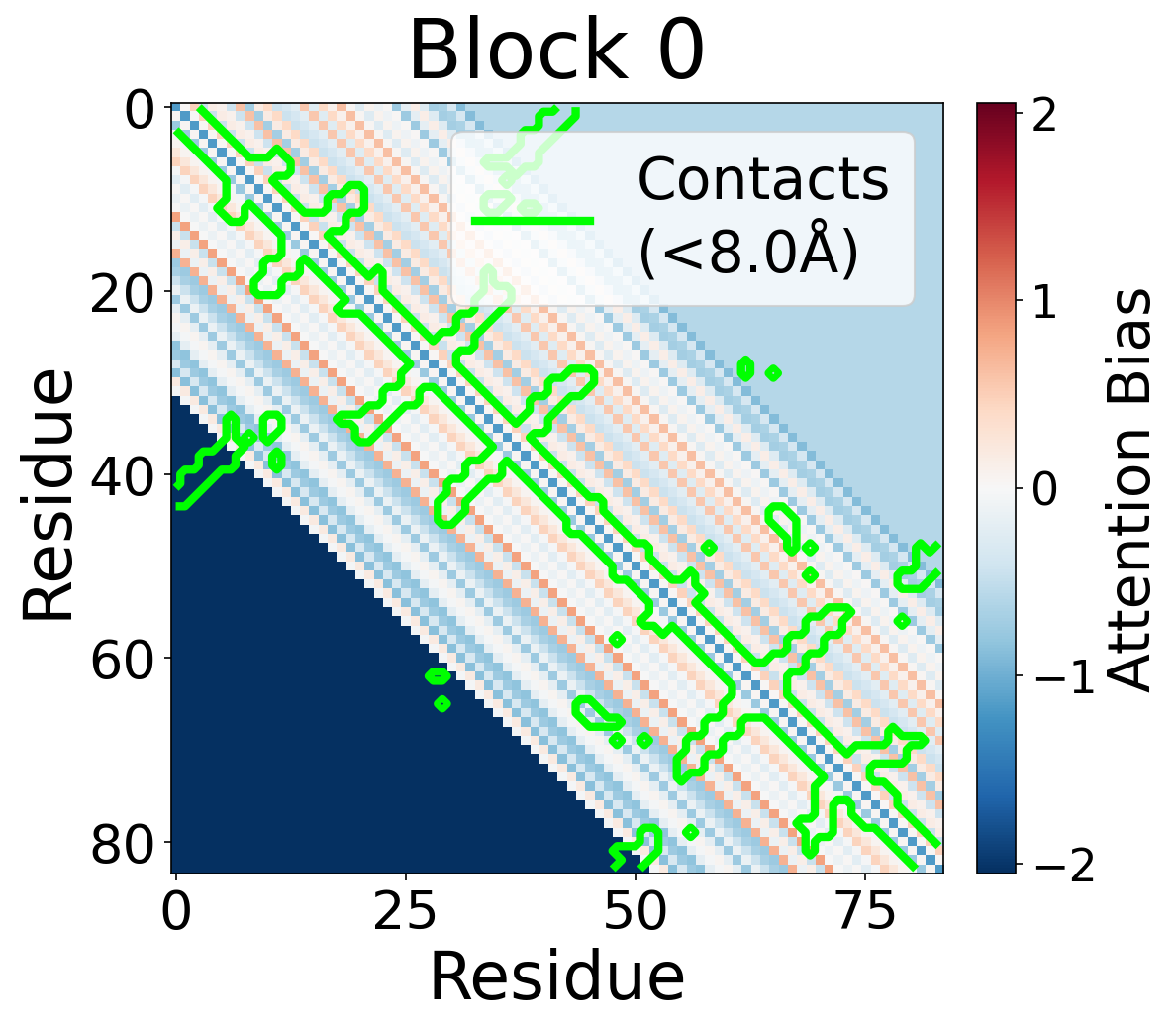}
    \end{subfigure}
    \hfill
    \begin{subfigure}[b]{0.24\textwidth}
        \includegraphics[width=\textwidth]{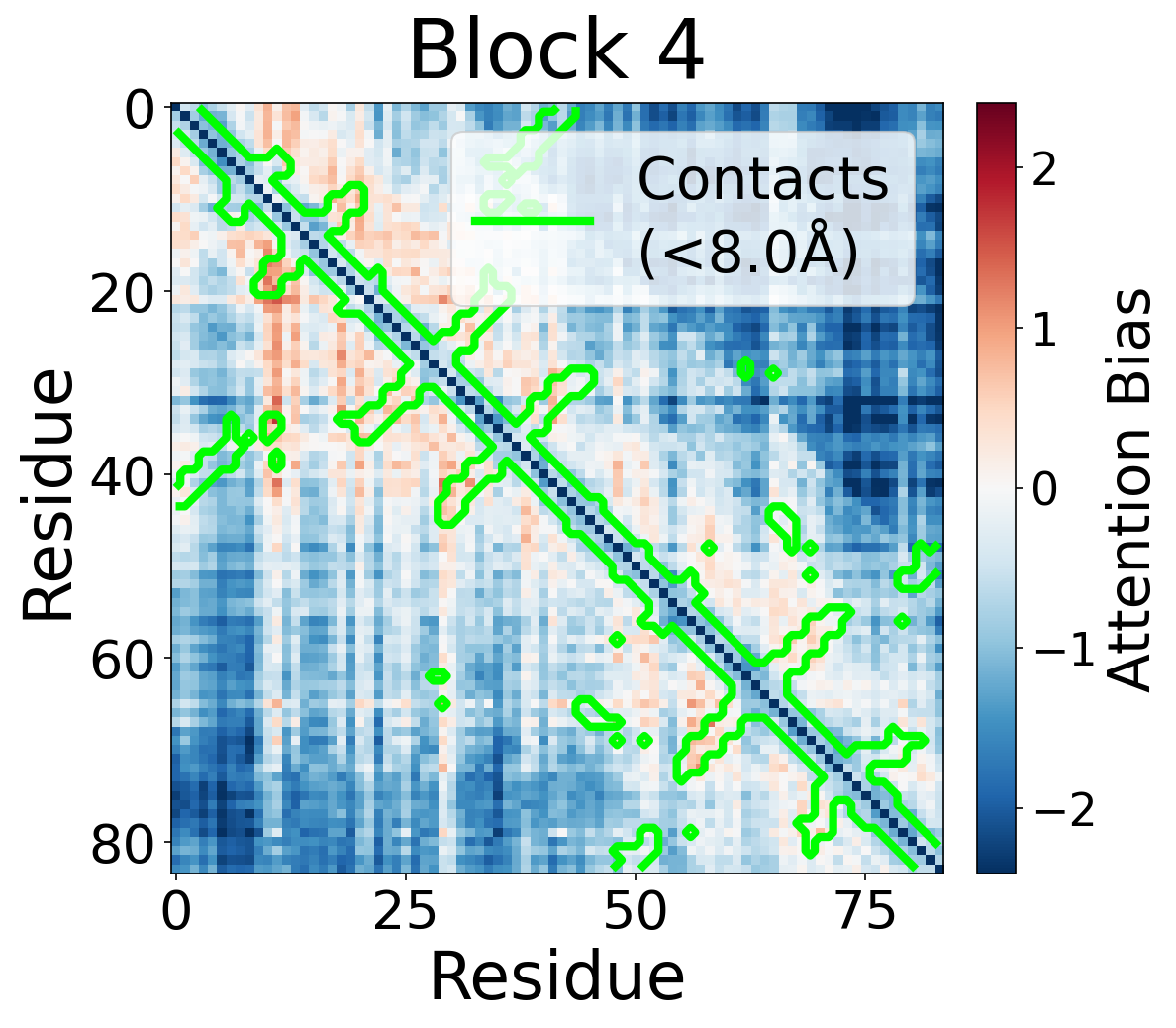}
    \end{subfigure}
    \hfill
    \begin{subfigure}[b]{0.24\textwidth}
        \includegraphics[width=\textwidth]{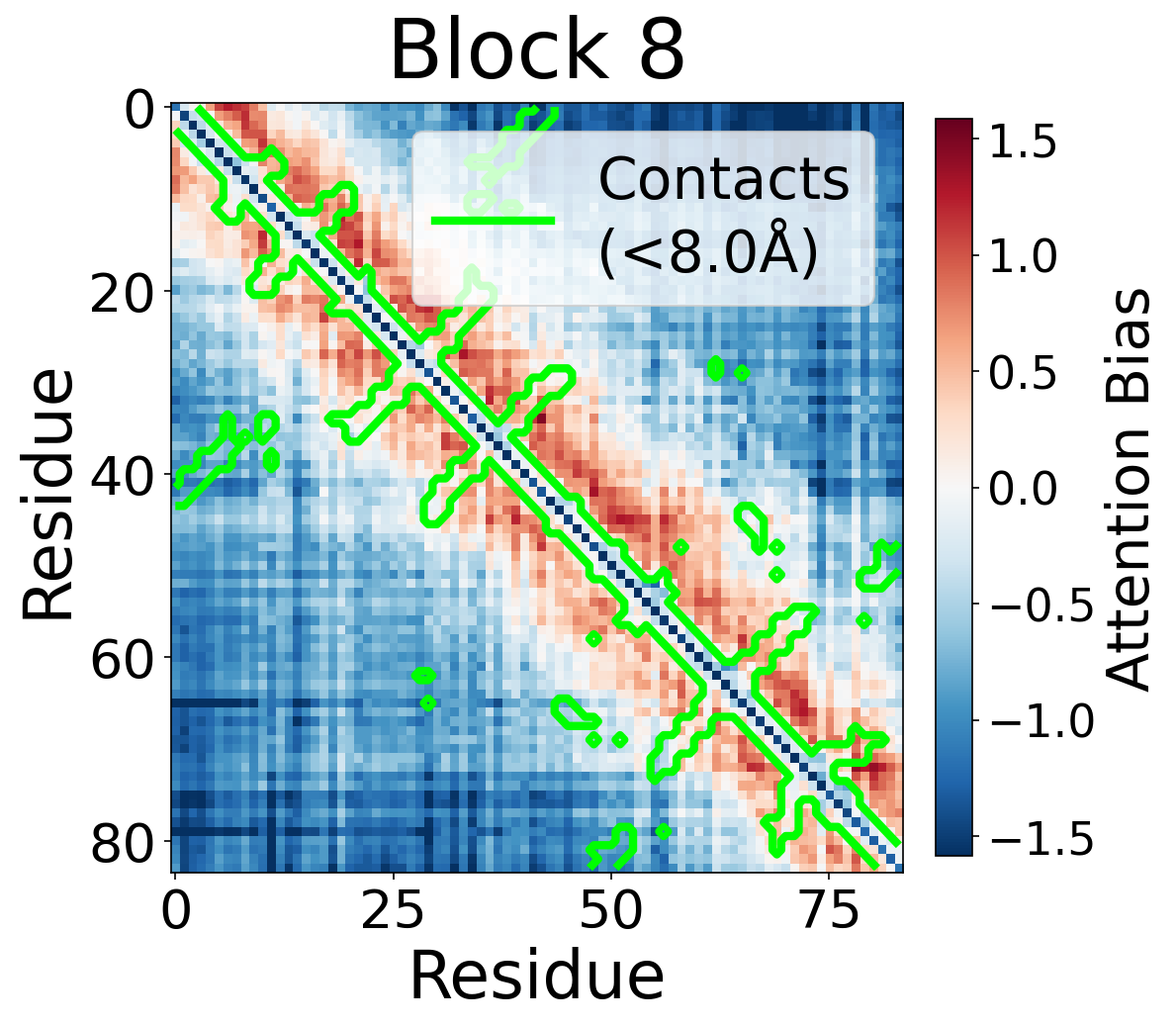}
    \end{subfigure}
    \hfill
    \begin{subfigure}[b]{0.24\textwidth}
        \includegraphics[width=\textwidth]{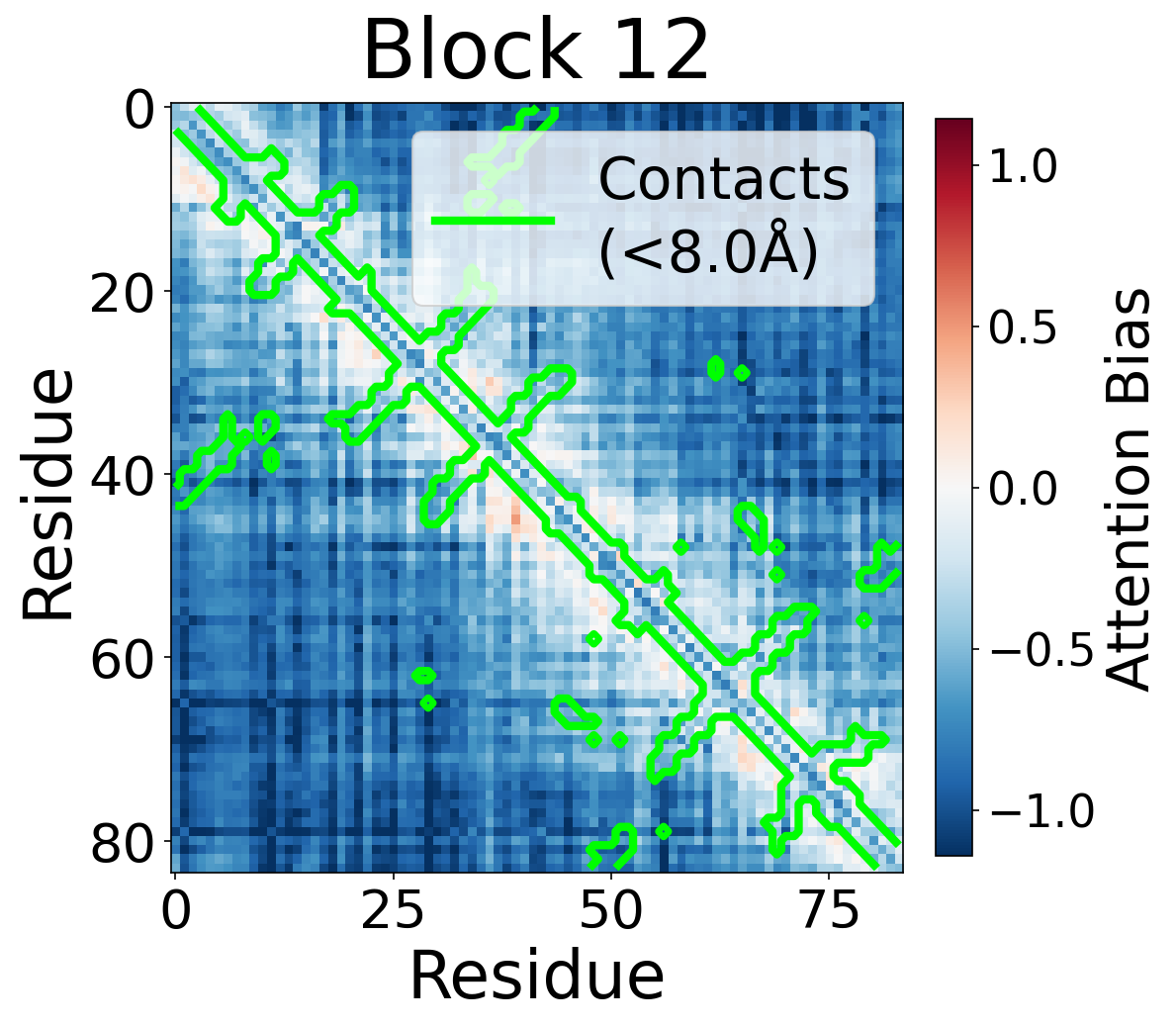}
    \end{subfigure}

    \vspace{0.5em}

    \begin{subfigure}[b]{0.24\textwidth}
        \includegraphics[width=\textwidth]{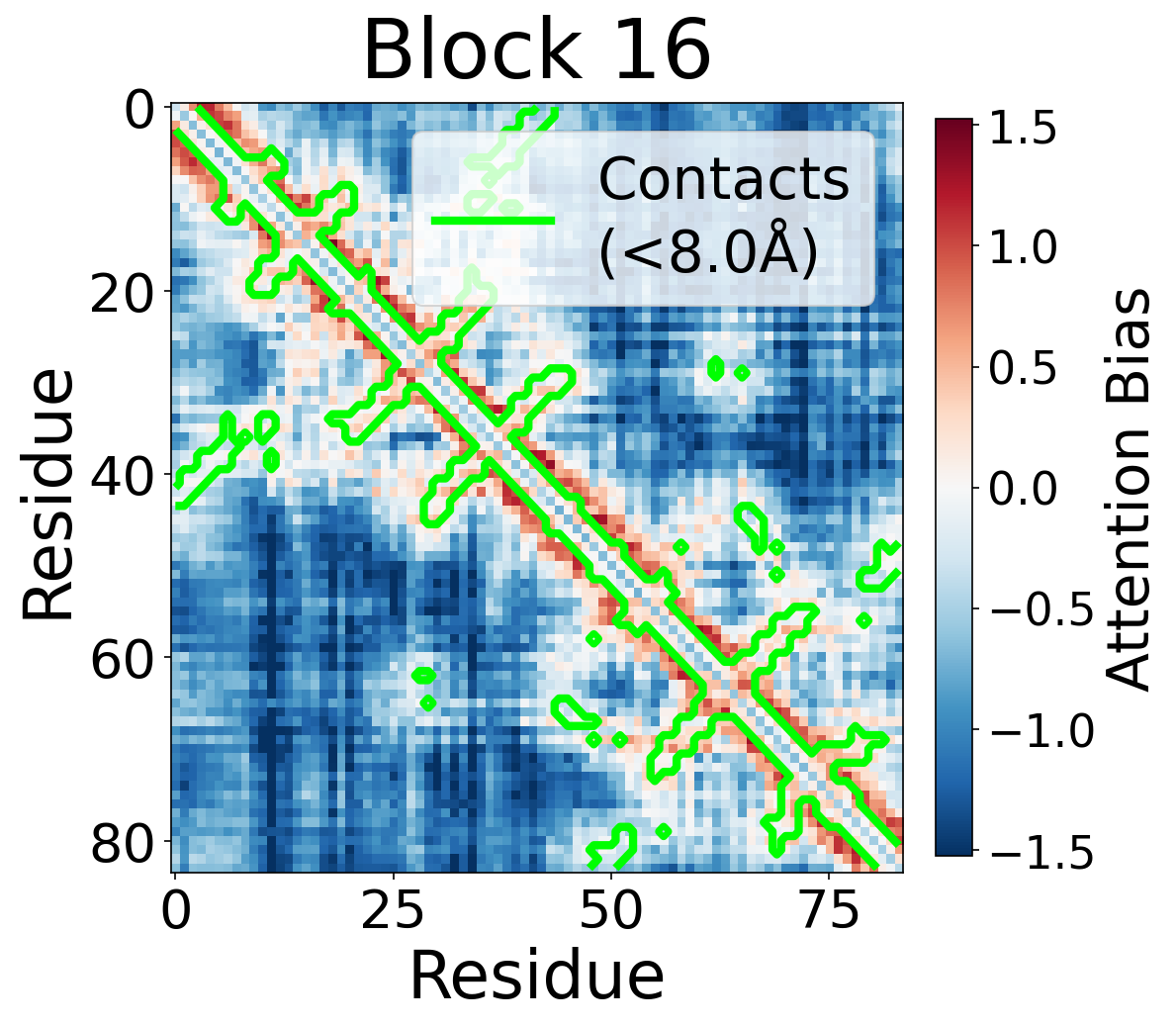}
    \end{subfigure}
    \hfill
    \begin{subfigure}[b]{0.24\textwidth}
        \includegraphics[width=\textwidth]{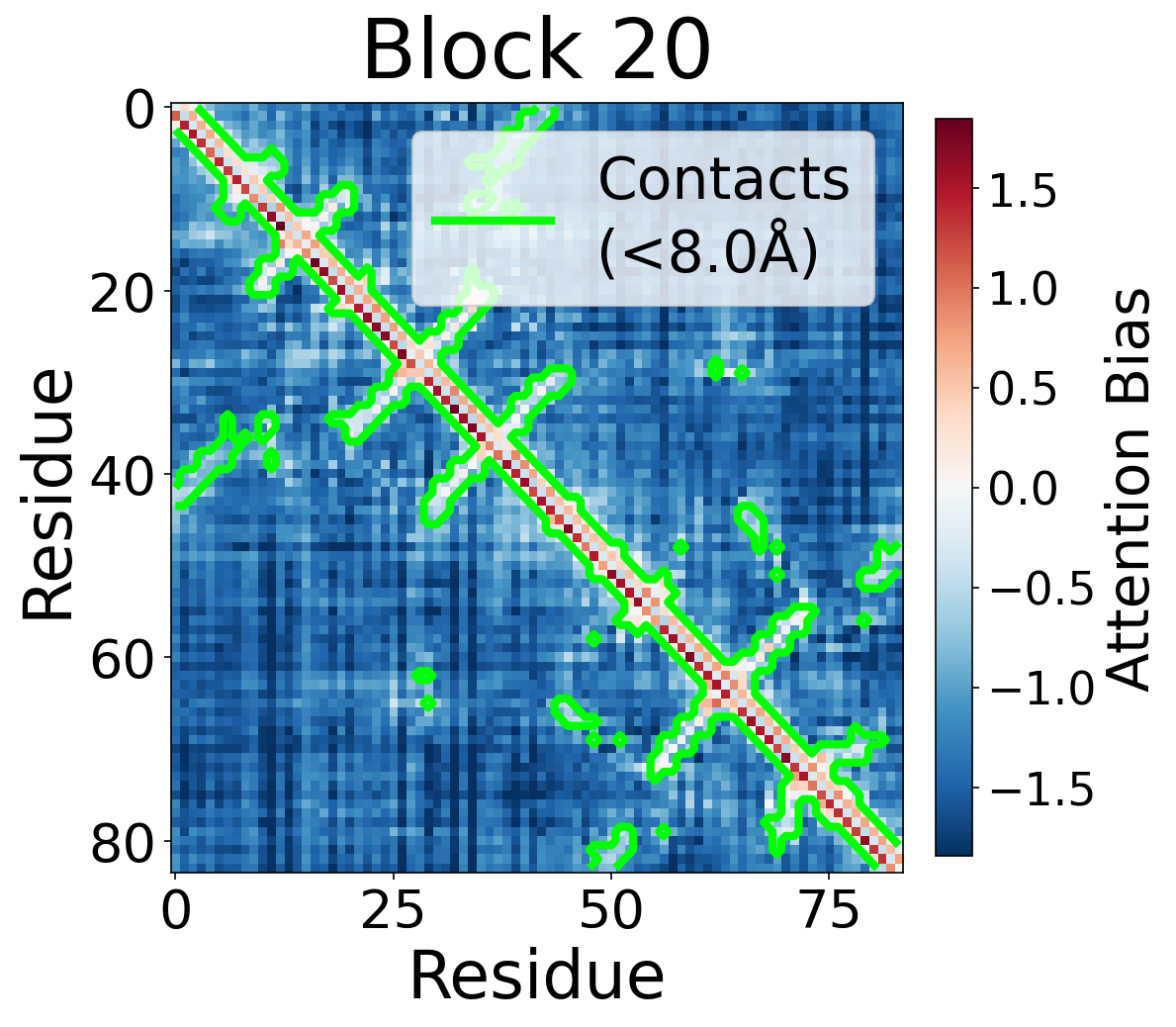}
    \end{subfigure}
    \hfill
    \begin{subfigure}[b]{0.24\textwidth}
        \includegraphics[width=\textwidth]{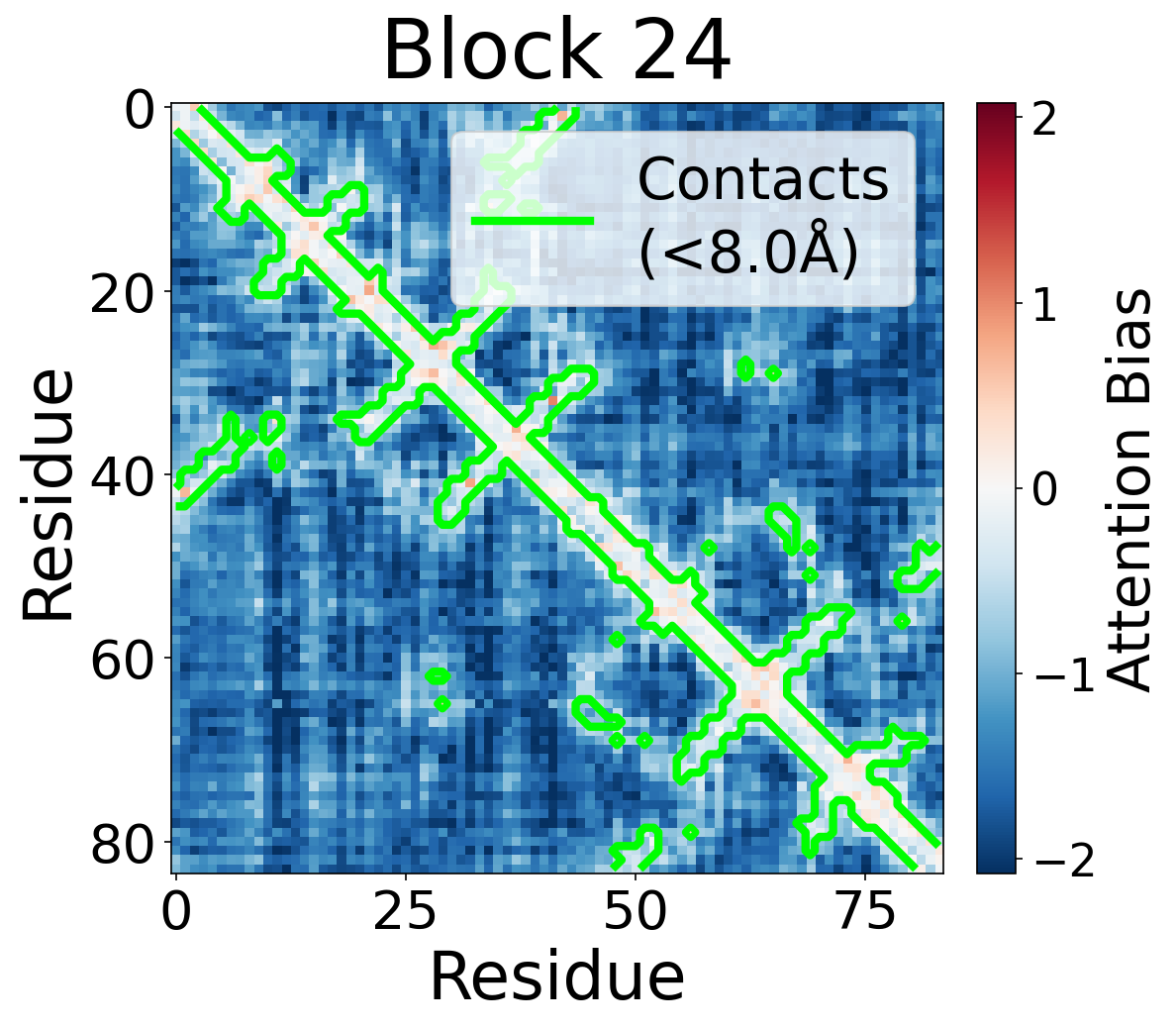}
    \end{subfigure}
    \hfill
    \begin{subfigure}[b]{0.24\textwidth}
        \includegraphics[width=\textwidth]{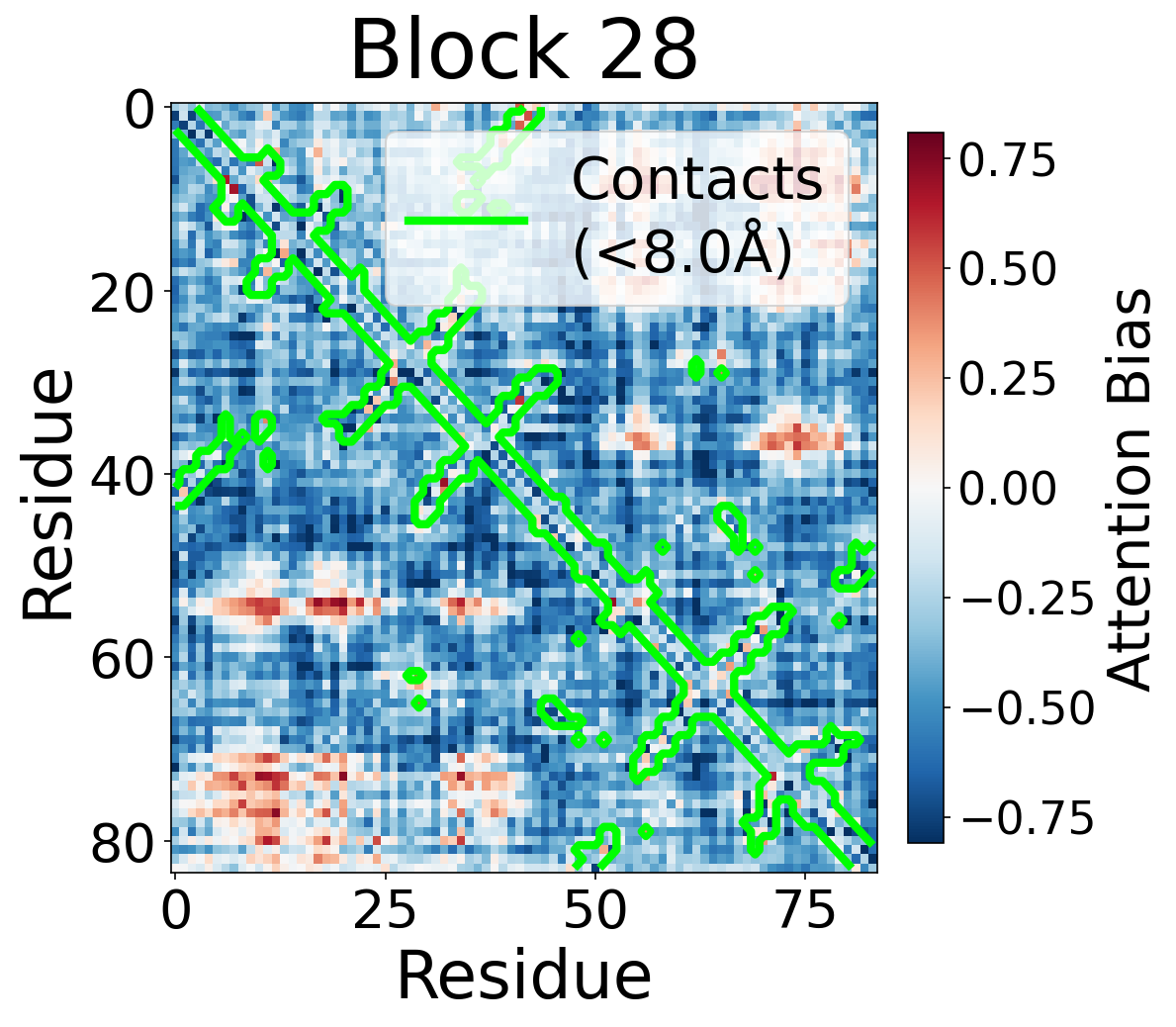}
    \end{subfigure}

    \vspace{0.5em}

    \begin{subfigure}[b]{0.24\textwidth}
        \includegraphics[width=\textwidth]{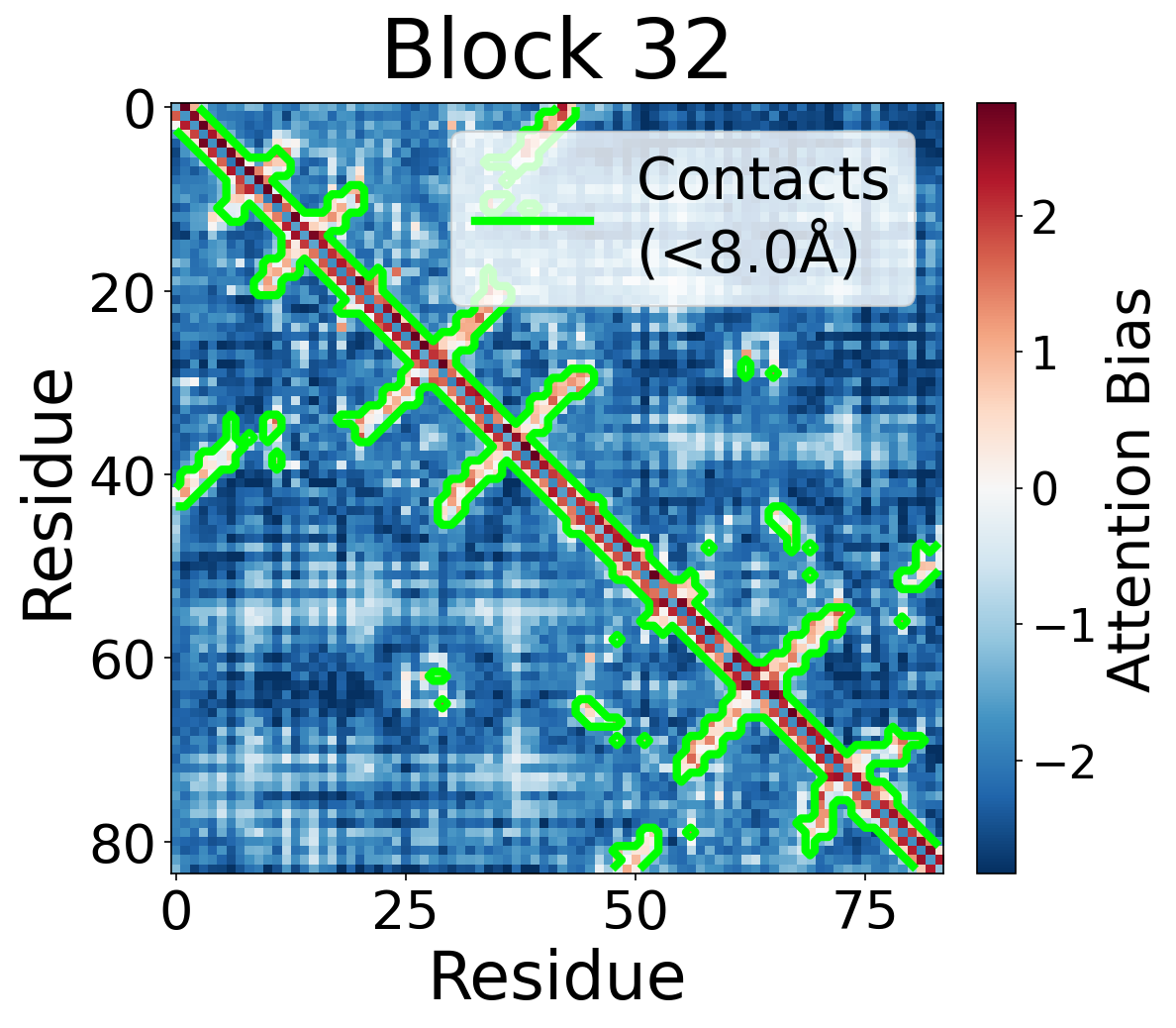}
    \end{subfigure}
    \hfill
    \begin{subfigure}[b]{0.24\textwidth}
        \includegraphics[width=\textwidth]{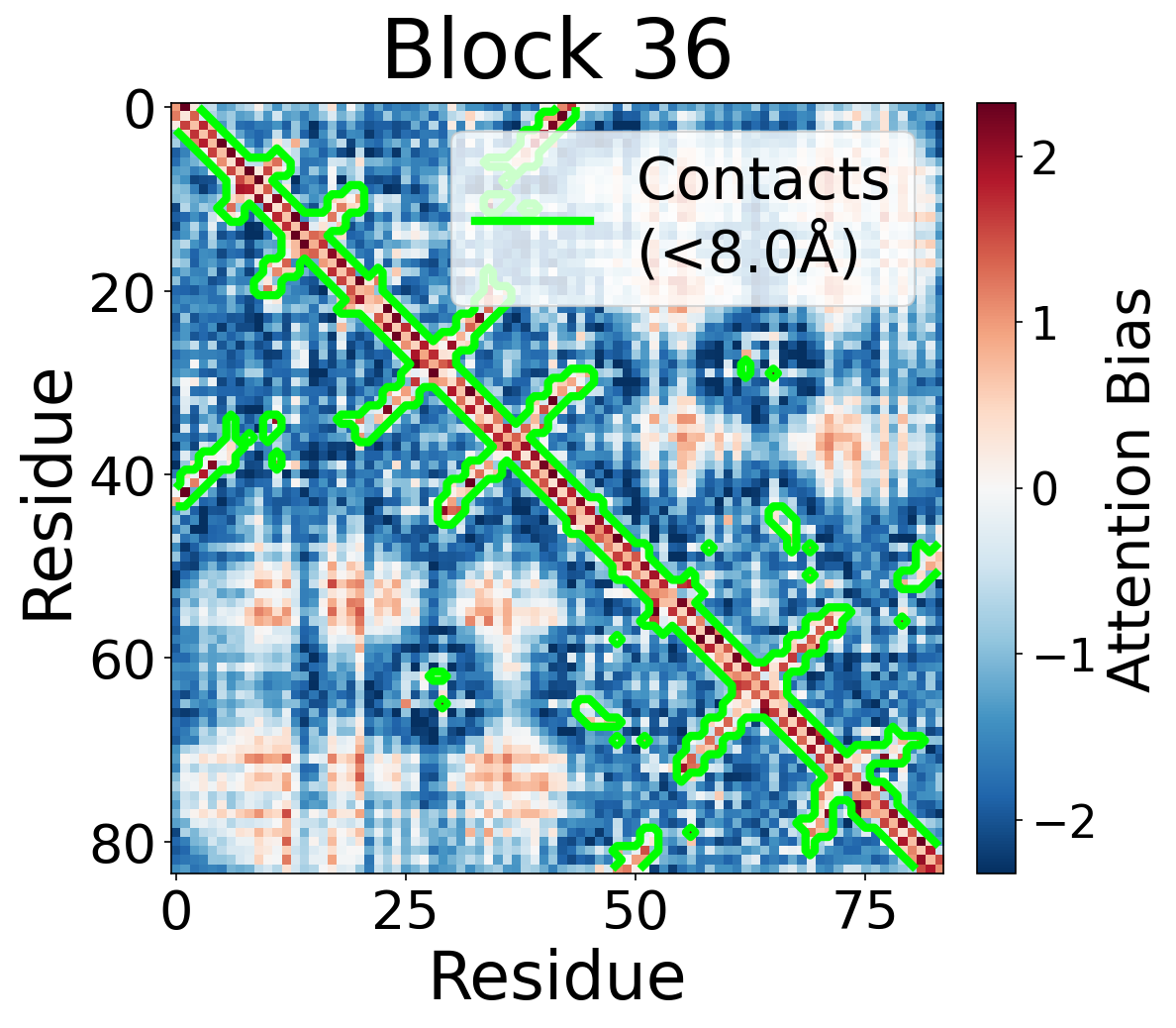}
    \end{subfigure}
    \hfill
    \begin{subfigure}[b]{0.24\textwidth}
        \includegraphics[width=\textwidth]{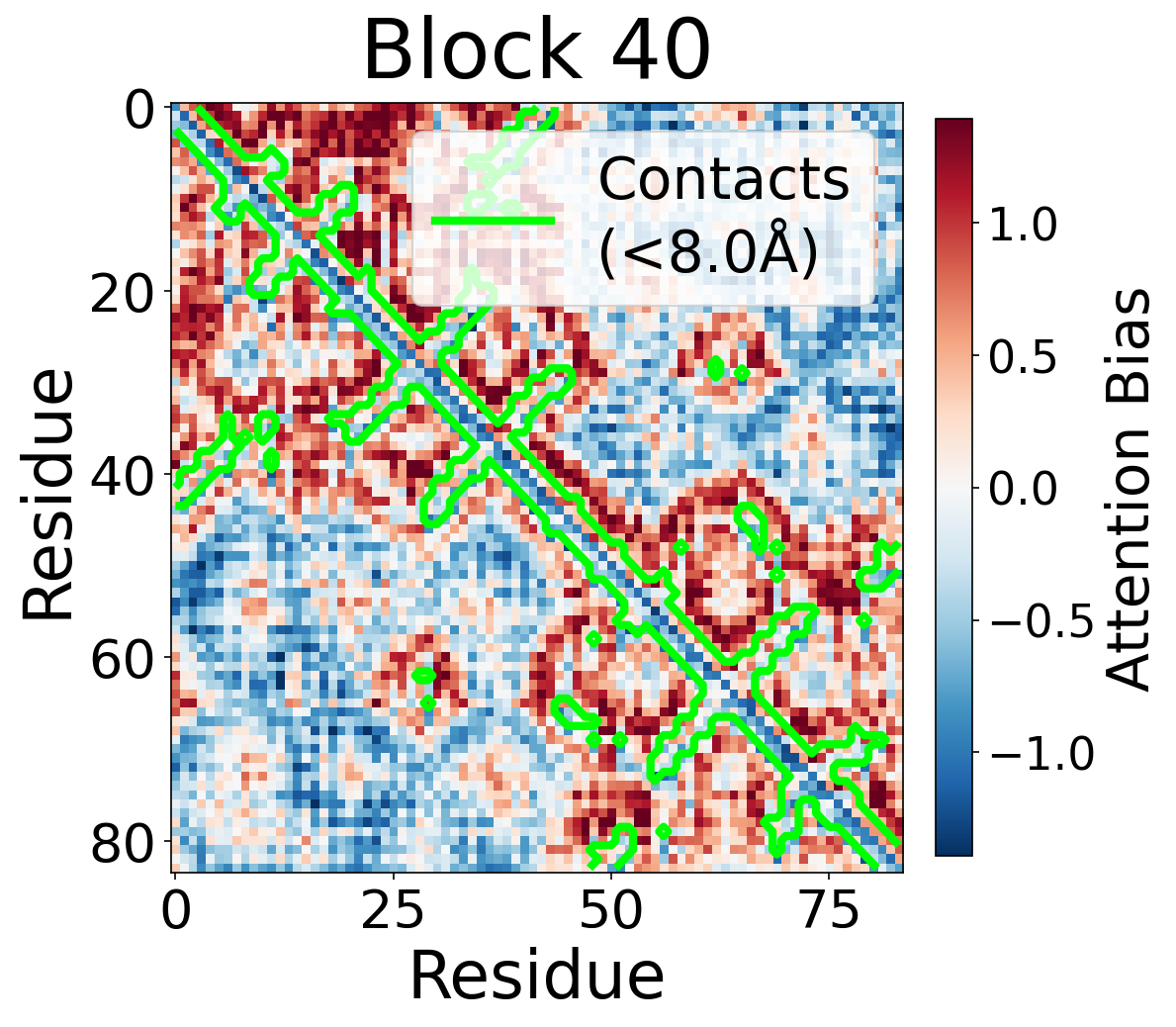}
    \end{subfigure}
    \hfill
    \begin{subfigure}[b]{0.24\textwidth}
        \includegraphics[width=\textwidth]{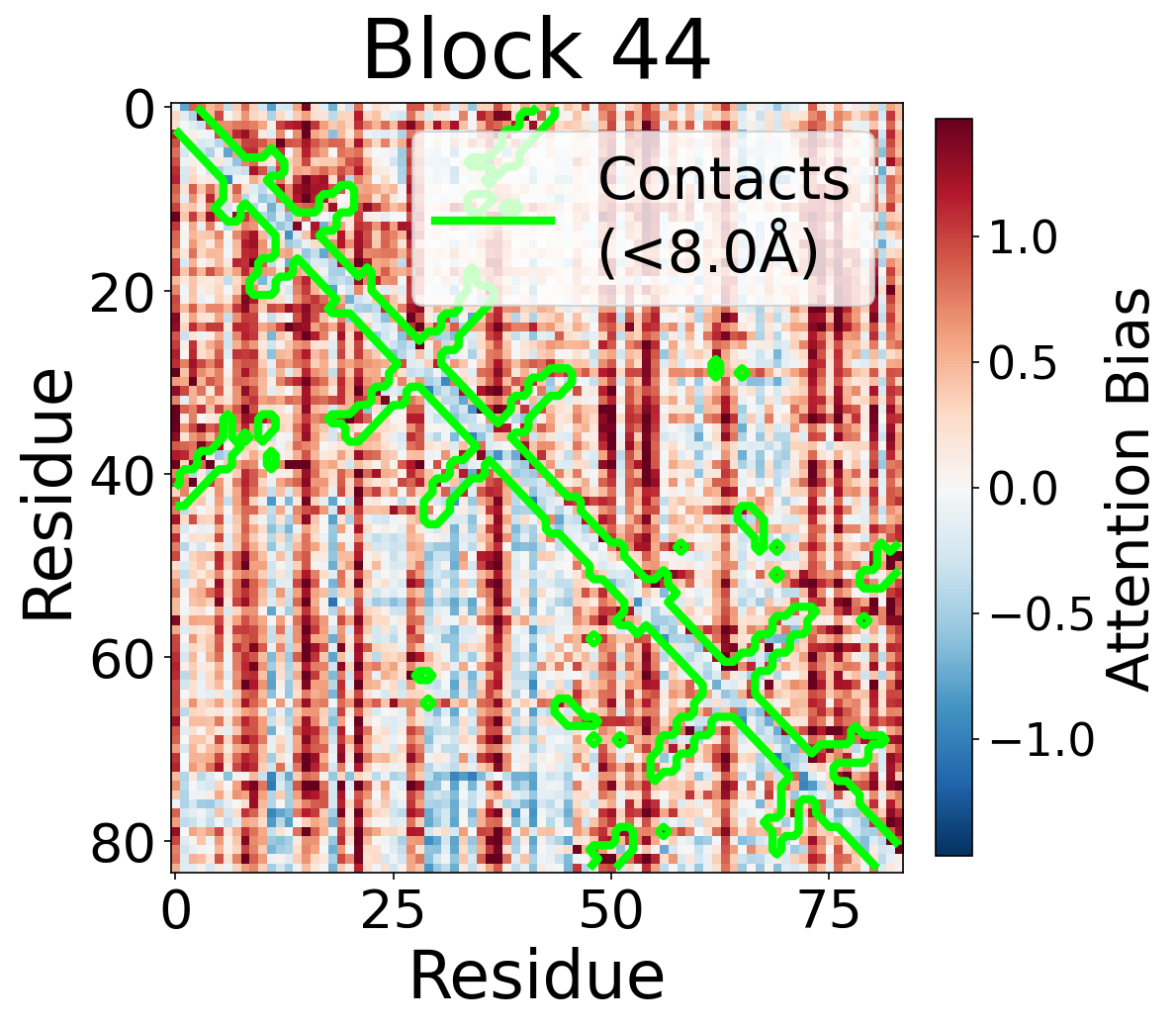}
    \end{subfigure}

    \caption{\textbf{\texttt{Pair2seq} attention bias across blocks (head-averaged).}
    Each panel shows the bias term $\beta_{ij}$ from Eq.~\eqref{eq:pair2seq}, averaged across all 8 attention heads, for protein 6rwc at selected blocks of the folding trunk. Red indicates positive bias (encouraging attention); blue indicates negative bias. Green contours mark structural contacts (C$\alpha$ $<$ 8\AA). In early blocks, the bias is near-uniform; by the middle blocks, it begins to align with the contact map, and by late blocks, contacting residue pairs receive substantially higher bias than non-contacts.}
    \label{fig:bias_maps_blocks}
\end{figure*}

\begin{figure*}[t]
    \centering

    \begin{subfigure}[b]{0.24\textwidth}
        \includegraphics[width=\textwidth]{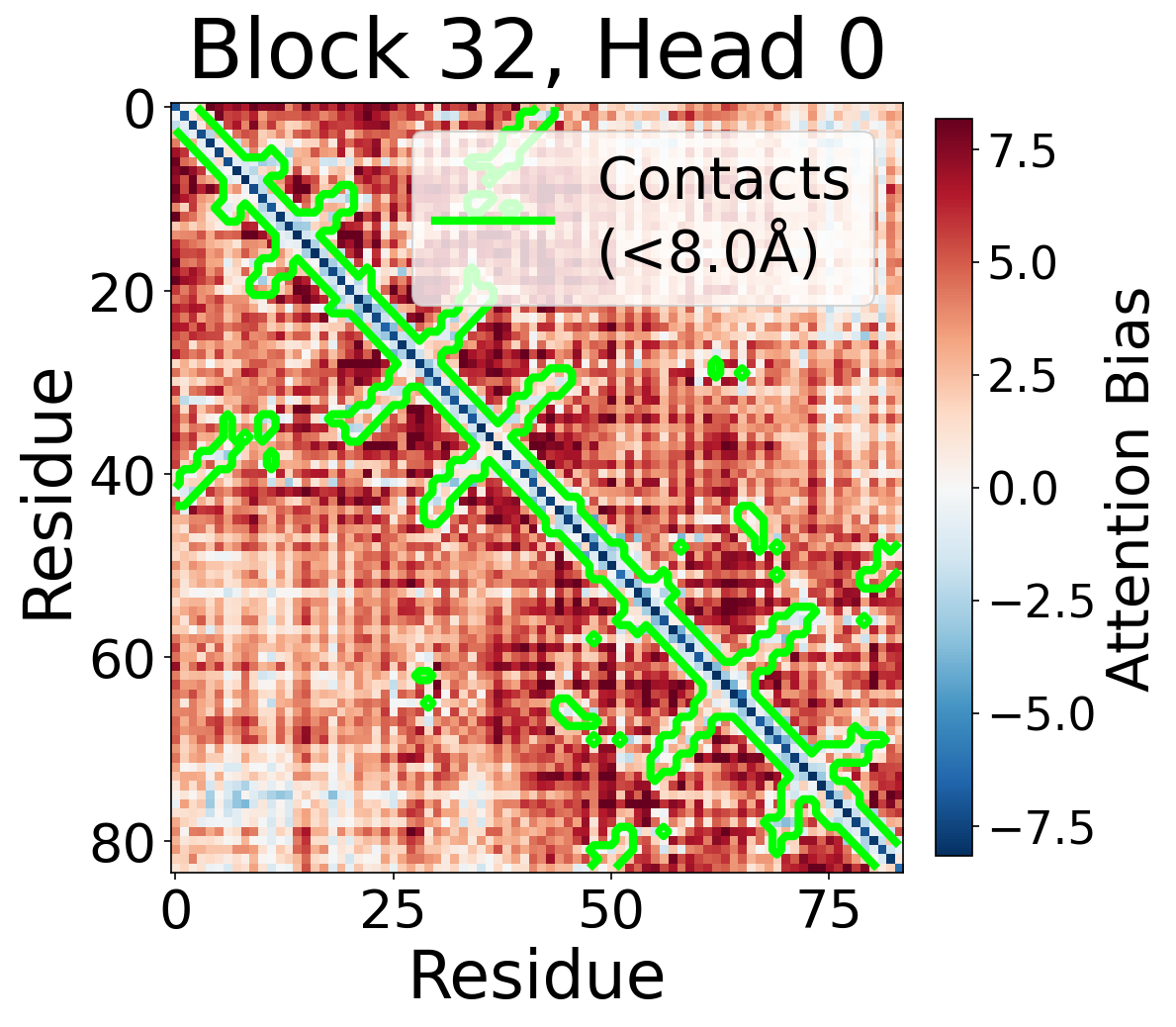}
    \end{subfigure}
    \hfill
    \begin{subfigure}[b]{0.24\textwidth}
        \includegraphics[width=\textwidth]{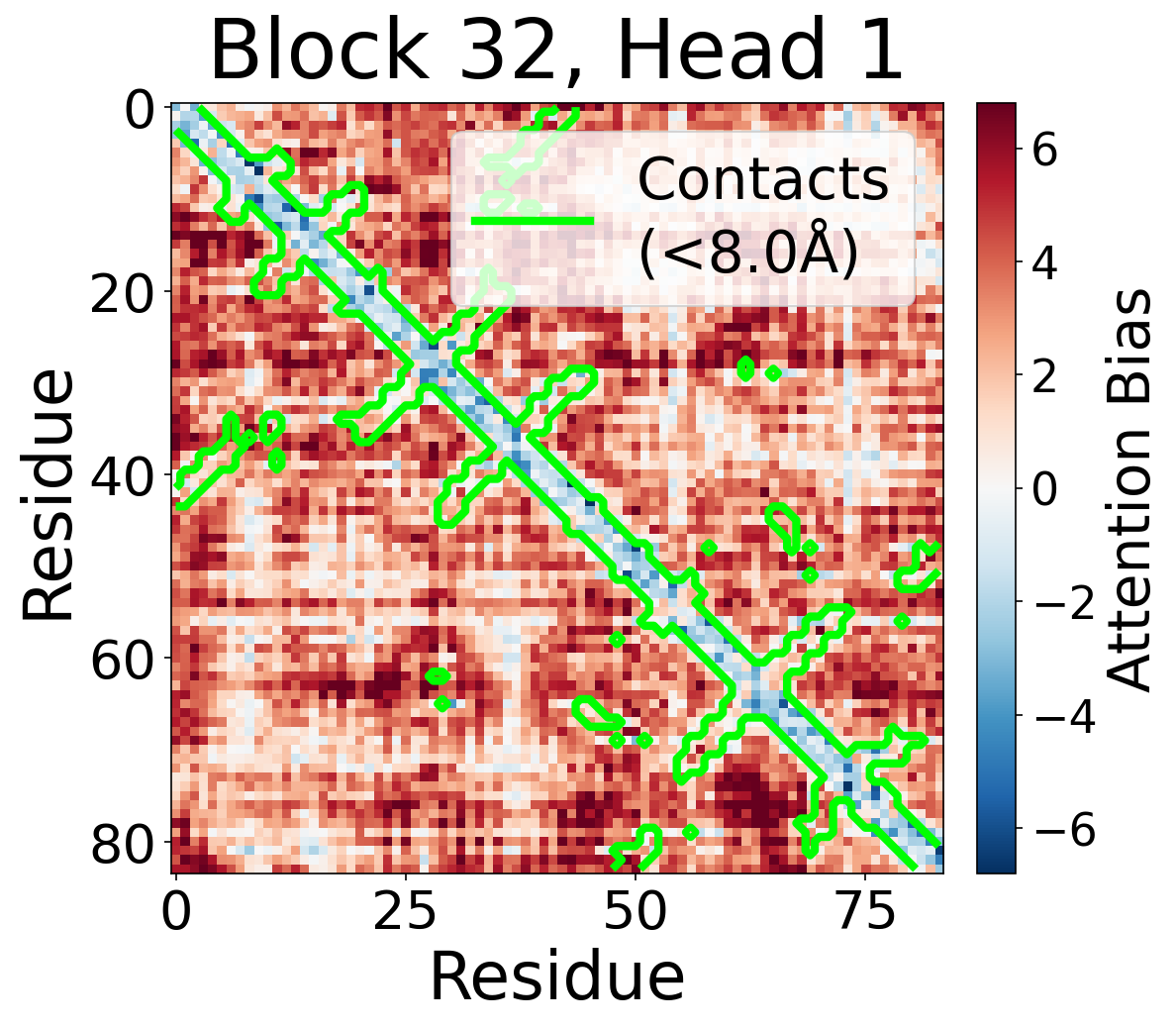}
    \end{subfigure}
    \hfill
    \begin{subfigure}[b]{0.24\textwidth}
        \includegraphics[width=\textwidth]{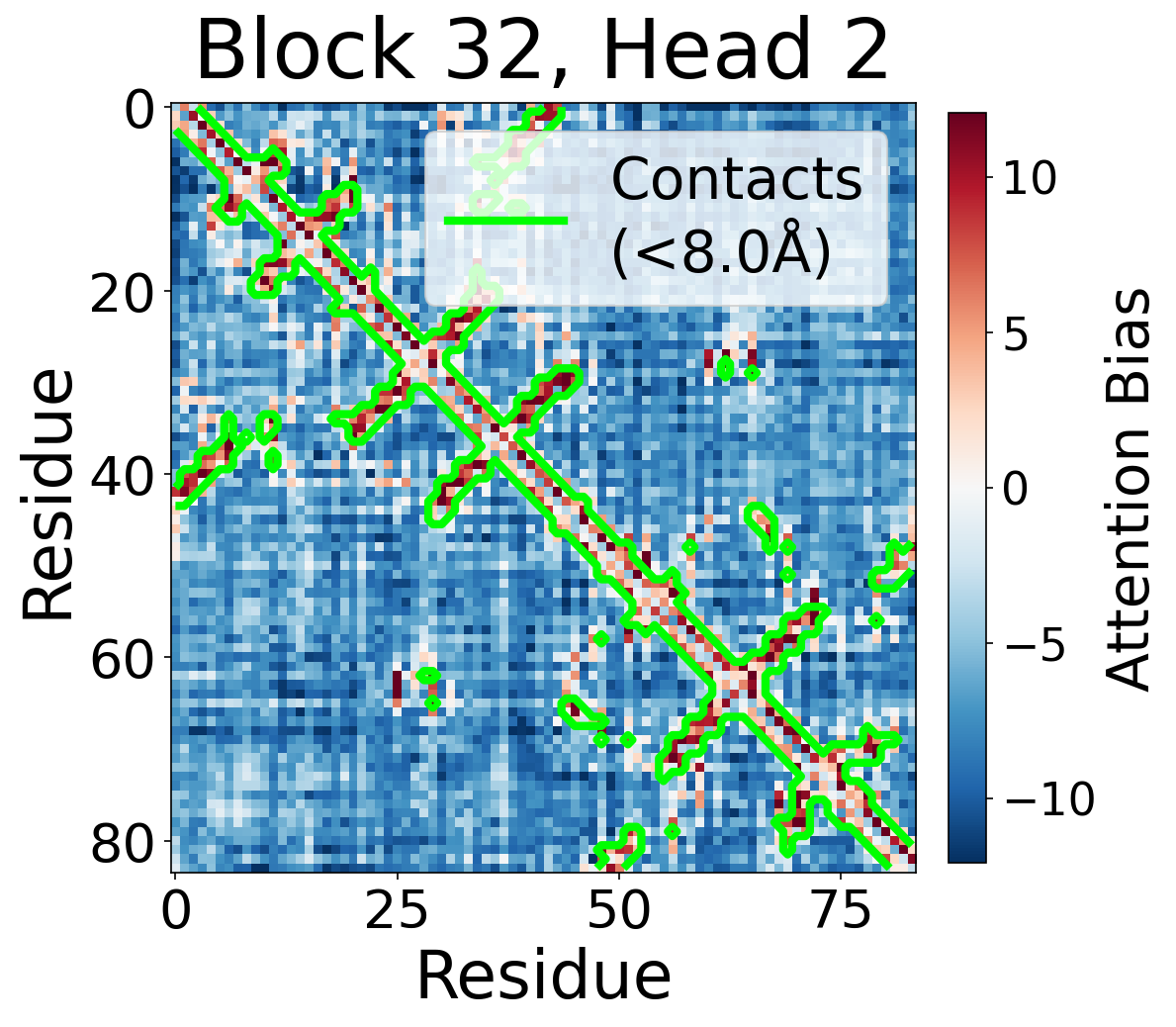}
    \end{subfigure}
    \hfill
    \begin{subfigure}[b]{0.24\textwidth}
        \includegraphics[width=\textwidth]{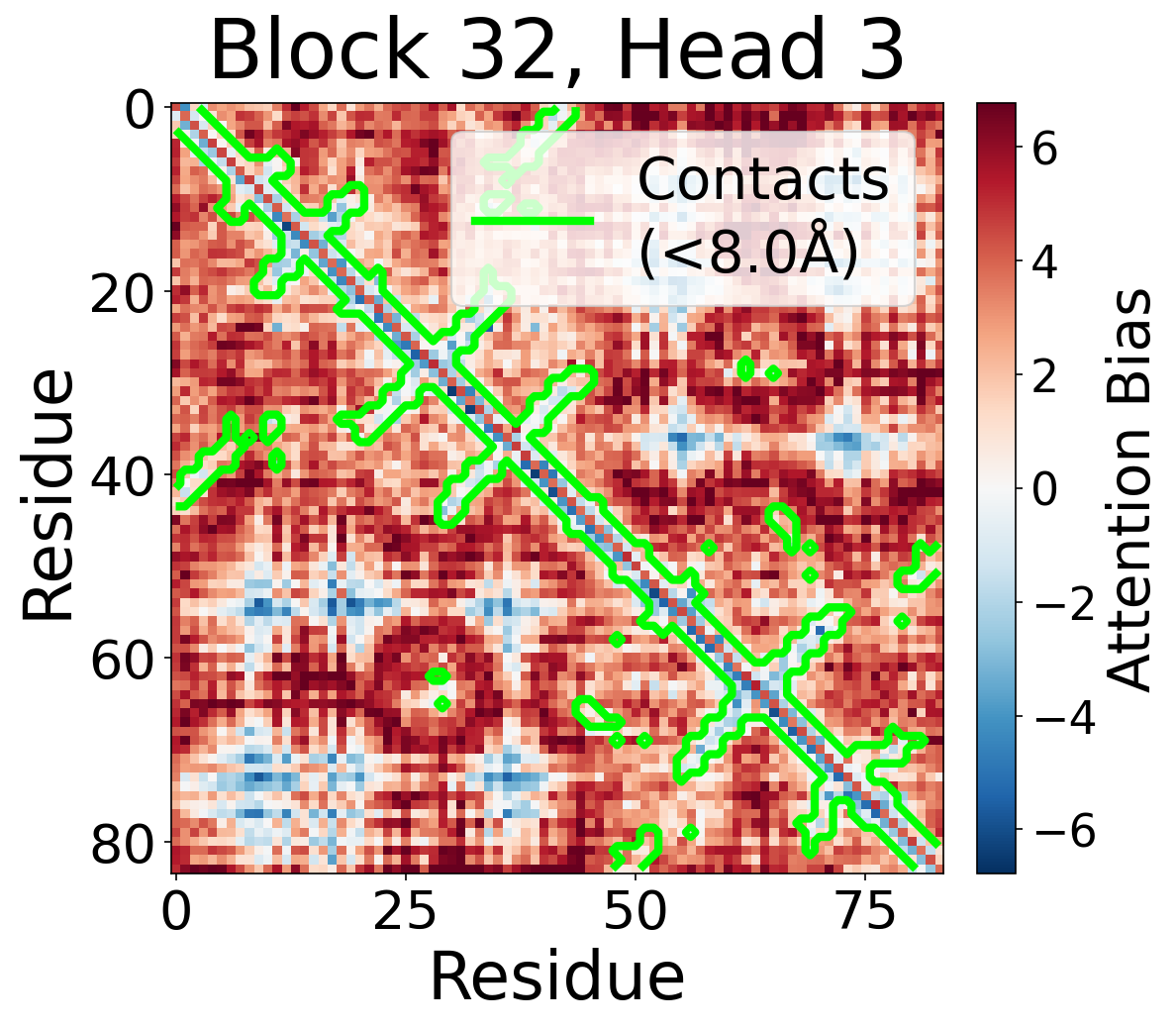}
    \end{subfigure}

    \vspace{0.5em}

    \begin{subfigure}[b]{0.24\textwidth}
        \includegraphics[width=\textwidth]{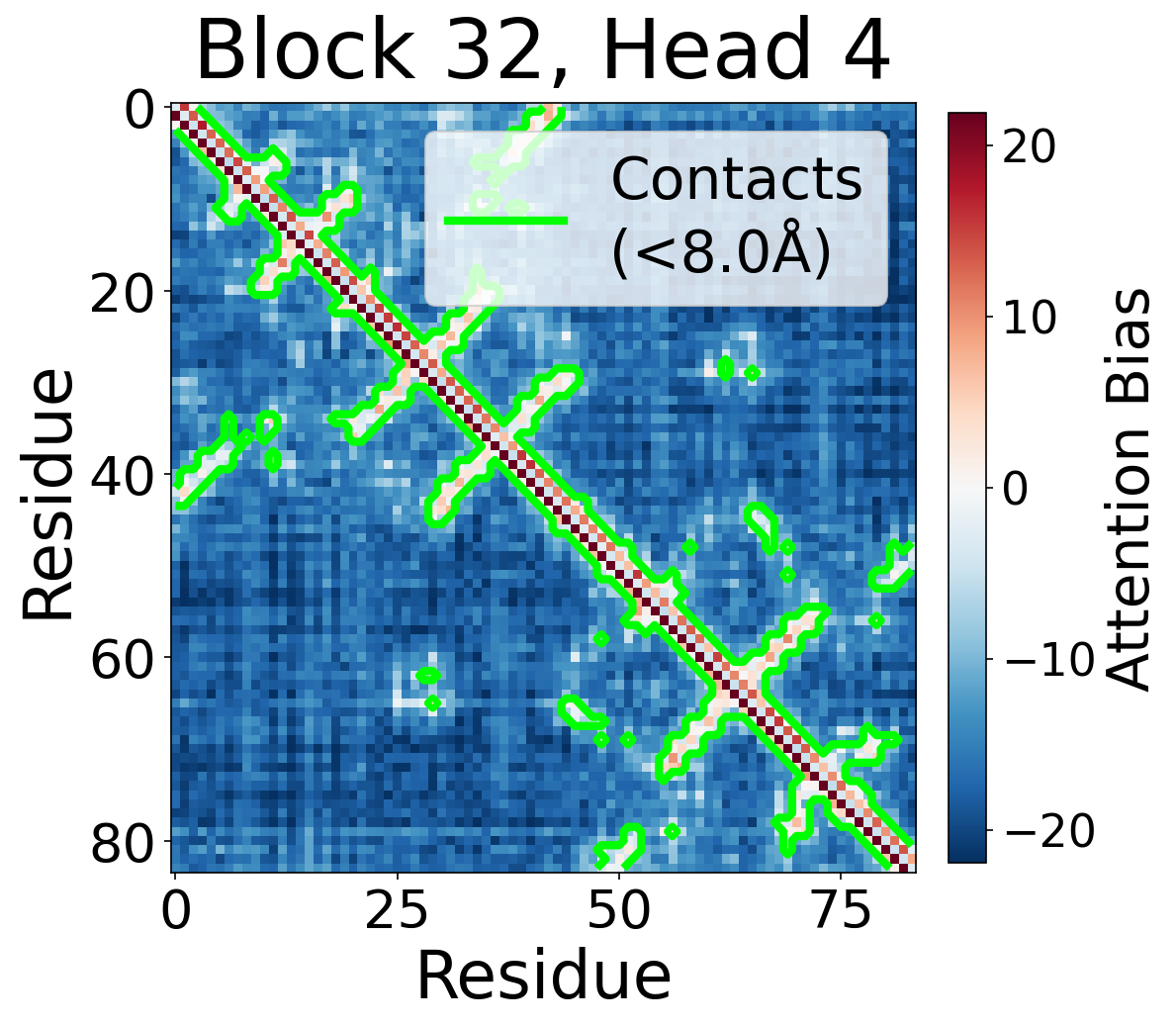}
    \end{subfigure}
    \hfill
    \begin{subfigure}[b]{0.24\textwidth}
        \includegraphics[width=\textwidth]{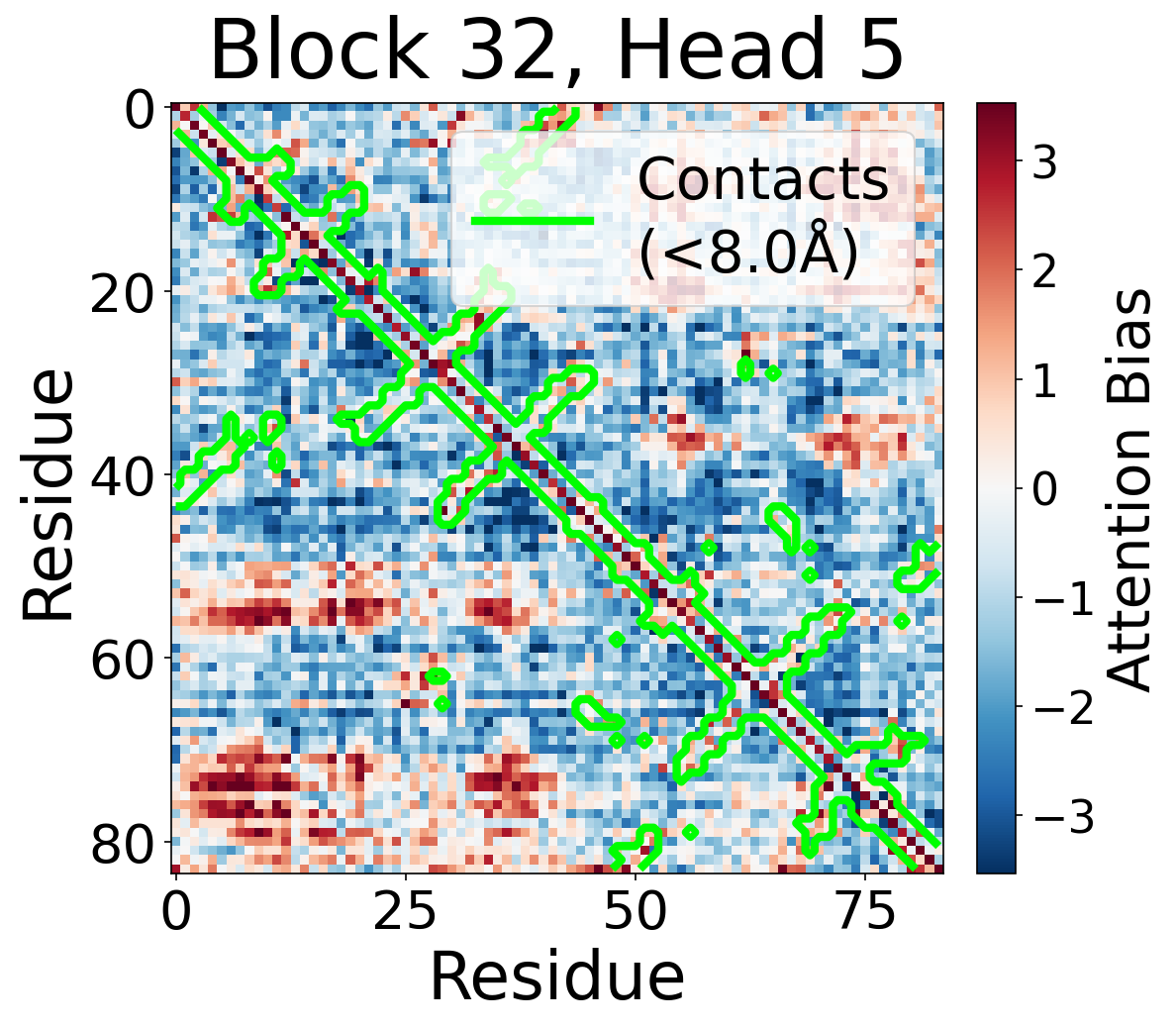}
    \end{subfigure}
    \hfill
    \begin{subfigure}[b]{0.24\textwidth}
        \includegraphics[width=\textwidth]{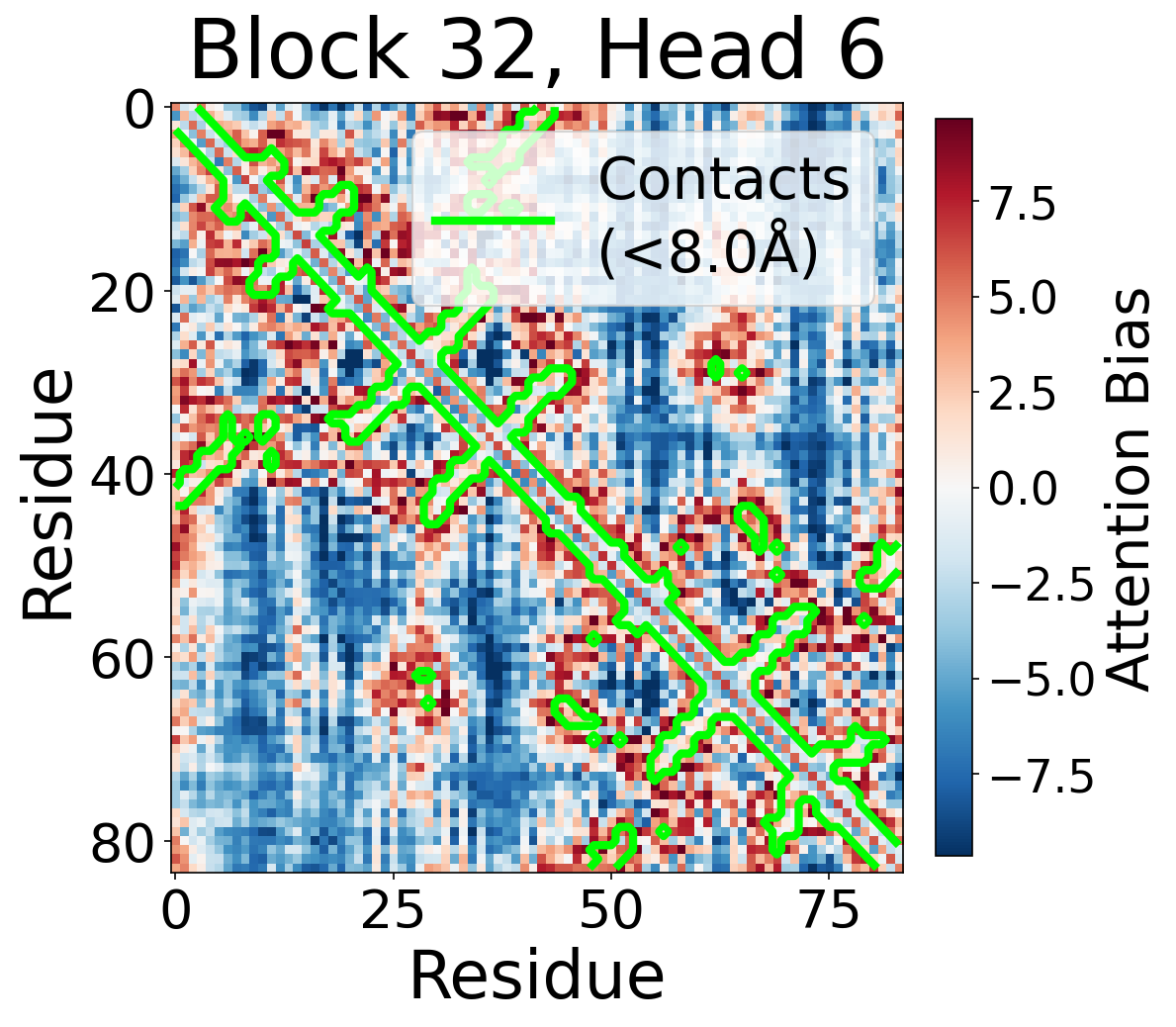}
    \end{subfigure}
    \hfill
    \begin{subfigure}[b]{0.24\textwidth}
        \includegraphics[width=\textwidth]{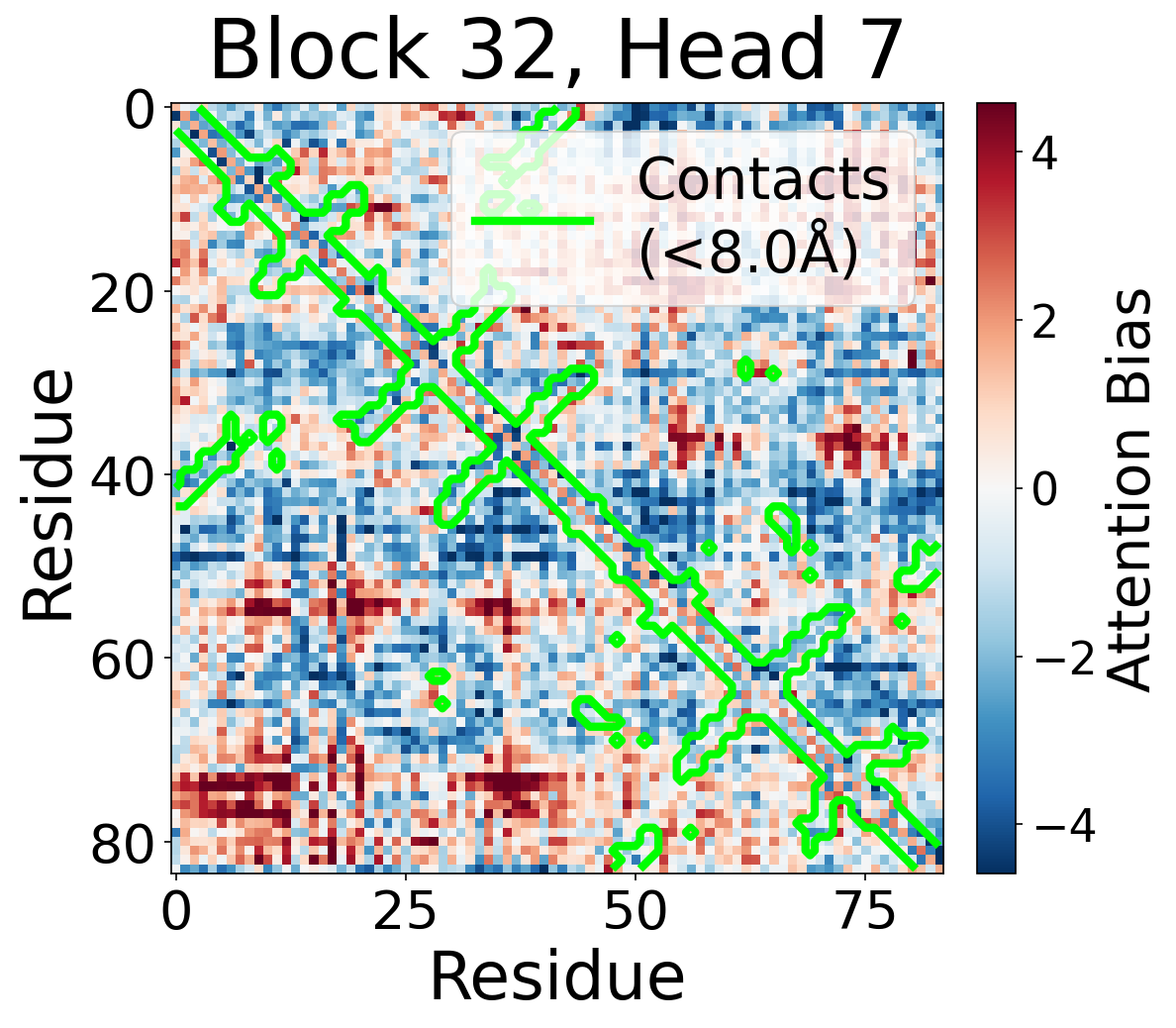}
    \end{subfigure}

    \caption{\textbf{Per-head \texttt{pair2seq} bias at block 32.}
    Individual attention head bias values for beta sheet protein 6rwc at block 32. Green contours mark structural contacts (C$\alpha$ $<$ 8\AA). Different heads exhibit distinct patterns: some heads show strong contact-aligned bias (e.g., heads that highlight the off-diagonal contact structure), while others capture different spatial relationships or show more diffuse patterns. The different bias patterns suggest possible head specialization; testing its effect on attention and structure requires further intervention.}
    \label{fig:bias_maps_heads}
\end{figure*}

\begin{figure*}[t]
    \centering

    \begin{subfigure}[b]{0.24\textwidth}
        \includegraphics[width=\textwidth]{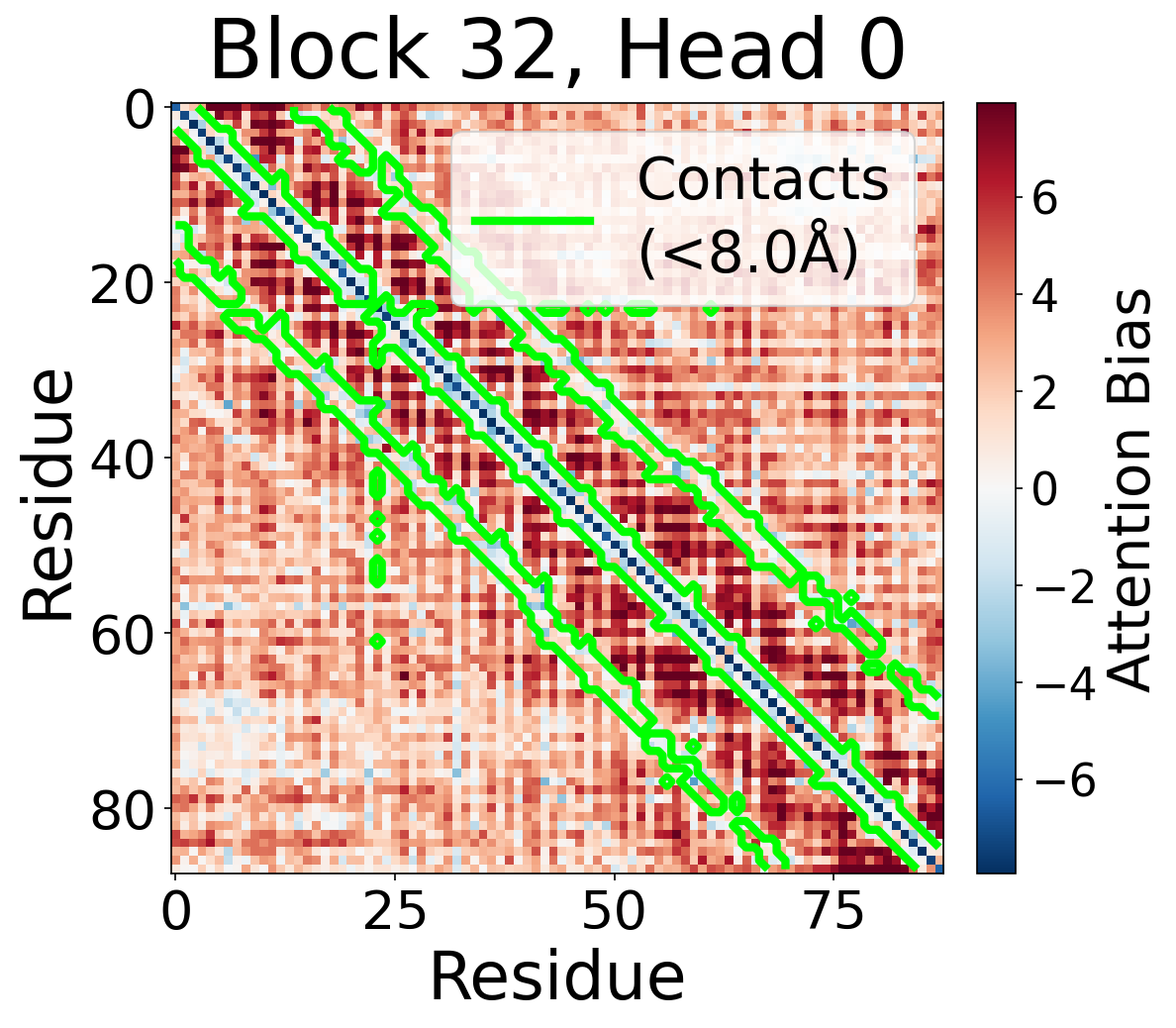}
    \end{subfigure}
    \hfill
    \begin{subfigure}[b]{0.24\textwidth}
        \includegraphics[width=\textwidth]{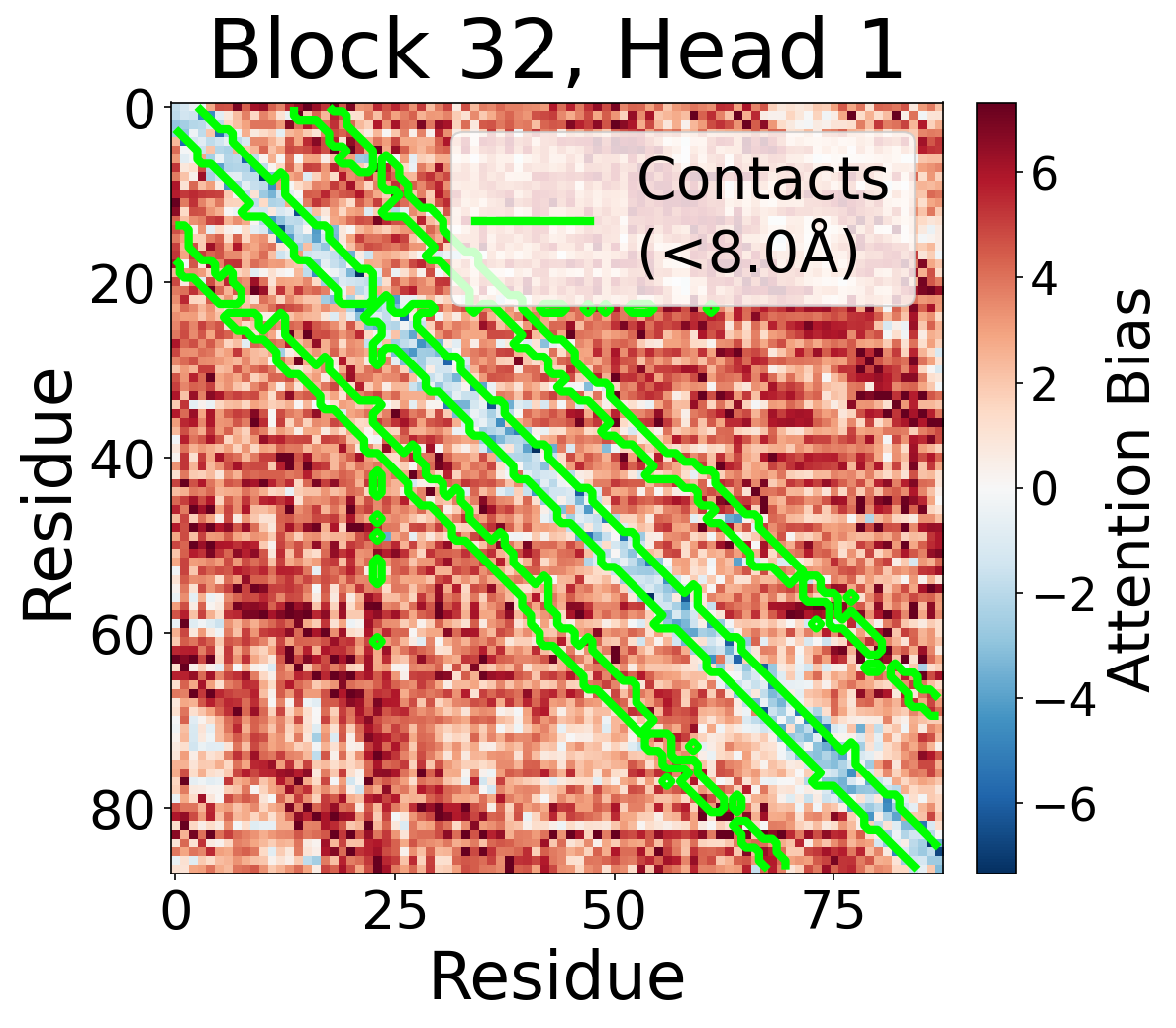}
    \end{subfigure}
    \hfill
    \begin{subfigure}[b]{0.24\textwidth}
        \includegraphics[width=\textwidth]{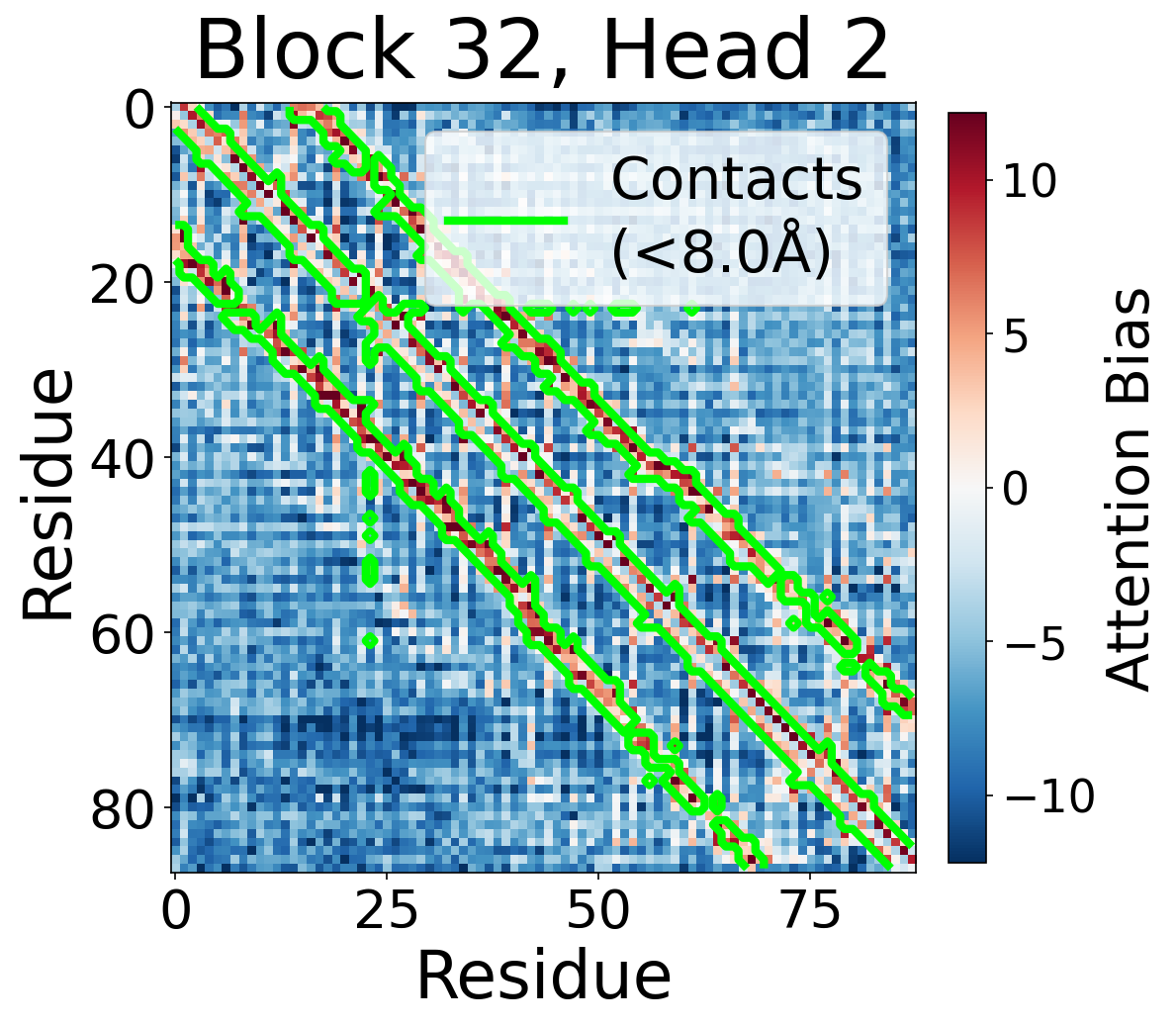}
    \end{subfigure}
    \hfill
    \begin{subfigure}[b]{0.24\textwidth}
        \includegraphics[width=\textwidth]{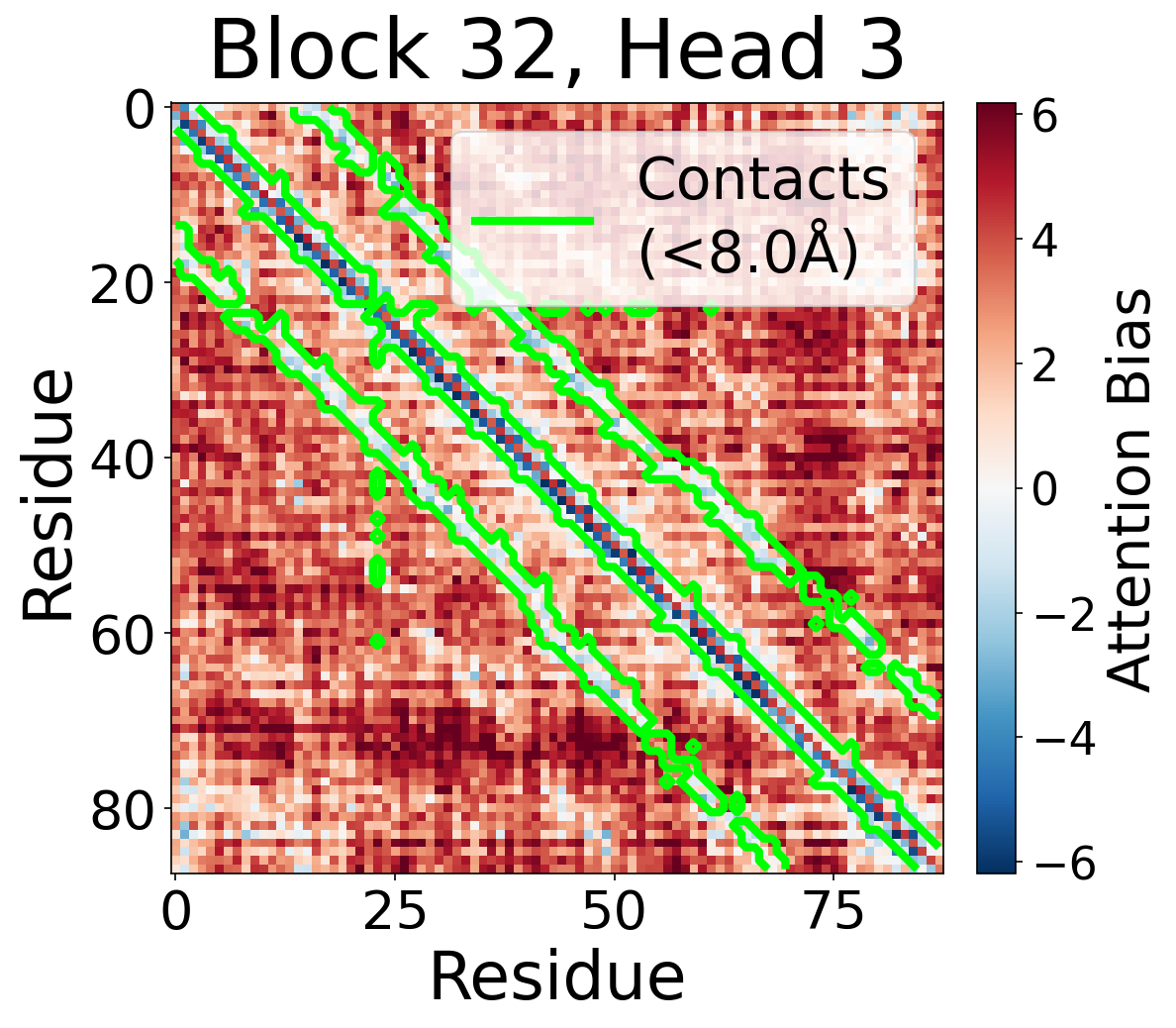}
    \end{subfigure}

    \vspace{0.5em}

    \begin{subfigure}[b]{0.24\textwidth}
        \includegraphics[width=\textwidth]{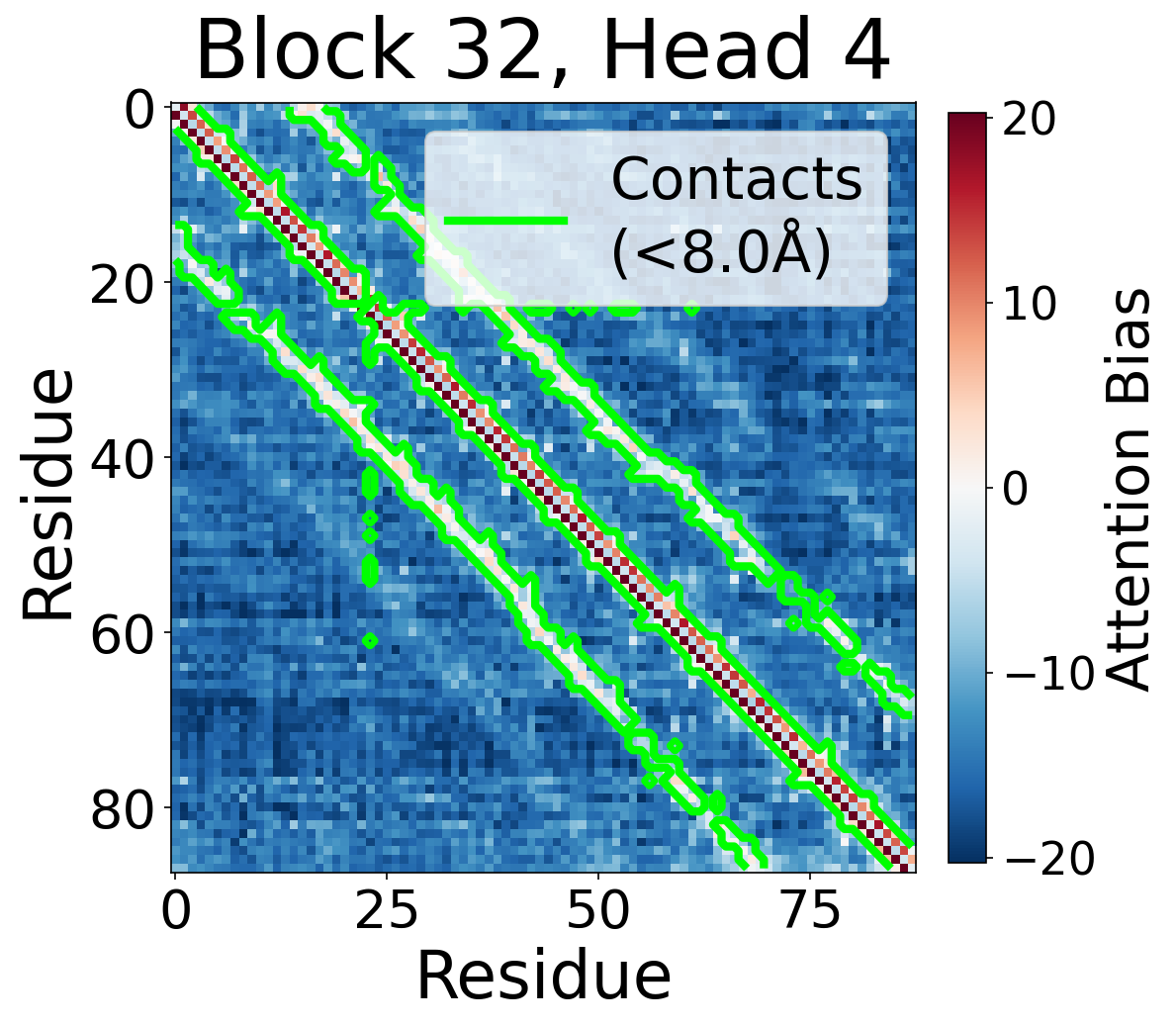}
    \end{subfigure}
    \hfill
    \begin{subfigure}[b]{0.24\textwidth}
        \includegraphics[width=\textwidth]{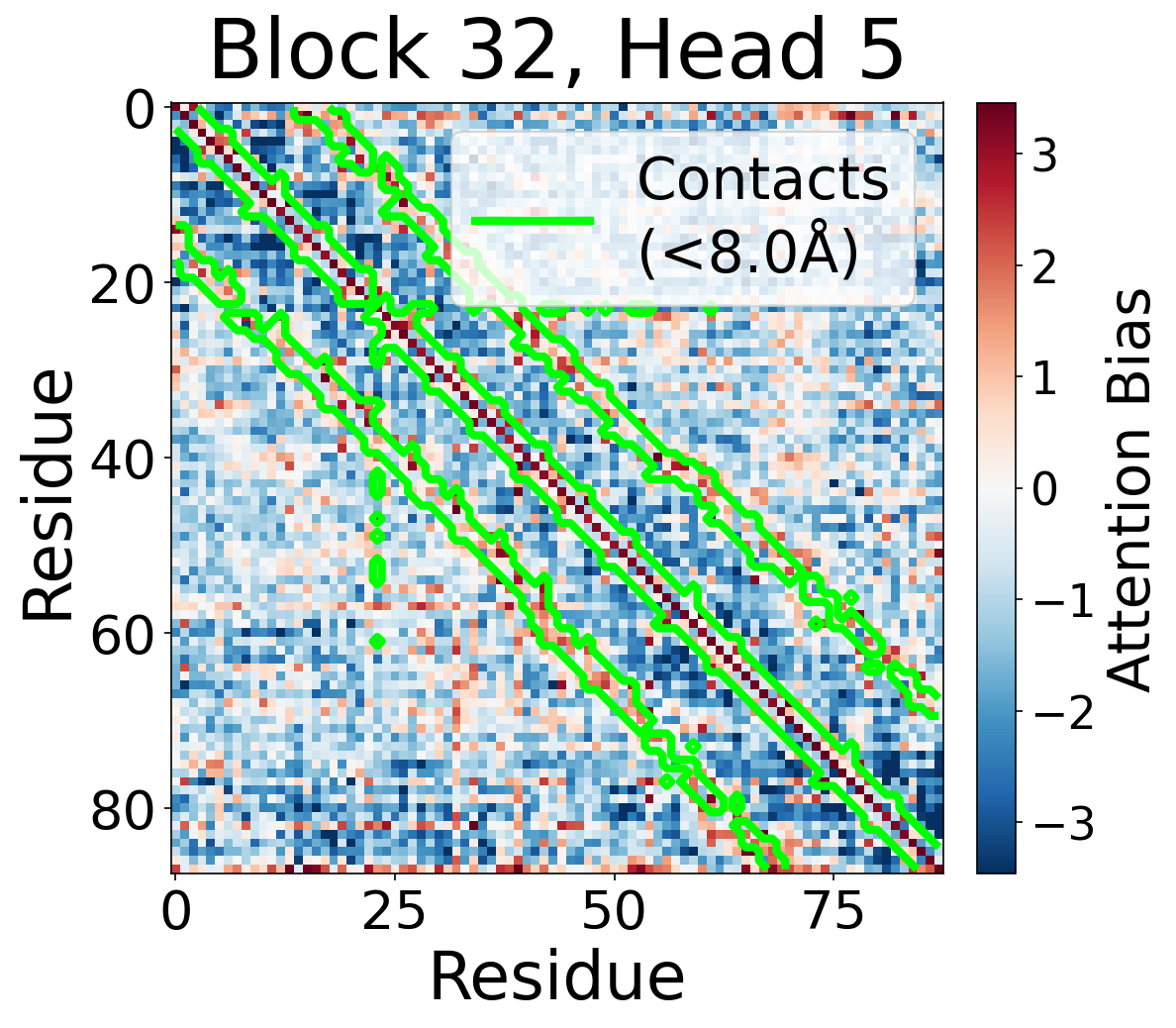}
    \end{subfigure}
    \hfill
    \begin{subfigure}[b]{0.24\textwidth}
        \includegraphics[width=\textwidth]{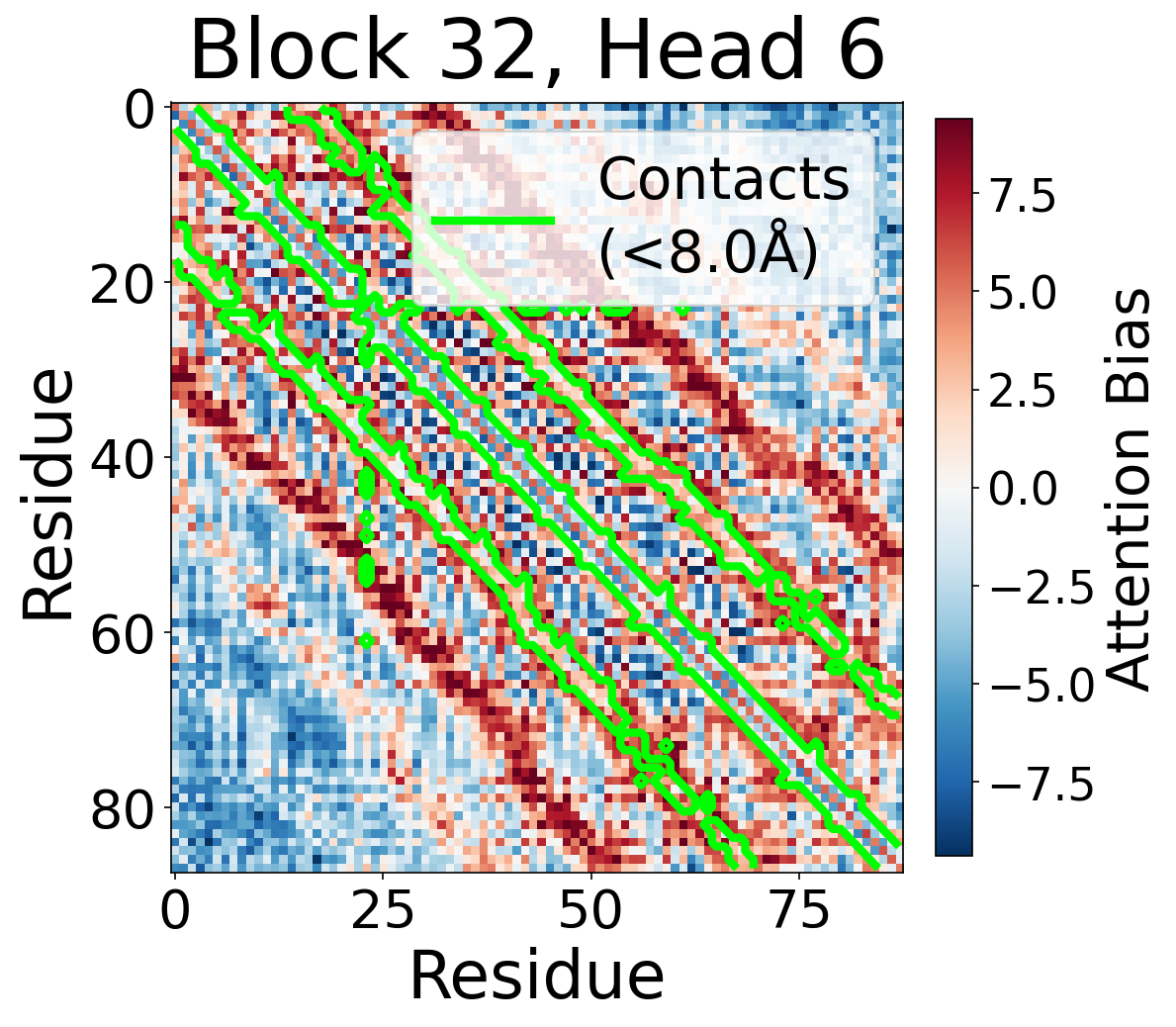}
    \end{subfigure}
    \hfill
    \begin{subfigure}[b]{0.24\textwidth}
        \includegraphics[width=\textwidth]{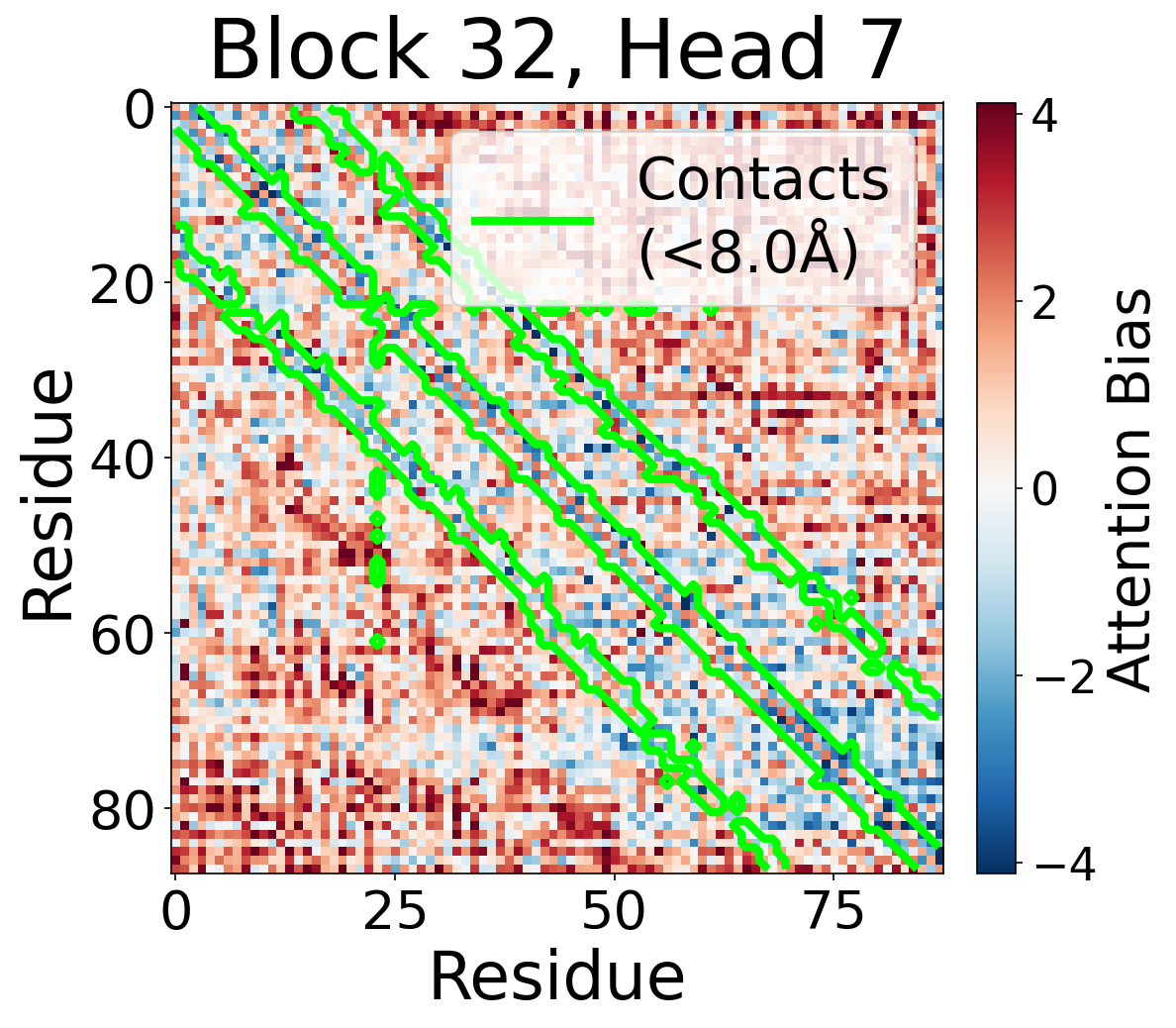}
    \end{subfigure}

    \caption{\textbf{Per-head \texttt{pair2seq} bias at block 32.}
    Individual attention head bias values for beta protein 1l0s at block 32. Green contours mark structural contacts (C$\alpha$ $<$ 8\AA). Different heads exhibit distinct patterns: some heads show strong contact-aligned bias (e.g., heads that highlight the off-diagonal contact structure), while others capture different spatial relationships or show more diffuse patterns. The different bias patterns suggest possible head specialization; testing its effect on attention and structure requires further intervention.}
    \label{fig:bias_maps_heads_alpha}
\end{figure*}

\begin{figure*}
    \includegraphics[width=1.0\linewidth]{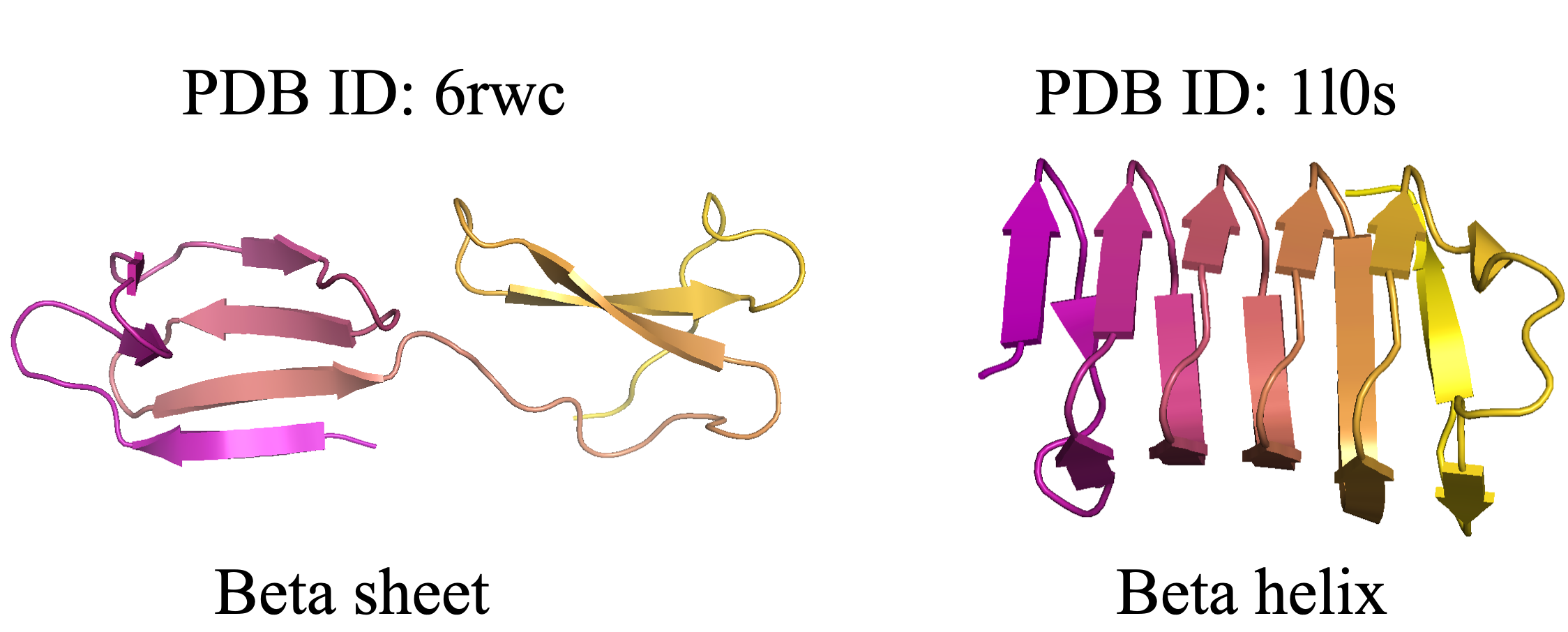}
    \caption{\textbf{Example proteins used throughout the analysis.}
PyMOL visualization of PDB 6rwc, a beta-sheet protein with antiparallel strands
connected by turns, and PDB: 1l0s, a left-handed beta-helix with repeating
triangular coils. These two structures illustrate distinct
beta-rich folds and serve as representative examples for the
pair2seq bias visualizations in Appendix \ref{app:bias_maps}.}
\end{figure*}

\section{All-Pairs Cross-Model Alignment and Patching}
\label{app:all_pairs_alignment}

\begin{figure}[h]
    \centering
    \includegraphics[width=\linewidth]{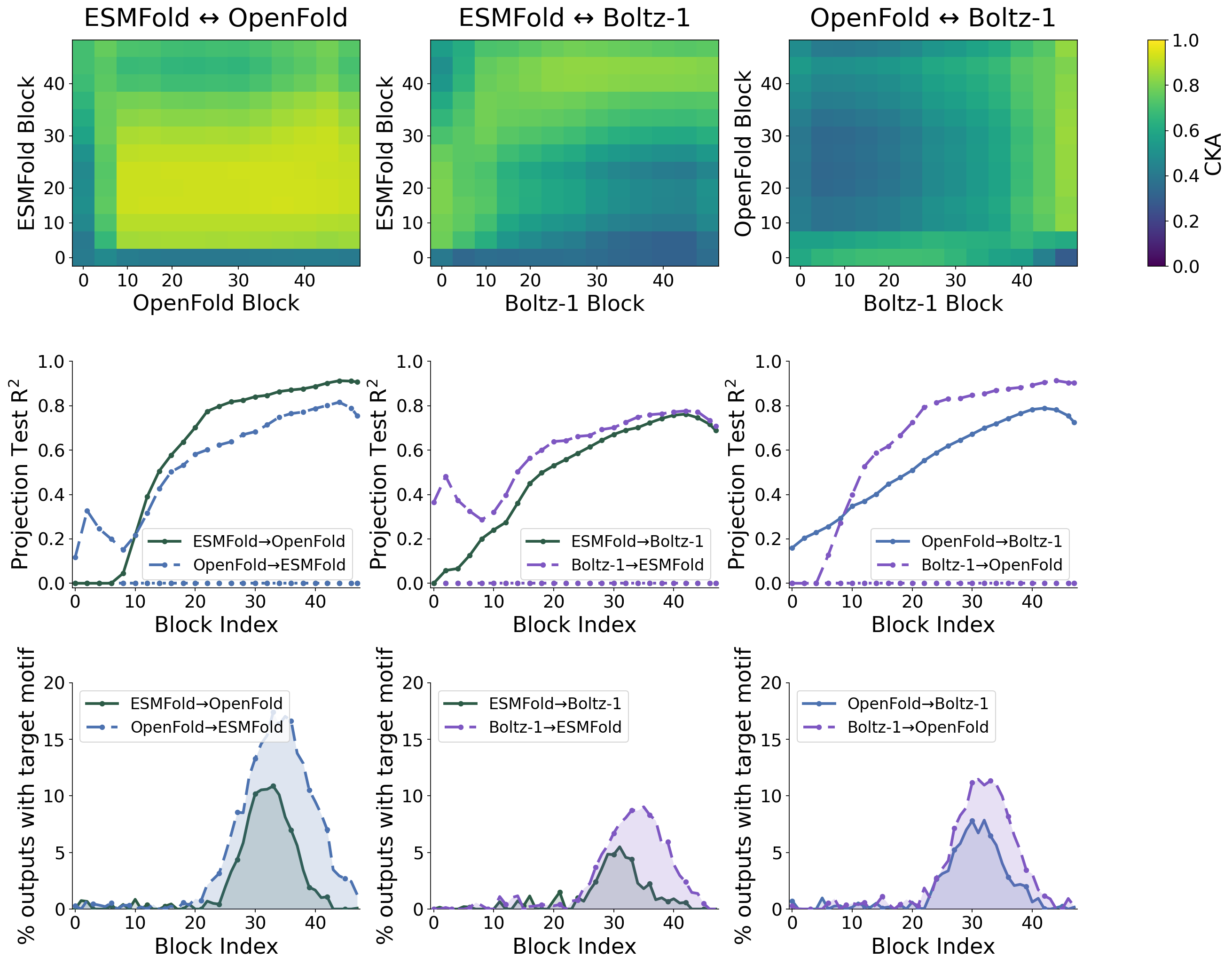}
    \caption{\textbf{Pairwise representation alignment and cross-model patching across all three model pairings.} Columns correspond to the three pairings: ESMFold $\leftrightarrow$ OpenFold (left), ESMFold $\leftrightarrow$ Boltz-1 (middle), OpenFold $\leftrightarrow$ Boltz-1 (right).
    \textbf{Top row:} CKA between pairwise representations at every block pair. Similarity is high throughout middle and late blocks for all three pairings, concentrating along the diagonal. ESMFold $\leftrightarrow$ OpenFold shows the highest overall similarity; pairings involving Boltz-1 are lower but still substantial.
    \textbf{Middle row:} Test $R^2$ of whitened Procrustes projections in both directions of each pairing, evaluated per block. Solid and dashed lines correspond to the two directions; dotted lines at the bottom of each panel are shuffled-correspondence baselines, which sit near zero across all blocks.
    \textbf{Bottom row:} Single-block pairwise patching success when donor representations come from one model (after Procrustes projection into the receiving model's space) and are patched into the other. Cross-model patching reproduces the late-block pairwise window in all six directions, with patching strength tracking projection quality from the middle row.}
    \label{fig:all_pairs_alignment}
\end{figure}

In §\ref{sec:cross-model}, we showed that pairwise representations from OpenFold and Boltz-1 can be linearly aligned to ESMFold's space and substituted into ESMFold at the corresponding stage, recovering the late-block patching window. Here we extend this analysis to all three pairings of folding architectures (ESMFold $\leftrightarrow$ OpenFold, ESMFold $\leftrightarrow$ Boltz-1, OpenFold $\leftrightarrow$ Boltz-1) and to both directions of each pairing. Results are summarized in Fig.~\ref{fig:all_pairs_alignment}.

\paragraph{CKA across all pairings.} The top row of Fig.~\ref{fig:all_pairs_alignment} extends the CKA analysis from §\ref{sec:cross-model} to the OpenFold $\leftrightarrow$ Boltz-1 pairing. All three pairings show high CKA throughout middle and late blocks: late blocks of one model are most similar to late blocks of another. ESMFold $\leftrightarrow$ OpenFold shows the highest overall similarity (CKA $> 0.8$ throughout the middle of the trunk), consistent with their shared lineage from the AlphaFold2 architecture. Pairings involving Boltz-1 show lower CKA (typically 0.5--0.8), showing shared linear structure despite the larger architectural differences involving Boltz-1.

\paragraph{Procrustes alignment in both directions.} The middle row reports per-block test $R^2$ of whitened Procrustes projections in both directions of each pairing. Alignment is poor at the earliest measured block, but rises rapidly through blocks 5--15 and plateaus thereafter. Plateau values are highest for ESMFold $\leftrightarrow$ OpenFold ($R^2 \approx 0.9$ in both directions), intermediate for OpenFold $\leftrightarrow$ Boltz-1 ($R^2 \approx 0.8$--$0.9$), and lowest for ESMFold $\leftrightarrow$ Boltz-1 ($R^2 \approx 0.7$--$0.8$). The two directions of each pairing track each other closely. Shuffled-correspondence baselines (dotted lines) perform poorly relative to true correspondences; their absolute values depend on the alignment and centering convention (App.~\ref{app:projection-comparison}).

A few panels show small positive $R^2$ at very early blocks (e.g., OpenFold $\to$ ESMFold and Boltz-1 $\to$ ESMFold near block 0--5). We attribute these to coincidental alignment of model-specific positional embedding structure that is recoverable by a linear map. These early-block alignments do not translate into successful patching (bottom row), consistent with the interpretation that they reflect surface-level structure rather than the shared computational geometry that develops in middle and late blocks.

\paragraph{Cross-model patching across all directions.} The bottom row applies the cross-model patching protocol from §\ref{sec:cross-model} to all six donor--receiver directions. In every direction, patching reproduces the late-block pairwise window observed in within-model patching: cross-model patches induce the target motif only when applied to blocks 25--40 of the receiving model, matching the pairwise stage identified in §\ref{sec:patching_setup} and App.~\ref{app:triple_model_info_flow}. Patching strength tracks alignment quality in the middle row. The highest cross-model patching rates occur for ESMFold $\leftrightarrow$ OpenFold (peaking near 17\% for OpenFold $\to$ ESMFold and 11\% for ESMFold $\to$ OpenFold). Pairings involving Boltz-1 show lower patching rates (typically 5--12\%), consistent with their lower projection $R^2$, but the late-block timing window is preserved in all six cases.

\paragraph{Asymmetry between directions.} For each pairing, the two directions of patching produce substantially different success rates even when projection $R^2$ is similar in both directions. We do not have a definitive explanation for this asymmetry, but plausible factors include differences in the receiving model's tolerance for off-distribution pairwise inputs, differences in how each model's downstream pathways read $z$, and small numerical differences in how the projection is conditioned. Matched calibration and decoder swaps (App.~\ref{app:revision_transfer}) further characterize this asymmetry, while leaving its full cause unresolved.

\paragraph{Summary.} Across all three pairings of folding architectures and all six donor--receiver directions, pairwise representations are linearly alignable, and aligned representations are partially functionally transplantable after linear alignment at the corresponding stage of computation. Combined with the staged structure (App.~\ref{app:triple_model_info_flow}) and stage-specific causal features (charge in §\ref{sec:biochem-features}, distance in §\ref{sec:distance_steering}) replicating across models, this supports the interpretation that folding trunks converge on a shared representational organization for the late-block pairwise stage, despite differences in training data, architecture, and input modality.

\subsection{Comparison of Linear Alignment Methods}
\label{app:projection-comparison}

In §\ref{sec:cross-model}, we used whitened Procrustes alignment to learn linear projections between models' pairwise representations. Here we compare against four alternatives under a shuffled-correspondence control.

\mypar{Methods} Ridge regression (unconstrained linear with L2), CCA, unwhitened Procrustes (orthogonal rotation), whitened Procrustes (the method used in §\ref{sec:cross-model}), and random orthogonal as a floor baseline.

\mypar{Shuffled control} For each method, we randomly shuffle residue-pair correspondences between $Z_A$ and $Z_B$ before alignment, then evaluate test $R^2$. A method exploiting genuine shared structure should perform poorly under shuffling; a method fitting model-specific artifacts (marginal statistics, dimensionality) can still achieve nontrivial $R^2$.

\begin{figure}[h]
    \centering
    \includegraphics[width=\linewidth]{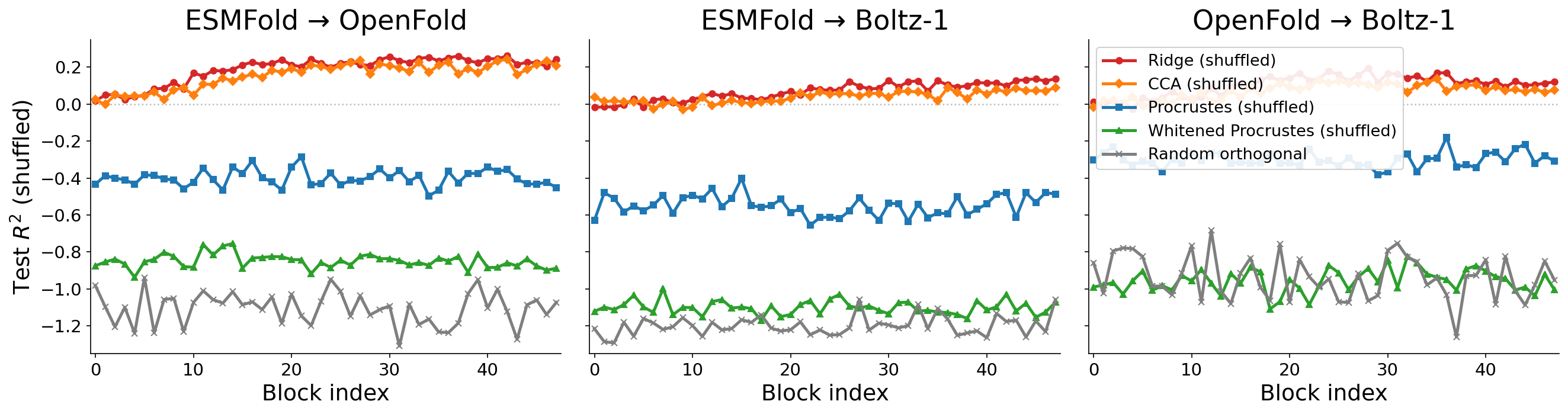}
    \caption{\textbf{Test $R^2$ under shuffled correspondences for five alignment methods.} Ridge and CCA achieve mildly positive shuffled $R^2$, indicating they fit marginal statistics. Procrustes-family methods produce negative $R^2$ under shuffling because the rotation constraint cannot recover the target mean; the more constrained the method, the more negative the shuffled $R^2$ can go. Whitened Procrustes sits near the random-orthogonal floor, indicating its rotation contains essentially no information transferable to shuffled data.}
    \label{fig:projection_shuffled}
\end{figure}

Ridge and CCA achieve shuffled $R^2$ around 0.1--0.25 across blocks. Whitened Procrustes sits near the random-orthogonal floor ($R^2 \approx -1$), with unwhitened Procrustes intermediate. Combined with the high true-correspondence $R^2$ reported in §\ref{sec:cross-model} ($R^2 \approx 0.9$ for ESMFold $\leftrightarrow$ OpenFold), this is why we use whitened Procrustes: it is the most constrained method that still aligns representations with high fidelity, and its near-floor shuffled control indicates the learned rotation is specific to genuine cross-model structure rather than model-specific artifacts.

\input{revision_evidence}

\clearpage
\input{checklist_camera_ready}

\end{document}

%% file: Figures/seq_example.tex

\begin{figure}[!h]
\centering
\begin{tikzpicture}[baseline=(pre.base)]
  \definecolor{strandCol}{RGB}{33,102,172} 
  \definecolor{loopCol}{RGB}{171,39,98}    
  \definecolor{braceS1}{RGB}{33,102,172}
  \definecolor{braceS2}{RGB}{0,109,44}
  \definecolor{braceLp}{RGB}{171,39,98}

  \node[inner sep=0pt, outer sep=0pt, anchor=base west] (pre) at (0,0)
    {{\ldots A\,Q\,T\,V\,D\,}};

  \node[inner sep=0pt, outer sep=0pt, anchor=base west] (s1) at (pre.base east)
    {{\textcolor{strandCol}{S\,W\,T\,W\,E\,}}};

  \node[inner sep=0pt, outer sep=0pt, anchor=base west] (lp) at (s1.base east)
    {{\textcolor{loopCol}{N\,G\,K\,W\,}}};

  \node[inner sep=0pt, outer sep=0pt, anchor=base west] (s2) at (lp.base east)
    {{\textcolor{strandCol}{T\,W\,K\,}}};

  \node[inner sep=0pt, outer sep=0pt, anchor=base west] (post) at (s2.base east)
    {{E\,Q\,L\,N\,S\ldots}};

  \draw[decorate,decoration={brace,amplitude=5pt},yshift=-30pt,
        line width=0.5pt, color=braceS1]
    (s1.south east) -- (s1.south west)
    node[midway,below=7pt, text=braceS1] {\scriptsize Strand 1};

  \draw[decorate,decoration={brace,amplitude=5pt},yshift=-12pt,
        line width=0.5pt, color=braceLp]
    (lp.south east) -- (lp.south west)
    node[midway,below=7pt, text=braceLp] {\scriptsize Loop};

  \draw[decorate,decoration={brace,amplitude=5pt},yshift=-12pt,
        line width=0.5pt, color=braceS1]
    (s2.south east) -- (s2.south west)
    node[midway,below=7pt, text=braceS1] {\scriptsize Strand 2};

\end{tikzpicture}
\caption{\textbf{Sequence view of a $\beta$-hairpin (highlighted) within a protein}: two $\beta$-strand segments separated by a loop. Colored braces label the strand--loop--strand decomposition.}
\label{fig:1d-sequence}
\end{figure}

%% file: revision_evidence.tex
\section{Scope, Conditioning, and Extended Interventions}
\label{app:revision_scope}

\paragraph{All-pair evaluation.}
The main single-block analysis uses donor--target pairs for which full patching succeeds. Table~\ref{tab:all_pairs_revision} compares that subset with all donor--target pairs. Each entry is the maximum per-block induction rate within the specified early (0--11) or late (36--47) region, rather than the maximum across the entire trunk. These regional peaks therefore need not equal the global peaks in other tables. Including all pairs lowers success rates. Sequence patches still act early in ESMFold and OpenFold, while pairwise patches remain more effective in Boltz-1. In Boltz-1, pairwise patches act over a broad interval rather than only in late blocks.

\begin{table}[htbp]
\centering\small
\caption{Single-block induction rates (\%) with and without conditioning on full-patch success. S: sequence; P: pairwise.}
\label{tab:all_pairs_revision}
\begin{tabular}{llrrrr}
\toprule
Model & Cases & S early & S late & P early & P late\\
\midrule
ESMFold & Transferable subset & 40.0 & 0.0 & 0.2 & 25.0\\
 & All pairs & 36.8 & 0.0 & 0.1 & 15.1\\
OpenFold & Transferable subset & 40.0 & 0.0 & 0.2 & 34.2\\
 & All pairs & 33.4 & 0.0 & 0.2 & 29.6\\
Boltz-1 & Transferable subset & 8.7 & 0.0 & 30.3 & 35.4\\
 & All pairs & 5.7 & 0.0 & 25.6 & 27.5\\
\bottomrule
\end{tabular}
\end{table}

\paragraph{Larger motifs and tertiary contacts.}
For four-strand sheets, we use 200 PISCES donor proteins and 100 size-matched targets of 400--600 residues. Donor sheets span 40--60 residues and are patched into non-beta target regions. We report intended-topology formation separately from DSSP beta content: increased beta content alone does not establish the intended sheet. For tertiary contacts, we use 300 donor--target pairs with contacting donor strands and non-contacting target segments. The median sequence separation is 132 residues (range 59--254), and the mean initial spatial separation is 24.5\AA. Sequence patches replace the selected segment states; pairwise patches replace cross-segment states. Contact success requires the intended C$\alpha$ pair to lie within 8\AA. Table~\ref{tab:extended_motifs_revision} reports both tasks, with matched contact controls in Tables~\ref{tab:contact_controls_distance}--\ref{tab:contact_controls_specificity}.

\begin{table}[htbp]
\centering\small
\caption{Extended interventions. Entries separated by a slash are sequence/pairwise results. Sheet beta-content increases and strict topology success are distinct outcomes.}
\label{tab:extended_motifs_revision}
\begin{tabular}{llccc}
\toprule
Task & Model & Effective blocks & Peak effect & Success (\%)\\
\midrule
Sheet & ESMFold & 0--6 / 21--42 & $+72\% / +65\%$ beta & 32 / 21\\
 & OpenFold & 0--4 / 19--45 & $+74\% / +81\%$ beta & 37 / 25\\
 & Boltz-1 & 0--3 / 7--47 & $+52\% / +83\%$ beta & 15 / 28\\
\midrule
Contact & ESMFold & 0--11 / 24--47 & $-15.5 / -17.3$\AA & 47 / 46\\
 & OpenFold & 0--15 / 22--47 & $-15.2 / -17.7$\AA & 55 / 52\\
 & Boltz-1 & 0--2 / 10--47 & $-16.0 / -18.5$\AA & 12 / 73\\
\bottomrule
\end{tabular}
\end{table}

\paragraph{Length and motif-independent timing.}
The original target proteins contain 100--400 residues. A length-stratified extension uses five bins with 60, 60, 40, 30, and 20 proteins, respectively, up to 1,000 residues. Comparing the shortest (75--250) and longest (801--1,000) bins, full-patch success declines from 43\% to 28\% for ESMFold, 41\% to 26\% for OpenFold, and 38\% to 25\% for Boltz-1. ESMFold's sequence window changes from 0--7 to 0--9 and its pairwise window from 27--37 to 25--35; OpenFold has the same reported windows. Their longest-bin sequence/pairwise peaks are 16.2\%/3.6\% and 11.3\%/11.8\%, respectively. Boltz-1 has no detected sequence-patching effect in this extension; its pairwise effects occupy 0--3 and 24--36 in the shortest bin and 0--3 and 23--34 in the longest, with a longest-bin peak of 18.2\%. Thus, longer proteins retain some intervention timing but provide weaker evidence for sequence sufficiency, especially in Boltz-1.

Separately, pathway-update norms measured across all residues of 1,000 diverse proteins (75--1,000 residues, including repeat, coiled-coil, and multidomain proteins) have sequence-to-pair peak/center-of-mass block indices of 2/6.4, 2/5.2, and 1/1.1 for ESMFold, OpenFold, and Boltz-1. Corresponding pair-to-sequence indices are 34/25.6, 32/26.3, and 11/22.5. This motif-independent measurement supports earlier sequence-to-pair updates without establishing causal use from update magnitude alone.

\paragraph{Boltz-1 indexing and interpretation.}
The upstream MSA module is indexed as blocks 0--3 in this analysis. After filtering, Boltz-1 pairwise patches transfer hairpins from blocks 3--47. The effect begins near the MSA--Pairformer boundary and continues through the subsequent pairwise processing. Sequence-to-pair freezing and ablation, donor-directed changes in $z$, and early biochemical steering provide evidence for functional sequence-to-pair transfer in the upstream module. Sequence patches produce the donor structure less often than in ESMFold or OpenFold. Thus, the models can perform related computations in the same order while differing in the blocks involved and the success of an intervention.

\section{Structural Quality, Designability, and Contact Specificity}
\label{app:revision_quality}

\paragraph{Evaluation criteria.}
A chain is intact when no adjacent C$\alpha$--C$\alpha$ distance exceeds 4.2\AA. A structure is clash-free when it has at most two non-adjacent heavy-atom pairs within 2\AA. Model confidence is reported as mean pLDDT on a 0--1 scale. ProteinMPNN~\citep{dauparas2022robust} designs eight sequences per backbone; the best-of-eight design/refold self-consistency TM-score is denoted scTM, with scTM $\geq 0.5$ used for the designability filter. The cross-model check refolds designed sequences with the two non-source folding models and reports their mean scTM. Motif recovery is conditional on successful patches and measures recovery within the intervention window. These checks ask whether the output has plausible geometry and whether folding models can recover it from designed sequences. They do not establish experimental stability.

\begin{table}[htbp]
\centering\footnotesize
\setlength{\tabcolsep}{4pt}
\caption{Independent-refolding evaluation, $n=2,000$ outputs per model/condition. Rates are percentages; brackets give reported 95\% intervals. ``Cross'' is cross-model scTM. All-block patching is more disruptive than a single-block intervention.}
\label{tab:quality_independent_revision}
\begin{tabular}{llcccc}
\toprule
Model & Condition & Success & Intact & Clash-free & Cross\\
\midrule
ESMFold & Baseline & 0 [0.0, 0.2] & 100 [99.8, 100] & 93 [91.8, 94.0] & .97 [.967, .973]\\
 & Single-block & 40 [37.9, 42.2] & 97 [96.2, 97.7] & 80 [78.2, 81.7] & .93 [.925, .935]\\
 & All-block & 42 [39.9, 44.2] & 83 [81.3, 84.6] & 63 [60.9, 65.1] & .91 [.904, .916]\\
OpenFold & Baseline & 0 [0.0, 0.2] & 97 [96.2, 97.7] & 80 [78.2, 81.7] & .96 [.956, .964]\\
 & Single-block & 40 [37.9, 42.2] & 90 [88.6, 91.2] & 63 [60.9, 65.1] & .94 [.935, .945]\\
 & All-block & 38 [35.9, 40.1] & 67 [64.9, 69.0] & 50 [47.8, 52.2] & .90 [.893, .907]\\
Boltz-1 & Baseline & 0 [0.0, 0.2] & 100 [99.8, 100] & 87 [85.5, 88.4] & .98 [.977, .983]\\
 & Single-block & 40 [37.9, 42.2] & 100 [99.8, 100] & 83 [81.3, 84.6] & .96 [.956, .964]\\
 & All-block & 41 [38.9, 43.2] & 100 [99.8, 100] & 70 [68.0, 72.0] & .95 [.945, .955]\\
\bottomrule
\end{tabular}

\medskip
\begin{tabular}{llccc}
\toprule
Model & Condition & pLDDT & scTM & Motif recovery (\%)\\
\midrule
ESMFold & Baseline & .92 [.916, .924] & .99 [.988, .992] & --\\
 & Single-block & .87 [.865, .875] & .95 [.946, .954] & 92 [89.9, 93.7]\\
 & All-block & .82 [.814, .826] & .93 [.925, .935] & 100 [99.5, 100]\\
OpenFold & Baseline & .85 [.844, .856] & .98 [.977, .983] & --\\
 & Single-block & .90 [.895, .905] & .96 [.956, .964] & 94 [92.1, 95.4]\\
 & All-block & .81 [.803, .817] & .94 [.935, .945] & 100 [99.5, 100]\\
Boltz-1 & Baseline & .92 [.916, .924] & .99 [.988, .992] & --\\
 & Single-block & .84 [.834, .846] & .98 [.977, .983] & 96 [94.4, 97.2]\\
 & All-block & .83 [.824, .836] & .97 [.966, .974] & 100 [99.5, 100]\\
\bottomrule
\end{tabular}
\end{table}

Rate intervals in Table~\ref{tab:quality_independent_revision} are Wilson 95\% intervals; continuous-metric intervals are obtained by resampling independent protein--target cases. Table~\ref{tab:steering_quality_revision} gives the separate steering and cross-model-transplant evaluations. Sample sizes and interventions differ, so its rates should not be interpreted as a matched comparison with the single-block cohort above.

\begin{table}[htbp]
\centering\small
\setlength{\tabcolsep}{4pt}
\caption{Quality of steering and transplantation outputs. Success is shown with 95\% interval half-widths. Intact, clash-free, and recovery are percentages; pLDDT and scTM are means. Recovery is conditional on successful interventions.}
\label{tab:steering_quality_revision}
\begin{tabular}{llrcccccc}
\toprule
Model & Intervention & $N$ & Success (\%) & Intact & Clash-free & pLDDT & scTM & Recovery\\
\midrule
ESMFold & Charge & 400 & $20\pm3.9$ & 72 & 68 & .83 & .75 & 77\\
 & Distance & 600 & $18\pm3.1$ & 67 & 72 & .85 & .78 & 89\\
 & Cross-model & 4,000 & $16\pm1.1$ & 83 & 75 & .78 & .95 & 85\\
OpenFold & Charge & 400 & $20\pm3.9$ & 77 & 65 & .74 & .90 & 75\\
 & Distance & 600 & $17\pm3.0$ & 80 & 68 & .72 & .90 & 72\\
 & Cross-model & 4,000 & $14\pm1.1$ & 91 & 72 & .81 & .96 & 100\\
Boltz-1 & Charge & 400 & $8\pm2.7$ & 95 & 83 & .82 & .84 & 62\\
 & Distance & 600 & $16\pm2.9$ & 92 & 75 & .79 & .93 & 87\\
 & Cross-model & 4,000 & $9\pm0.9$ & 93 & 91 & .84 & .97 & 91\\
\bottomrule
\end{tabular}
\end{table}

\begin{table}[htbp]
\centering\small
\caption{Localization after structural-validity and designability filtering. Peaks are induction rates (\%). The Boltz-1 pairwise interval is 3--47. Each interval applies to the stated intervention and mask.}
\label{tab:filtered_windows_revision}
\begin{tabular}{llrrrr}
\toprule
Model & Patch & Raw peak & Validity-filtered & Designability-filtered & Window\\
\midrule
ESMFold & Sequence & 40 & 37 & 35 & 0--6\\
 & Pairwise & 25 & 21 & 22 & 19--42\\
OpenFold & Sequence & 41 & 39 & 34 & 0--7\\
 & Pairwise & 34 & 26 & 24 & 18--43\\
Boltz-1 & Sequence & 15 & 10 & 7 & 0--3\\
 & Pairwise & 35 & 33 & 33 & 3--47\\
\bottomrule
\end{tabular}
\end{table}

\paragraph{Targeted contact transfer versus collapse.}
The long-range experiment is repeated with a donor whose corresponding segments do not contact, and with a random pairwise perturbation matched to the donor patch's norm and mask. Tables~\ref{tab:contact_controls_distance} and \ref{tab:contact_controls_specificity} report the same conditions. S and P denote sequence and pairwise patching. Donor-contact precision is the fraction of newly formed cross-segment contacts present in the donor. Off-target density measures tertiary contact-density change outside the targeted cross-segment pairs. ``Contacting'' denotes the original donor condition; ``non-contacting'' denotes the matched donor control. Continuous quantities are shown with 95\% interval half-widths; rates use Wilson intervals.

\begin{table}[htbp]
\centering\footnotesize
\setlength{\tabcolsep}{4pt}
\caption{Long-range contact controls: targeted and global distance changes. Rg is radius of gyration; all distances are in \AA.}
\label{tab:contact_controls_distance}
\begin{tabular}{llrrrr}
\toprule
Model & Condition & $\Delta d_{\rm intended}$ & Contact success (\%) & $\Delta$Rg & $\Delta d_{\rm chain}$\\
\midrule
ESMFold & Contacting S & $-15.51\pm5.13$ & $47.7\pm5.8$ & $-.65\pm.74$ & $-.46\pm.21$\\
 & Contacting P & $-17.32\pm3.56$ & $46.1\pm4.9$ & $-.15\pm.17$ & $-.14\pm.19$\\
 & Non-contacting S & $6.72\pm1.54$ & 0 [0, 3] & $1.87\pm.50$ & $2.65\pm.55$\\
 & Non-contacting P & $5.28\pm1.31$ & 0 [0, 2] & $1.01\pm.66$ & $2.01\pm1.06$\\
 & Random P & $-5.63\pm2.82$ & $7.8\pm4.3$ & $-3.19\pm.23$ & $-2.21\pm.28$\\
OpenFold & Contacting S & $-15.20\pm3.17$ & $55.5\pm4.2$ & $.22\pm.27$ & $.35\pm.33$\\
 & Contacting P & $-17.70\pm3.28$ & $52.3\pm6.6$ & $-.41\pm.24$ & $-.05\pm.28$\\
 & Non-contacting S & $4.20\pm.96$ & 0 [0, 6] & $.51\pm.27$ & $.69\pm.35$\\
 & Non-contacting P & $4.36\pm1.02$ & 0 [0, 7] & $1.05\pm.19$ & $1.06\pm.24$\\
 & Random P & $-8.79\pm3.14$ & $9.2\pm3.1$ & $-2.44\pm2.21$ & $-1.79\pm2.84$\\
Boltz-1 & Contacting S & $-16.04\pm4.47$ & $12.4\pm4.7$ & $-.42\pm.14$ & $-.12\pm.05$\\
 & Contacting P & $-18.53\pm5.85$ & $73.2\pm7.3$ & $-.22\pm.11$ & $-.27\pm.12$\\
 & Non-contacting S & $3.04\pm.04$ & 0 [0, 3] & $1.05\pm.13$ & $1.07\pm.13$\\
 & Non-contacting P & $3.03\pm1.92$ & 0 [0, 4] & $2.31\pm.97$ & $1.21\pm.57$\\
 & Random P & $-6.03\pm1.23$ & $6.0\pm2.4$ & $-3.24\pm.07$ & $-1.32\pm1.08$\\
\bottomrule
\end{tabular}
\end{table}

\begin{table}[htbp]
\centering\small
\caption{Long-range contact controls: specificity for the donor pattern and off-target changes. Density changes are in percentage points (pp). Rows correspond to Table~\ref{tab:contact_controls_distance}.}
\label{tab:contact_controls_specificity}
\begin{tabular}{llrr}
\toprule
Model & Condition & Donor-contact precision & $\Delta$ off-target density (pp)\\
\midrule
ESMFold & Contacting S & $.92\pm.06$ & $.3\pm.2$\\
 & Contacting P & $.87\pm.08$ & $.1\pm.1$\\
 & Non-contacting S & $.00\pm.00$ & $-1.2\pm.4$\\
 & Non-contacting P & $.00\pm.00$ & $-.9\pm.5$\\
 & Random P & $.12\pm.04$ & $4.1\pm.9$\\
OpenFold & Contacting S & $.88\pm.07$ & $-.1\pm.2$\\
 & Contacting P & $.79\pm.13$ & $.1\pm.2$\\
 & Non-contacting S & $.00\pm.00$ & $-.4\pm.3$\\
 & Non-contacting P & $.00\pm.00$ & $-.5\pm.3$\\
 & Random P & $.13\pm.09$ & $-.8\pm.8$\\
Boltz-1 & Contacting S & $.76\pm.12$ & $.4\pm.2$\\
 & Contacting P & $.83\pm.22$ & $.2\pm.2$\\
 & Non-contacting S & $.00\pm.00$ & $-.6\pm.2$\\
 & Non-contacting P & $.00\pm.00$ & $-.7\pm.4$\\
 & Random P & $.19\pm.09$ & $.8\pm.3$\\
\bottomrule
\end{tabular}
\end{table}

Contacting donors recover the donor contact pattern with precision .76--.92 and modest whole-chain changes ($|\Delta\mathrm{Rg}|\leq .65$\AA). Random perturbations produce lower contact success and precision, with larger whole-chain contraction. Non-contacting donors increase segment separation. The donor patches therefore do more than pull the protein together: they preferentially recover the intended contacts. Local distortions caused by the intervention can still occur.

\section{Biochemical Steering Controls}
\label{app:revision_biochem}

\paragraph{Charge controls.}
Norm- and variance-matched random directions use the same steering strengths, residue positions, and block windows as the charge intervention. Charge-steering peaks are 20\%, 20\%, and 8\% for ESMFold, OpenFold, and Boltz-1, whereas random directions have median peak effects at most .2\% and 95th-percentile effects below 1\% across models. Together with sign-sensitive attraction and repulsion, this argues against arbitrary perturbation as the sole explanation. The direction can nevertheless encode charge together with correlated residue properties.

\paragraph{An uncharged-residue hydrophobicity construction.}
To separate hydrophobicity from charged-residue labels, we construct a difference-of-means direction from two sets containing only uncharged residues: $\{\mathrm{I,V,L,F,M,A}\}$ and $\{\mathrm{N,Q,S,T,G}\}$. On 500 helical targets, we steer the originally hydrophobic face toward the polar direction and the originally polar face toward the hydrophobic direction, then measure DSSP-derived solvent exposure. Table~\ref{tab:uncharged_hydropathy_revision} gives the results. Late-block and matched-random interventions do not produce the systematic exposure reversal. The effect persists without charged residues in either label set, supporting a hydrophobicity-related feature. Other properties of these residues may also contribute.

\begin{table}[htbp]
\centering\small
\caption{Steering from uncharged residue sets, $N=500$ helical targets. Exposure rates are percentages. Parentheses give change in percentage points with reported 95\% interval half-widths.}
\label{tab:uncharged_hydropathy_revision}
\begin{tabular}{lccc}
\toprule
Model & Effective blocks & Originally hydrophobic face & Originally polar face\\
\midrule
ESMFold & 0--18 & $23\to66$ ($+43\pm6.1$) & $49\to26$ ($-23\pm2.3$)\\
OpenFold & 0--14 & $23\to62$ ($+39\pm5.6$) & $49\to22$ ($-27\pm3.7$)\\
Boltz-1 & 0--3 & $23\to47$ ($+24\pm4.4$) & $49\to17$ ($-32\pm4.9$)\\
\bottomrule
\end{tabular}
\end{table}

\section{Coherent Scale Sensitivity and Directional Geometry Controls}
\label{app:revision_geometry}

Multiplying the entire pairwise input by a scalar tests the effect of its overall scale; it does not isolate one feature direction. In a magnitude-matched comparison, coherent rescaling changes mean output pairwise distance by 6.89\AA, versus .05\AA\ for Gaussian noise. The decoder therefore responds differently to uniform scaling and to noise of the same magnitude.

The learned distance direction is tested separately on the cross-strand pairs targeted in distance steering. In the reverse intervention, steering toward greater separation changes targeted distances by $+10.86$\AA\ averaged across the three models, compared with $+.20$\AA\ for a matched random direction. Preserving the global norm of $z$ gives $+10.85$\AA. The distance change is therefore not explained solely by an increase in the overall norm, although individual pair vectors can still change in norm. Separately, contact readouts from the attention bias show that attention has access to contact information; they do not establish how much the output depends on its use.

\section{Calibration, Decoder Swaps, and Disagreement Analysis}
\label{app:revision_transfer}

\paragraph{Matched intermediate-state calibration.}
Each off-diagonal donor--receiver direction is evaluated on approximately 2,000 matched protein--target cases across the 48 blocks. We divide peak cross-model success by the receiver's peak within-model success, using matched cases, receivers, and masks. Bootstrap intervals resample independent matched cases and recompute the peak ratio within each replicate; block evaluations are not treated as independent samples. Table~\ref{tab:transfer_ratios_revision} shows that success depends strongly on the receiving model. Patches from the other models recover approximately 64\% of ESMFold's within-model peak, 35\% of OpenFold's, and 22.5\% of Boltz-1's. The mean of the six cross-model ratios is 40.3\%.

\begin{table}[htbp]
\centering\small
\caption{Peak cross-model/within-model success ratios with case-bootstrap 95\% intervals. Rows are donor models; columns are receivers. Diagonals are normalized references.}
\label{tab:transfer_ratios_revision}
\begin{tabular}{lccc}
\toprule
Donor & ESMFold & OpenFold & Boltz-1\\
\midrule
ESMFold & 1.00 & .33 [.29, .37] & .20 [.17, .23]\\
OpenFold & .75 [.69, .81] & 1.00 & .25 [.21, .29]\\
Boltz-1 & .52 [.47, .57] & .37 [.33, .42] & 1.00\\
\bottomrule
\end{tabular}
\end{table}

\paragraph{Transferring final states between decoders.}
For each source--decoder combination, we use 500 cases in which the source intervention produces a hairpin. For each case, we finish the source trunk, align its final $z$ to the receiver basis, and pass it to the receiving decoder. This bypasses subsequent receiver-trunk computation. Table~\ref{tab:decoder_swaps_revision} reports the median C$\alpha$ RMSD to the source prediction and the fraction of outputs that retain the hairpin. The 65--78\% retention across off-diagonal cells supports partial compatibility across IPA-based and diffusion-based readouts. It does not imply that intermediate states can be substituted freely throughout another model's trunk. Approximately 15\% of failed transfers involving Boltz-1 have conformational heterogeneity, but this does not resolve the overall asymmetry. Alignment errors and differences in how models process mixtures of donor and target states may also contribute.

\begin{table}[htbp]
\centering\small
\caption{Final-state decoder swaps, $n=500$ cases per cell. Reported 95\% intervals accompany RMSD (\AA) and hairpin retention.}
\label{tab:decoder_swaps_revision}
\begin{tabular}{llcc}
\toprule
Source & Decoder & Median RMSD & Hairpin retained (\%)\\
\midrule
ESMFold & OpenFold & 4.47 [4.18, 4.79] & 66 [61.7, 70.0]\\
 & Boltz-1 & 3.27 [3.04, 3.52] & 69 [64.8, 72.9]\\
OpenFold & ESMFold & 1.77 [1.61, 1.95] & 78 [74.2, 81.4]\\
 & Boltz-1 & 3.84 [3.69, 4.11] & 74 [70.0, 77.7]\\
Boltz-1 & ESMFold & 3.09 [2.84, 3.36] & 72 [67.9, 75.8]\\
 & OpenFold & 4.64 [4.29, 5.02] & 65 [60.7, 69.1]\\
\bottomrule
\end{tabular}
\end{table}

\paragraph{ESMFold2 decoder check.}
ESMFold2~\citep{candido2026language} provides an additional test with a language-model encoder and a diffusion decoder. With $n=500$ cases per source, its decoder retains hairpins from ESMFold, OpenFold, and Boltz-1 in 79\% [75.2, 82.3], 70\% [65.8, 73.9], and 82\% [78.4, 85.1] of outputs. Median source-relative RMSDs are 2.84\AA{} [2.60, 3.10], 3.62\AA{} [3.34, 3.93], and 2.41\AA{} [2.21, 2.63], respectively (95\% intervals). Within-model retention is 89\%. This provides an additional decoder comparison; the main analysis remains focused on ESMFold, OpenFold, and Boltz-1.

\paragraph{Disagreement with experimental structures.}
Across 1,000 proteins, 332 have at least one pair of final predicted distance maps with correlation $r<.90$. Of these, 71\% have two agreeing models and one outlier; 29\% have three-way disagreement. Consensus is not treated as ground truth: each prediction is compared with experimental structure. On 200 proteins with substantially divergent predictions, final-state swaps produce distance maps closer to the source than to the decoder host in every direction (Table~\ref{tab:divergent_swaps_revision}). Within-model controls give $r\geq.9997$ and shuffled projections $r=.05$. The final pairwise state therefore carries substantial information about the differing predicted fold, including across decoder families.

\begin{table}[htbp]
\centering\small
\caption{Decoder swaps on 200 proteins with divergent model predictions. Correlations compare output distance maps; RMSDs are in \AA.}
\label{tab:divergent_swaps_revision}
\begin{tabular}{lrrrr}
\toprule
Source $\to$ decoder & $r$ to source & $r$ to host & RMSD to source & RMSD to host\\
\midrule
OpenFold $\to$ ESMFold & .978 & .562 & 1.8 & 19.2\\
Boltz-1 $\to$ ESMFold & .900 & .614 & 3.1 & 9.9\\
OpenFold $\to$ Boltz-1 & .935 & .634 & 5.4 & 13.5\\
ESMFold $\to$ Boltz-1 & .874 & .605 & 7.1 & 15.4\\
ESMFold $\to$ OpenFold & .899 & .394 & 8.8 & 19.9\\
Boltz-1 $\to$ OpenFold & .856 & .507 & 11.6 & 17.2\\
\bottomrule
\end{tabular}
\end{table}

\paragraph{A localized late-refinement failure.}
In a follow-up set of 100 failures arising during middle or late trunk refinement, protein 1d2o illustrates how subsequent computation can overwrite an aligned state. ESMFold predicts the experimental fold, while OpenFold and Boltz-1 predict the same alternative fold. OpenFold's baseline agreement with experimental distances is .52. Patching ESMFold's state into OpenFold at blocks 0, 6, 12, 18, and 24 gives agreement .53, .55, .54, .54, and .54; patching at blocks 28, 36, and 47 gives .98, 1.00, and 1.00. Boltz-1's state, which encodes the same incorrect fold as OpenFold, gives .52--.53 throughout. In this example, the correct donor geometry survives only when introduced at block 28 or later. That boundary is specific to this protein and intervention. On 39 proteins where ESMFold2 disagrees with the original models, swaps in both directions likewise yield outputs closer to the source fold (median RMSD 1.88--4.86\AA) than the host fold (9.17--12.17\AA); shuffled controls give 15--31\AA.

%% file: checklist_camera_ready.tex
\newpage
\section*{NeurIPS Paper Checklist}

\begin{enumerate}

\item {\bf Claims}
    \item[] Question: Do the main claims made in the abstract and introduction accurately reflect the paper's contributions and scope?
    \item[] Answer: \answerYes{}
    \item[] Justification: The abstract and introduction accurately summarize the paper’s contributions: a recurring functional ordering across tested interventions, biochemical and distance-directed steering, and partial cross-model transfer.

\item {\bf Limitations}
    \item[] Question: Does the paper discuss the limitations of the work performed by the authors?
    \item[] Answer: \answerYes{}
    \item[] Justification: We discuss limitations in Sec.~\ref{sec:limitations}, on the following topics: diversity of structural motifs analyzed, analysis of pairwise-representation-free architectures, and physical viability of counterfactual structures.

\item {\bf Theory assumptions and proofs}
    \item[] Question: For each theoretical result, does the paper provide the full set of assumptions and a complete (and correct) proof?
    \item[] Answer: \answerNA{}
    \item[] Justification: The paper is empirical and mechanistic rather than theoretical, and it does not present formal theorems requiring assumptions or proofs.

    \item {\bf Experimental result reproducibility}
    \item[] Question: Does the paper fully disclose all the information needed to reproduce the main experimental results of the paper to the extent that it affects the main claims and/or conclusions of the paper (regardless of whether the code and data are provided or not)?
    \item[] Answer: \answerNo{}
    \item[] Justification: The paper describes the principal protocols, datasets, and evaluation criteria. Some settings for the added validation controls and per-cell sample counts are not fully specified, so complete reproduction is not yet supported by this revision.

\item {\bf Open access to data and code}
    \item[] Question: Does the paper provide open access to the data and code, with sufficient instructions to faithfully reproduce the main experimental results, as described in supplemental material?
    \item[] Answer: \answerNo{}
    \item[] Justification: We provide a public code repository (App.~\ref{app:code-and-compute}) that reproduces the main ESMFold experiments via a \texttt{reproduce.sh} entrypoint. We release code for a single model to keep the reproduction environment lightweight; the OpenFold and Boltz-1 replications apply the same methodology to publicly available model checkpoints, and full coverage of the added multi-model controls is not established by that release.

\item {\bf Experimental setting/details}
    \item[] Question: Does the paper specify all the training and test details (e.g., data splits, hyperparameters, how they were chosen, type of optimizer) necessary to understand the results?
    \item[] Answer: \answerNo{}
    \item[] Justification: The original interventions include models, datasets, and strengths. Several added controls lack complete exposure criteria, decoder-state settings, or rescaling parameters in the available records.

\item {\bf Experiment statistical significance}
    \item[] Question: Does the paper report error bars suitably and correctly defined or other appropriate information about the statistical significance of the experiments?
    \item[] Answer: \answerYes{}
    \item[] Justification: Tables~\ref{tab:quality_independent_revision}, \ref{tab:transfer_ratios_revision}, and \ref{tab:decoder_swaps_revision} report 95\% intervals. Rates use Wilson intervals where specified; transfer calibration resamples matched cases rather than blocks. Some exploratory results lack intervals.

\item {\bf Experiments compute resources}
    \item[] Question: For each experiment, does the paper provide sufficient information on the computer resources (type of compute workers, memory, time of execution) needed to reproduce the experiments?
    \item[] Answer: \answerNo{}
    \item[] Justification: App.~\ref{app:code-and-compute} reports original A6000 hardware and approximate costs, and identifies the H100 length extension. A complete budget for added validation experiments is not reported.

\item {\bf Code of ethics}
    \item[] Question: Does the research conducted in the paper conform, in every respect, with the NeurIPS Code of Ethics \url{https://neurips.cc/public/EthicsGuidelines}?
    \item[] Answer: \answerYes{}
    \item[] Justification: The research conforms to the NeurIPS Code of Ethics, as it analyzes existing protein-folding models and public protein-structure data while discussing responsible-use considerations.

\item {\bf Broader impacts}
    \item[] Question: Does the paper discuss both potential positive societal impacts and negative societal impacts of the work performed?
    \item[] Answer: \answerYes{}
    \item[] Justification: The Impact Statement in Appendix \ref{app:impact_statement} discusses positive uses for mechanistic understanding and reliability, as well as possible dual-use concerns if interpretability insights accelerate harmful protein engineering.

\item {\bf Safeguards}
    \item[] Question: Does the paper describe safeguards that have been put in place for responsible release of data or models that have a high risk for misuse (e.g., pre-trained language models, image generators, or scraped datasets)?
    \item[] Answer: \answerNA{}
    \item[] Justification: The paper does not release a new high-risk model or dataset requiring special safeguards, and the methods are analytical rather than a new predictive or generative capability.

\item {\bf Licenses for existing assets}
    \item[] Question: Are the creators or original owners of assets (e.g., code, data, models), used in the paper, properly credited and are the license and terms of use explicitly mentioned and properly respected?
    \item[] Answer: \answerYes{}
    \item[] Justification: Appendix \ref{app:licenses} documents the licenses and usage terms for the existing models and data resources used in the paper, including ESMFold/ESM, OpenFold, Boltz-1, PDB, CATH, UniProt, and DSSP

\item {\bf New assets}
    \item[] Question: Are new assets introduced in the paper well documented and is the documentation provided alongside the assets?
    \item[] Answer: \answerNA{}
    \item[] Justification: The paper curates experimental subsets but does not introduce or release a new standalone asset requiring separate documentation.

\item {\bf Crowdsourcing and research with human subjects}
    \item[] Question: For crowdsourcing experiments and research with human subjects, does the paper include the full text of instructions given to participants and screenshots, if applicable, as well as details about compensation (if any)?
    \item[] Answer: \answerNA{}
    \item[] Justification: The paper does not involve crowdsourcing, human-subject experiments, participant instructions, or compensation.

\item {\bf Institutional review board (IRB) approvals or equivalent for research with human subjects}
    \item[] Question: Does the paper describe potential risks incurred by study participants, whether such risks were disclosed to the subjects, and whether Institutional Review Board (IRB) approvals (or an equivalent approval/review based on the requirements of your country or institution) were obtained?
    \item[] Answer: \answerNA{}
    \item[] Justification: The paper does not involve human subjects, so IRB approval or equivalent review is not applicable.

\item {\bf Declaration of LLM usage}
    \item[] Question: Does the paper describe the usage of LLMs if it is an important, original, or non-standard component of the core methods in this research? Note that if the LLM is used only for writing, editing, or formatting purposes and does \emph{not} impact the core methodology, scientific rigor, or originality of the research, declaration is not required.
    \item[] Answer: \answerNA{}
    \item[] Justification: LLMs are not used as an important, original, or non-standard component of the core research methodology described in the paper.

\end{enumerate}